\documentclass[12pt]{article}
\usepackage{rotating}

\usepackage{url}
\usepackage{graphicx}
\usepackage{amssymb}
\usepackage{amsmath}
\usepackage{amsbsy}
\usepackage{latexsym}
\usepackage{subfigure}
\usepackage{url}
\usepackage{amsthm}
\usepackage{multirow}
\usepackage{bm}
\usepackage{fixltx2e}
\usepackage{xcolor}

\newtheorem{theorem}{Theorem}

\newtheorem{corollary}[theorem]{Corollary}

\DeclareMathOperator{\tr}{\mathrm{tr}}

\DeclareMathOperator{\var}{\mathbf{Var}}
\DeclareMathOperator{\cov}{\mathbf{Cov}}

\newcommand{\T}{{\hspace{-0.25ex}\top\hspace{-0.25ex}}}
\newcommand{\bE}{\mathbb{E}}
\newcommand{\cN}{\mathcal{N}}
\newcommand{\cJ}{\mathcal{J}}

\title{
\begin{flushleft}
  \normalsize
  \sl
  \textit{Neural Computation},
 to appear.
\end{flushleft}
\vspace*{5mm}
Efficient Sample Reuse\\
  in Policy Gradients with Parameter-based Exploration}
\author{Tingting Zhao, Hirotaka Hachiya, Voot Tangkaratt, \\
Jun Morimoto, and Masashi Sugiyama}

\date{}

\begin{document}
\maketitle

\begin{abstract}\noindent
The policy gradient approach is a flexible and powerful reinforcement learning method
particularly for problems with continuous actions such as robot control.
A common challenge in this scenario is how to reduce the variance of
policy gradient estimates for reliable policy updates.
In this paper, we combine the following three ideas and give
a highly effective policy gradient method:
(a) the \emph{policy gradients with parameter based exploration},
which is a recently proposed policy search method with low variance of gradient estimates,
(b) an \emph{importance sampling technique}, which allows us to reuse
previously gathered data in a consistent way,
and
(c) an \emph{optimal baseline}, which minimizes the variance of gradient estimates
with their unbiasedness being maintained.
For the proposed method,
we give theoretical analysis of the variance of gradient estimates
and show its usefulness through extensive experiments.
\end{abstract}

\section{Introduction}

The objective of \emph{reinforcement learning} (RL) is to let an agent
optimize its decision-making policy
through interaction with an unknown environment \cite{book:Sutton+Barto:1998}.
Among possible approaches,
\emph{policy search} has become a popular method
because of its direct nature for policy learning \cite{Bagnell}.
Particularly, in high-dimensional problems with continuous states and actions,
policy search has been shown to be highly useful in practice
\cite{Ng00pegasus:a,IROS:Peters+Schaal:2006}.

Among policy search methods \cite{BusBab:10-002},
gradient-based methods are popular in physical control tasks
because policies are changed gradually \cite{Sutton99policygradient,nips02-CN11,IROS:Peters+Schaal:2006} and thus
steady performance improvement is ensured until a local optimal policy has been obtained.
However, since the gradients estimated with these methods tend to have large variance and thus
they may suffer from slow convergence.

Recently, a novel approach to using policy gradients
called \emph{policy gradients with parameter based exploration} (PGPE) was proposed
\cite{NN:Sehnke+etal:2010}.
PGPE tends to produce gradient estimates with low variance by
removing unnecessary randomness from policies and
introducing useful stochasticity by considering
a prior distribution for policy parameters.
PGPE was shown to be more promising than alternative approaches
experimentally \cite{NN:Sehnke+etal:2010,NN2012-ting}.
However, PGPE still requires a relatively large number of samples
to obtain accurate gradient estimates, which can be
a critical bottleneck in real-world applications
that require large costs and time in data collection.

To overcome this weakness,
an \emph{importance sampling} technique \cite{book:Fishman:1996}
is useful under the \emph{off-policy} RL scenario, where
a data-collecting policy and the current target policy are different
in general \cite{book:Sutton+Barto:1998}.
An importance sampling technique allows us to reuse previously collected
data, which are collected following policies different from the current one
in a consistent manner \cite{book:Sutton+Barto:1998,Shimodaira2000}.
However, naively using an importance sampling technique significantly
increases the variance of gradient estimates,
which can cause sudden changes in policy updates \cite{She01,Leonid02,R3,truncated}.
To mitigate this problem, variance reduction techniques such as
decomposition \cite{Precup00}, truncation \cite{truncated,Uchibe04},
normalization \cite{She01,Leonid02}, and flattening \cite{R3} of importance weights
are often used.
However, these methods commonly suffer from the bias-variance trade-off,
meaning that the variance is reduced at the expense of increasing the bias.

The purpose of this paper is to propose a new approach to
systematically addressing the large variance problem in policy search.
Basically, this work is an extension of our previous research \cite{NN2012-ting} to an \emph{off-policy} scenario using an importance weighting technique.
More specifically, we first give an off-policy implementation of PGPE
called the \emph{importance-weighted PGPE} (IW-PGPE) method
for consistent sample reuse.
We then derive the optimal baseline for IW-PGPE
to minimize the variance of importance-weighted gradient estimates,
following \cite{JMLR:Greensmith+Bartlett+Baxter:2004,UAI:Weaver+Tao:2001}.
We show that
the proposed method can achieve significant performance improvement
over alternative approaches in experiments with an artificial domain.
We also investigate that combining the proposed method with the truncation technique
can further improve the performance in high-dimensional problems.

\section{Formulations of Policy Gradient}
In this paper, we consider the standard framework of episodic reinforcement learning (RL)
in which an agent interacts with an environment modeled as a \emph{Markov decision process}
(MDP)
\cite{book:Sutton+Barto:1998}.
In this section, we first review a standard formulation of policy gradient methods
\cite{mach:Williams:1992,nips02-CN11,IROS:Peters+Schaal:2006}.
Then we show an alternative formulation adopted in the PGPE (policy gradients with
parameter based exploration) method \cite{NN:Sehnke+etal:2010}.

\subsection{Standard Formulation}
\label{subsec:RLformulation}
We assume that the underlying control problem is a discrete-time MDP.
At each discrete time step $t$, the agent observes a state $\bm{s}_t \in {\cal S}$, selects an action $a_t \in {\cal A}$,
and then receives an immediate reward $r_t$ resulting from a state transition in the environment.
The state ${\cal S}$ and action ${\cal A}$ are both defined as continuous spaces in this paper\footnote{Note that continuous formulation is not an essential restriction.}.
The dynamics of the environment are characterized by $p(\bm{s}_{t+1}|\bm{s}_{t},a_t)$, which represents the transition probability density from the current state $\bm{s}_t$ to
the next state $\bm{s}_{t+1}$ when action $a_t$ is taken, and $p(\bm{s}_1)$ is the probability density of initial states.
The immediate reward $r_t$ is given according to the reward function $r(\bm{s}_t,a_t,\bm{s}_{t+1})$.

The agent's decision making procedure at each time step $t$ is characterized by a parameterized policy $p(a_t|\bm{s}_t,\bm{\theta})$
with parameter $\bm{\theta}$,
which represents the conditional probability
density of taking action $a_t$ in state $\bm{s}_t$.
We assume that the policy is continuously differentiable with respect to its parameter $\bm{\theta}$.

A sequence of states and actions forms a \emph{trajectory} denoted by
\[
h:=[\bm{s}_{1},a_{1},\ldots,\bm{s}_{T},a_{T}],
\]
where $T$ denotes the number of steps called horizon length.
In this paper, we assume that $T$ is a fixed deterministic number.
Note that the action $a_t$ is chosen independently of the trajectory given $\bm{s}_t$ and $\bm{\theta}$.
Then the discounted cumulative reward along $h$, called the \emph{return},
is given by
\[
R(h):=
\sum_{t=1}^T \gamma ^{t-1}r(\bm{s}_t,a_t,\bm{s}_{t+1}),
\]
where $\gamma \in [0, 1)$ is the discount factor for future rewards.

The goal is to optimize the policy parameter $\bm{\theta}$ so that the
\emph{expected return} is maximized.
The expected return for policy parameter $\bm{\theta}$ is defined by
\[
J(\bm{\theta}):=\int p(h|\bm{\theta})R(h)\mathrm{d}h, \]
where
\[ p(h|\bm{\theta})=p(\bm{s}_1)\prod_{t=1} ^T p(\bm{s}_{t+1}|\bm{s}_t,a_t)p(a_t|\bm{s}_t,\bm{\theta}).\]
The most straightforward way to update the policy parameter is to follow the gradient in policy parameter space using gradient ascent:
\[
\bm{\theta}\longleftarrow\bm{\theta}+\varepsilon\nabla_{\bm{\theta}} J(\bm{\theta}),
\]
where $\varepsilon$ is a small positive constant, called the learning rate.

This is a standard formulation of policy gradient methods \cite{mach:Williams:1992,nips02-CN11,IROS:Peters+Schaal:2006}.
The central problem is to estimate the policy gradient $\nabla_{\bm{\theta}} J(\bm{\theta})$ accurately from trajectory samples.

\subsection{Alternative Formulation}

However, standard policy gradient methods were shown to suffer from high variance in the gradient estimation
due to randomness introduced by the stochastic policy model $p(a|\bm{s},\bm{\theta})$ \cite{NN2012-ting}.
To cope with this problem, an alternative method
called \emph{policy gradients with parameter based exploration} (PGPE) was proposed
recently \cite{NN:Sehnke+etal:2010}.
The basic idea of PGPE is to use a deterministic policy
and introduce stochasticity by drawing parameters from a prior distribution.
More specifically, parameters are sampled from the prior distribution at the start of each trajectory,
and thereafter the controller is deterministic\footnote{Note that transitions are stochastic, and thus trajectories are also stochastic even though the policy is deterministic.}.
Thanks to this per-trajectory formulation,
the variance of gradient estimates in PGPE does not increase with respect to trajectory length $T$.
Below, we review PGPE.

PGPE uses a deterministic policy with typically a linear architecture:
\begin{align}
p(a|\bm{s},\bm{\theta})
=\delta(a=\bm{\theta}^\T \bm{\phi} (\bm{s})),
\label{linear-policy-model}
\end{align}
where $\delta(\cdot)$ is the \emph{Dirac delta function},
$\bm{\phi} (\bm{s})$ is an $\ell$-dimensional basis function vector,
and $^\T$ denotes the transpose.
The policy parameter $\bm{\theta}$ is drawn from
a prior distribution $p(\bm{\theta}|\bm{\rho})$ with hyper-parameter $\bm{\rho}$.

The expected return in the PGPE formulation is defined
in terms of expectations over both $h$ and $\bm{\theta}$
as a function of hyper-parameter $\bm{\rho}$:
\[
\cJ(\bm{\rho}):=\iint p(h|\bm{\theta})p(\bm{\theta}|\bm{\rho})
R(h)\mathrm{d}h\mathrm{d}\bm{\theta}.
\]
In PGPE, the hyper-parameter $\bm{\rho}$ is optimized so as to maximize $\cJ(\bm{\rho})$,
i.e., the optimal hyper-parameter $\bm{\rho}^*$ is given by
\[
\bm{\rho}^*:= \arg\max_{\bm{\rho}} \cJ(\bm{\rho}).
\]

In practice, a gradient method is used to find $\bm{\rho}^*$:
\[
\bm{\rho}\longleftarrow\bm{\rho}+\varepsilon\nabla_{\bm{\rho}} \cJ(\bm{\rho}),
\]
where
$\nabla_{\bm{\rho}} \cJ(\bm{\rho})$ is the derivative of $\cJ$ with respect to $\bm{\rho}$:
\begin{align}
\nabla_{\bm{\rho}} \cJ(\bm{\rho})=\iint p(h|\bm{\theta})p(\bm{\theta}|\bm{\rho})  \nabla_{\bm{\rho}} \log p(\bm{\theta}|\bm{\rho}) R(h)\mathrm{d}h\mathrm{d}\bm{\theta}.
\end{align}
Note that, in the derivation of the gradient, the logarithmic derivative,
\[
\nabla_{\bm{\rho}}  \log p(\bm{\theta}|\bm{\rho})=
\frac{\nabla_{\bm{\rho}}p(\bm{\theta}|\bm{\rho})}{p(\bm{\theta}|\bm{\rho})},
\]
was used.
The expectations over $h$ and $\bm{\theta}$ are approximated by the empirical averages:
\begin{align}
\label{emp_gra}
\nabla_{\bm{\rho}} \widehat{\cJ}(\bm{\rho}) = \frac {1} {N} \sum ^N_{n=1} \nabla_{\bm{\rho}} \log p(\bm{\theta}_n|\bm{\rho})R(h_n),
\end{align}
where each trajectory sample $h_n$ is drawn independently from $p(h|\bm{\theta}_n)$ and
parameter $\bm{\theta}_n$ is drawn from $p(\bm{\theta}_n|\bm{\rho})$.
We denote samples collected at the current iteration as
\[
D=\{\left(\bm{\theta}_n, h_n\right)\}_{n=1}^N.
\]

Following \cite{NN:Sehnke+etal:2010}, in this paper we employ a Gaussian distribution as the distribution of the policy parameter $\bm{\theta}$
with the hyper-parameter $\bm{\rho}$.
However, other distributions can also be allowed. When assuming a Gaussian distribution,
the hyper-parameter $\bm{\rho}$ consists of a set of means $\{{\eta}_i\}$ and standard deviations $\{\tau_i\}$,
which determine the prior distribution for each element $\theta_i$ in $\bm{\theta}$ of the form
\[p(\theta_i|\rho_i)=\cN(\theta_i|\eta_i,\tau_i^2),\]
where $\cN(\theta_i|\eta_i,\tau_i^2)$ denotes the normal distribution with mean $\eta_i$ and variance $\tau_i^2$.
Then the derivative of $\log p(\bm{\theta}|\bm{\rho})$ with respect to $\eta_i$ and $\tau_i$ are given as
\begin{align*}
  \nabla_{\eta_i}\log p(\bm{\theta}|\bm{\rho})=&\frac{\theta_i-\eta_i}{\tau_i^2},\\
  \nabla_{\tau_i} \log p(\bm{\theta}|\bm{\rho})=&\frac{(\theta_i-\eta_i)^2-\tau_i^2}{\tau_i^3},
\end{align*}
which can be substituted into Eq.(\ref{emp_gra}) to approximate the
gradients with respect to $\bm{\eta}$ and $\bm{\tau}$.
These gradients give the PGPE update rules.

An advantage of PGPE is its low variance of gradient estimates:
Compared with a standard policy gradient method REINFORCE \cite{mach:Williams:1992},
PGPE was empirically demonstrated to be better in some settings \cite{NN:Sehnke+etal:2010, NN2012-ting}.
The variance of gradient estimates in PGPE
can be further reduced by subtracting an optimal baseline (Theorem 4 of \cite{NN2012-ting}).

Another advantage of PGPE is its high flexibility:
In standard policy gradient methods, the parameter $\bm{\theta}$ is used to determine
a stochastic policy model $p(a|\bm{s},\bm{\theta})$,
and policy gradients are calculated by differentiating the policy with respect to the parameter.
However, because PGPE needs not calculate the derivative of the policy,
a non-differentiable controller is also allowed.

\section{Off-Policy Extension of PGPE}
In real-world applications such as robot control,
gathering roll-out data is often costly.
Thus, we want to keep the number of samples as small as possible.
However, when the number of samples is small,
policy gradients estimated by the original PGPE are not reliable enough.

The original PGPE is categorized as an \emph{on-policy} algorithm \cite{book:Sutton+Barto:1998},
where data drawn from the current target policy is used to estimate policy gradients.
On the other hand, \emph{off-policy} algorithms are more flexible in the sense
that a data-collecting policy and the current target policy can be different.
In this section, we extend PGPE to an \emph{off-policy} scenario using importance-weighting,
which allows us to reuse previously collected data in a consistent manner.
We also theoretically analyze properties of the extended method.

\subsection{Importance-Weighted PGPE}

Let us consider an off-policy scenario where
a data-collecting policy and the current target policy are different in general.
In the context of PGPE, we consider two hyper-parameters, $\bm{\rho}$ for the target policy to learn
and $\bm{\rho}'$ for data collection. Let us denote data samples collected with hyper-parameter $\bm{\rho}'$ by $D'$:
\[
D'=\{\left(\bm{\theta}'_n, h'_n\right)\}_{n=1}^{N'} \overset{i.i.d}{\sim}
p(h,\bm{\theta}|\bm{\rho}')=p(h|\bm{\theta})p(\bm{\theta}|\bm{\rho}').
\]
If we naively use data $D'$ to estimate policy gradients by Eq.\eqref{emp_gra},
we have an inconsistency problem:
\[
\frac {1} {N'} \sum ^{N'}_{n=1} \nabla_{\bm{\rho}} \log p(\bm{\theta}'_n|\bm{\rho})R(h_n')
\overset{N'\rightarrow\infty}{\nrightarrow}
\nabla_{\bm{\rho}} \cJ(\bm{\rho}),
\]
which we refer to as ``\emph{non-importance-weighted PGPE}'' (NIW-PGPE).

\emph{Importance sampling} \cite{book:Fishman:1996}
is a technique to systematically resolve this distribution mismatch problem.
The basic idea of importance sampling is to weight samples drawn from a sampling distribution
to match the target distribution, which gives a consistent gradient estimator:
\[
\nabla_{\bm{\rho}}\widehat{\cJ}_{\mathrm{IW}}(\bm{\rho})
:=\frac {1} {N'}\sum ^{N'}_{n=1} w(\bm{\theta}'_n) \nabla_{\bm{\rho}} \log p(\bm{\theta}'_n|\bm{\rho})R(h_n')
\overset{N'\rightarrow\infty}{\longrightarrow} \nabla_{\bm{\rho}} \cJ(\bm{\rho}),
\]
where
\[
w(\bm{\theta})=\frac{p(\bm{\theta}|\bm{\rho})}{p(\bm{\theta}|\bm{\rho}')}
\]
is called the \emph{importance weight}.

An intuition behind importance sampling is that if we know how ``important''
a sample drawn from the sampling distribution is in the target distribution,
we can make adjustment by importance weighting.
We call this extended method \emph{importance-weighted PGPE} (IW-PGPE).

Now we analyze the variance of gradient estimates in IW-PGPE.
For a multi-dimensional space,
we consider the \emph{trace} of the covariance matrix of gradient vectors.
That is, for a random vector $\bm{A}=(A_1,\ldots,A_\ell)^\T$, we define
\begin{align}
\label{def-var}
\var(\bm{A})=&\tr\Big(\bE\big[(\bm{A}-\bE[\bm{A}])(\bm{A}-\bE[\bm{A}])^\T\big]\Big)\nonumber,\\
=&\sum_{m=1}^\ell \bE\Big[(A_m-\bE[A_m])^2\Big],
\end{align}
where $\bE$ denotes the expectation.

Let
\[
B=\sum_{i=1}^{\ell}\tau_i^{-2},
\]
where $\ell$ is the dimensionality of
the basis function vector $\bm{\phi}(\bm{s})$.
For a $\bm{\rho}=(\bm{\eta}, \bm{\tau})$, we have the following theorem\footnote{
Proofs of all theorems are provided in Appendix,
which are basically extensions of the proofs for the plain PGPE given
in \cite{NN2012-ting} to importance-weighting scenarios.
}:

\begin{theorem}
\label{theorem:variance-bound-PGPE}
Assume that for all $\bm{s}$, $a$, and $\bm{s}'$, there exists $\beta>0$ such that $r(\bm{s},a,\bm{s}')\in [-\beta,\beta]$,
and, for all $\bm{\theta}$, there exists $0<w_{\max}< \infty$ such that $0<w(\bm{\theta})\leq w_{\max}$.
Then we have the following upper bounds:
\begin{align*}
\var\left[\nabla_{\bm{\eta}} \widehat{\cJ}_{\mathrm{IW}}(\bm{\rho})\right]&\le
\frac{\beta^2(1-\gamma^T)^2 B}{N'(1-\gamma)^2}w_{\max},\\
\var\left[\nabla_{\bm{\tau}} \widehat{\cJ}_{\mathrm{IW}}(\bm{\rho})\right]&\le
\frac{2\beta^2(1-\gamma^T)^2 B}{N'(1-\gamma)^2}w_{\max}.
\end{align*}
\end{theorem}

Theorem \ref{theorem:variance-bound-PGPE} shows that the upper bound of the variance of $\nabla_{\bm{\eta}} \widehat{\cJ}_{\mathrm{IW}}(\bm{\rho})$
is proportional to $\beta^2$ (the upper bound of squared rewards), $w_{\max}$ (the upper bound of the importance weight $w(\bm{\theta})$),
$B$ (the trace of the inverse Gaussian covariance),
and $(1-\gamma^T)^2/(1-\gamma)^2$,
and is inverse-proportional to sample size $N'$.
It is interesting to see that
the upper bound of the variance of
$\nabla_{\bm{\tau}} \widehat{\cJ}_{\mathrm{IW}}(\bm{\rho})$ is twice larger than
that of $\nabla_{\bm{\eta}} \widehat{\cJ}_{\mathrm{IW}}(\bm{\rho})$.

It is also interesting to see that the upper bounds are the same as the upper bounds for the plain PGPE (Theorem 1 of \cite{NN2012-ting}) except for the factor $w_{max}$;
when $w_{\max}=1$, the bounds are reduced to those of the plain PGPE method.
However, if the sampling distribution is significantly different from the target distribution,
$w_{\max}$ can take a large value and thus IW-PGPE tends to produce a gradient estimator with large variance (at least in terms of its upper bound).
Therefore, IW-PGPE may not be a reliable approach as it is.

Below, we give a variance reduction technique for IW-PGPE,
which leads to a highly effective policy gradient algorithm.

\subsection{Variance Reduction by Baseline Subtraction for IW-PGPE}
\label{sec:variance}

To cope with the large variance of gradient estimates in IW-PGPE,
several techniques have been developed in the context of sample reuse,
for example,
by flattening \cite{R3},
truncating \cite{truncated},
and normalizing \cite{She01} the importance weight.
Indeed, from Theorem~\ref{theorem:variance-bound-PGPE},
we can see that
decreasing $w_{\max}$ by flattening or truncating the importance weight
reduces the upper bounds of the variance of gradient estimates.
However, all of those techniques are based on the bias-variance trade-off,
and thus they lead to biased estimators.

Another, and possibly more promising variance reduction technique
is subtraction of a constant \emph{baseline} \cite{sutton:dissertation84,Williams:88,JMLR:Greensmith+Bartlett+Baxter:2004,UAI:Weaver+Tao:2001},
which reduces the variance \emph{without} increasing the bias.
Here, we derive an optimal baseline for IW-PGPE to minimize the variance,
and analyze its theoretical properties.

A policy gradient estimator with a baseline $b\in\mathbb{R}$ is defined as
\[
\nabla_{\bm{\rho}}\widehat{\cJ}^b_{\mathrm{IW}}(\bm{\rho})
:=\frac{1}{N'}\sum_{n=1}^{N'} (R(h'_n)-b) w(\bm{\theta}'_n) \nabla_{\bm{\rho}} \log p(\bm{\theta}'_n|\bm{\rho}).
\]
 It is well known that $\nabla_{\bm{\rho}}\widehat{\cJ}^b_{\mathrm{IW}}(\bm{\rho})$
is still a consistent estimator of the true gradient for any constant $b$ \cite{JMLR:Greensmith+Bartlett+Baxter:2004}.
Here, we determine the constant baseline $b$ so that the variance is minimized,
following the line of \cite{NN2012-ting}.
Let $b^*$ be the optimal constant baseline for IW-PGPE that minimizes the variance:
\[
b^*:= \arg\min_b \var[\nabla_{\bm{\rho}}\widehat{\cJ}_{\mathrm{IW}}^b(\bm{\rho})].
\]
Then the following theorem gives the optimal constant baseline for IW-PGPE:

\begin{theorem}
\label{theorem:optimal-baseline}
The optimal constant baseline for IW-PGPE is given by
\[
b^*= \frac{\bE_{p(h,\bm{\theta}|\bm{\rho'})}[R(h)w^2(\bm{\theta})\| \nabla_{\bm{\rho}} \log p(\bm{\theta}|\bm{\rho})\|^2]}
      {\bE_{p(h,\bm{\theta}|\bm{\rho'})}[w^2(\bm{\theta})\|\nabla_{\bm{\rho}} \log p(\bm{\theta}|\bm{\rho})\|^2]},\]
and the excess variance for a constant baseline $b$ is given by
\[
\var[\nabla_{\bm{\rho}} \widehat{\cJ}^b_{\mathrm{IW}}(\bm{\rho})]-\var[\nabla_{\bm{\rho}} \widehat{\cJ}^{b^*}_{\mathrm{IW}}(\bm{\rho})]
=\frac{(b-b^*)^2}{N'} \bE_{p(h,\bm{\theta}|\bm{\rho'})}[w^2(\bm{\theta})\| \nabla_{\bm{\rho}} \log p(\bm{\theta}|\bm{\rho})\|^2],
\]
where $\bE_{p(h,\bm{\theta}|\bm{\rho'})}[\cdot]$ denotes the expectation of the function of random variables $h$ and $\bm{\theta}$ with
respect to $\left(h, \bm{\theta}\right)\sim p(h,\bm{\theta}|\bm{\rho'})$.
\end{theorem}

The above theorem gives an analytic expression of the optimal constant baseline for IW-PGPE.
It also shows that the excess variance is
proportional to the squared difference of baselines $(b-b^*)^2$
and
the expectation of the product of squared importance weight $w(\bm{\theta})$
and the squared norm of characteristic eligibility
$\|\nabla_{\bm{\rho}} \log p(\bm{\theta}|\bm{\rho})\|^2$,
and is inverse-proportional to sample size $N'$.

Next, we analyze contributions of the optimal baseline to variance reduction in IW-PGPE:

\begin{theorem}
\label{theorem:variance-bound-gap}
Assume that for all $\bm{s}$, $a$, and $\bm{s}'$, there exists $\alpha>0$ such that $r(\bm{s},a,\bm{s}') \geq \alpha$,
and, for all $\bm{\theta}$, there exists $w_{\min}>0$ such that $w(\bm{\theta})\geq w_{\min}$.
Then we have the following lower bounds:
\begin{align*}
 \var\left[\nabla_{\bm{\eta}} \widehat{\cJ}_{\mathrm{IW}}(\bm{\rho})\right]-\var\left[\nabla_{\bm{\eta}} \widehat{\cJ}_{\mathrm{IW}}^{b^*}(\bm{\rho})\right] & \geq
 \frac{\alpha^2(1-\gamma^T)^2 B}{N'(1-\gamma)^2}w_{\min},\\
\var\left[\nabla_{\bm{\tau}} \widehat{\cJ}_{\mathrm{IW}}(\bm{\rho})\right]-\var\left[\nabla_{\bm{\tau}} \widehat{\cJ}_{\mathrm{IW}}^{b^*}(\bm{\rho})\right] & \geq
\frac{2\alpha^2(1-\gamma^T)^2 B}{N'(1-\gamma)^2}w_{\min}.
\end{align*}
Assume that for all $\bm{s}$, $a$, and $\bm{s}'$, there exists $\beta>0$ such that $r(\bm{s},a,\bm{s}')\in [-\beta,\beta]$,
and, for all $\bm{\theta}$, there exists $0<w_{\max}< \infty$ such that $0<w(\bm{\theta})\leq w_{\max}$.
Then we have the following upper bounds:
\begin{align*}
\var\left[\nabla_{\bm{\eta}} \widehat{\cJ}_{\mathrm{IW}}(\bm{\rho})\right]-\var\left[\nabla_{\bm{\eta}} \widehat{\cJ}_{\mathrm{IW}}^{b^*}(\bm{\rho})\right]&\le
\frac{\beta^2(1-\gamma^T)^2 B}{N'(1-\gamma)^2}w_{\max},\\
\var\left[\nabla_{\bm{\tau}} \widehat{\cJ}_{\mathrm{IW}}(\bm{\rho})\right]-\var\left[\nabla_{\bm{\tau}} \widehat{\cJ}_{\mathrm{IW}}^{b^*}(\bm{\rho})\right]&\le
\frac{2\beta^2(1-\gamma^T)^2 B}{N'(1-\gamma)^2}w_{\max}.
\end{align*}
\end{theorem}

This theorem shows that the bounds of the variance reduction in IW-PGPE brought by the optimal constant baseline depend on the bounds of the importance weight.
If importance weights are larger, using the optimal baseline can reduce the variance more.

Based on Theorems \ref{theorem:variance-bound-PGPE} and \ref{theorem:variance-bound-gap}, we get the following corollary:

\begin{corollary}
\label{corollary:variance-ob}
Assume that for all $\bm{s}$, $a$, and $\bm{s}'$, there exists $0<\alpha<\beta$ such that $r(\bm{s},a,\bm{s}')\in [\alpha,\beta]$,
and, for all $\bm{\theta}$, there exists $0<w_{\min}<w_{\max}< \infty$ such that $w_{\min} \leq w(\bm{\theta})\leq w_{\max}$.
Then we have the following upper bounds:
\begin{align*}
\var\left[\nabla_{\bm{\eta}} \widehat{\cJ}_{\mathrm{IW}}^{b^*}(\bm{\rho})\right] & \le
\frac{(1-\gamma^T)^2 B}{N'(1-\gamma)^2}(\beta^2 w_{\max}-\alpha^2 w_{\min}),\\
\var\left[\nabla_{\bm{\tau}} \widehat{\cJ}_{\mathrm{IW}}^{b^*}(\bm{\rho})\right] & \le
\frac{2(1-\gamma^T)^2 B}{N'(1-\gamma)^2}(\beta^2 w_{\max}-\alpha^2 w_{\min}).
\end{align*}
\end{corollary}

Comparing Theorem \ref{theorem:variance-bound-PGPE} and this corollary,
we can see that the upper bounds for IW-PGPE
with the optimal constant baseline
are smaller than those for IW-PGPE with no baseline
because $\alpha^2 w_{\min}>0$.
Although they are just upper bounds, they can still intuitively show that
subtraction of the optimal constant baseline contributes to mitigating
the large variance caused by importance weighting.
If $w_{\min}$ is larger, then the upper bounds for IW-PGPE with the optimal constant baseline
can be much smaller than those for IW-PGPE with no baseline.

\section{Experimental Results}
In this section, we experimentally investigate the usefulness of the proposed method,
importance-weighted PGPE with the optimal constant baseline
(which we denote by IW-PGPE\textsubscript{OB} hereafter).
In the experiments, we estimate the optimal constant baseline using all
collected data, as suggested in
\cite{JMLR:Greensmith+Bartlett+Baxter:2004,IROS:Peters+Schaal:2006,UAI:Weaver+Tao:2001}.
This approach introduces bias into the method because the same sample-set is used both for estimating the gradient and the baseline.
Another possibility is to split the data into two parts:
One is used for estimating the optimal constant baseline
and the other is used for estimating the gradient.
However, we found that this splitting approach does not work well in our preliminary experiments.
The MATLAB implementation of IW-PGPE\textsubscript{OB} is available from:
\url{http://sugiyama-www.cs.titech.ac.jp/~tingting/software.html}.

\subsection{Illustrative Example}
First, we illustrate the behavior of PGPE methods using a toy dataset.

\subsubsection{Setup}

The dynamics of the environment is defined as
\[
s_{t+1}=s_t+a_t+\varepsilon,
\]
where $s_t\in \mathbb{R}$, $a_t \in \mathbb{R}$, and $\varepsilon \sim \cN(0,0.5^2)$ is stochastic noise.
The initial state $s_1$ is randomly chosen from the standard normal distribution.
The linear deterministic controller is represented by $a_t=\theta s_t $ for $\theta \in \mathbb{R}$.
The immediate reward function is given by
\[
r(s_t,a_t)=\exp\left(-s_t^2/2-a_t^2/2\right)+1,
\]
which is bounded in $(1, 2]$.
In the toy dataset experiments, we always set the discount factor at $\gamma=0.9$, and we always use
the adaptive learning rate $\varepsilon=0.1/\|\nabla_{\rho} \hat{\cJ}(\bm{\rho})\|$ \cite{MatsubaraMM10}.

Here, we compare the following PGPE methods:
\begin{itemize}
 \item \textbf{PGPE:}
    Plain PGPE without data reuse \cite{NN:Sehnke+etal:2010}.
  \item \textbf{PGPE\textsubscript{OB}:}
    Plain PGPE with the optimal constant baseline without data reuse \cite{NN2012-ting}.
  \item \textbf{NIW-PGPE:}
    Data-reuse PGPE without importance weights.
  \item \textbf{NIW-PGPE\textsubscript{OB}:}
    Data-reuse PGPE\textsubscript{OB} without importance weights.
  \item \textbf{IW-PGPE:}
    Importance-weighted PGPE.
  \item \textbf{IW-PGPE\textsubscript{OB}:}
    Importance-weighted PGPE with the optimal baseline.
\end{itemize}


Suppose that a small amount of samples consisting of $N$ trajectories with length $T$ is available at each iteration.
More specifically,
given the hyper-parameter $\bm{\rho}_L=(\eta_L,\tau_L)$ at the $L$th iteration,
we first choose the policy parameter $\theta_n^L$ from $p(\theta|\bm{\rho}_L)$,
and then run the agent to generate trajectory $h_n^L$ according to $p(h|\theta_n^L)$.
Initially, the agent starts from a randomly selected state $s_1$ following the initial state probability density $p(s_1)$
and chooses an action based on the policy $p(a_t|s_t, \theta_n^L)$.
Then the agent makes a transition following the dynamics of the environment $p(s_{t+1}|s_t, a_t)$ and receives a reward $r_t=r(s_t,a_t,s_{t+1})$.
The transition is repeated $T$ times to get a trajectory, which is denoted as $h_n^L=\{s_t,a_t,r_t,s_{t+1}\}_{t=1}^T$.
We repeat the procedure $N$ times,
and, the samples gathered at the $L$th iteration is obtained, which is expressed as $D^L=\{(\theta_n^L, h_n^L)\}_{n=1}^N$.

In the data-reuse methods, we estimate gradients at each iteration
based on the current data and all previously collected data $D^{1:L}=\{D^l\}_{l=1}^L$,
by the estimated gradients to update the policy hyper-parameters (i.e., mean $\eta$ and standard deviation $\tau$).
In the plain PGPE method and the plain PGPE\textsubscript{OB} method, we only use the on-policy data $D^L$ to estimate the gradients at each iteration,
by the estimated gradients to update the policy hyper-parameters.
If the deviation parameter $\tau$ takes a value smaller than $0.05$
during the parameter-update process, we set it at $0.05$.

Below, we experimentally evaluate
the variance, bias, and mean squared error of the estimated gradients,
trajectories of learned hyper-parameters, and obtained returns.

\subsubsection{Estimated Gradients}
We investigate how data reuse influences estimated gradients over iterations.
Below, we focus on gradients with respect to the mean parameter $\eta$.

\begin{figure}[t]
\centering
\subfigure[Variance]{\includegraphics[width=0.7\columnwidth]{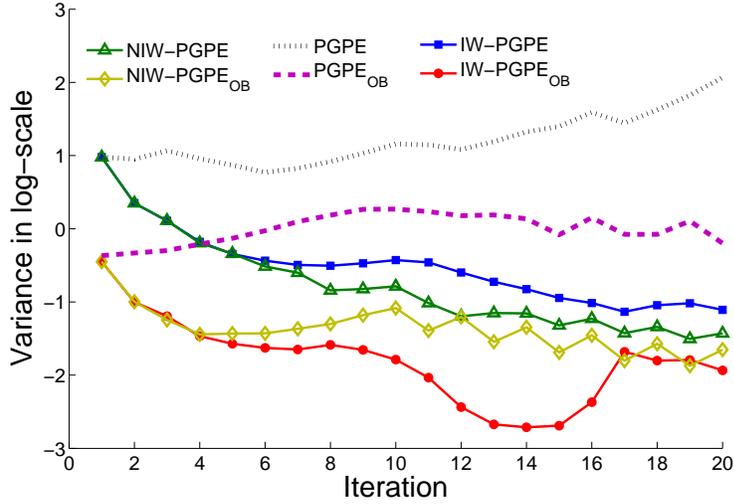}
\label{fig:var}
}
\subfigure[$\operatorname{Bias}^2$]{\includegraphics[width=0.7\columnwidth]{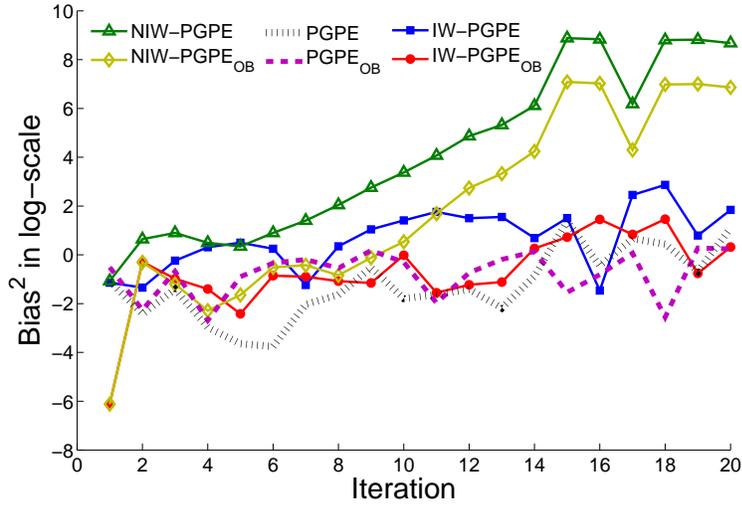}
\label{fig:bias}
}
\caption{Variance and $\operatorname{Bias}^2$ of gradient estimates with respect
  to the mean parameter $\eta$ through parameters update iterations.}
\label{fig:var-bias-mse}
\end{figure}

We randomly choose initial mean parameter $\eta$ from the standard normal distribution,
and fix the initial deviation parameter at $\tau=1$.
We collect $N=10$ trajectories with the trajectory length $T=10$ at each iteration,
and update hyper-parameters over $20$ iterations.
Here, the variance and squared bias of estimated gradients at each iteration (e.g., at the $L$th iteration, $L=1,\ldots, 20$)
are investigated for $M=10000$ trials:
\begin{align*}
\operatorname{Var}&:=\frac{1}{M} \sum_{m=1}^M
\left\|\nabla_{\eta_L} \widehat{\cJ}^m(\bm{\rho}_L)-
\frac{1}{M} \sum_{m'=1}^M  \nabla_{\eta_L} \widehat{\cJ}^{m'}(\bm{\rho}_L)\right\|^2,\\
\operatorname{Bias}^2&:=\left\|\frac{1}{M} \sum_{m=1}^M
\nabla_{\eta_L} \widehat{\cJ}^m(\bm{\rho}_L)-\nabla_{\eta_L} \cJ(\bm{\rho}_L)\right\|^2,
\end{align*}
where $\nabla_{{\eta}_L} \widehat{\cJ}^m(\bm{\rho}_L)$
is an estimated gradient in the $m$-th trial.
More specifically, we estimate the gradients $M$ times with different random seeds at the $L$th iteration as follows:
We generate samples $D_m^{1:L}=\{D_m^l\}_{l=1}^L$ following the corresponding distributions $\{D_m^l\overset{i.i.d}{\sim} p(h, \theta|\bm{\rho}_l)\}_{l=1}^L$ in each trial ($m=1,\ldots,M$),
and we estimate the gradient $\nabla_{\eta_L} \widehat{\cJ}^m(\bm{\rho}_L)$ with the generated samples $D_m^{1:L}$.
The variance and squared bias at the $L$th iteration are calculated based on the estimated gradients from $M$ trials.
In this experiment, the true gradient $\nabla_{\eta_L} \cJ(\bm{\rho}_L)$ at the $L$th iteration is
approximated by the plain PGPE method using Eq.(\ref{emp_gra}) with $N=10000$ on-policy samples.
Note that the sum of the variance and squared bias
agrees with the mean squared error:
\begin{align}
  \operatorname{Var}+\operatorname{Bias}^2
=
  \frac{1}{M} \sum_{m=1}^M
\|\nabla_{\eta_L} \widehat{\cJ}^m(\bm{\rho}_L)-\nabla_{\eta_L} \cJ(\bm{\rho}_L)\|^2.
\label{bias^2+var=MSE}
\end{align}
We update the hyper-parameters $\bm{\rho}_L$ based on the estimated true gradient $\nabla_{\eta_L} \cJ(\bm{\rho}_L)$, and obtain $\bm{\rho}_{L+1}$.
Then, we investigate the variance and bias at the next iteration, i.e., the $(L+1)$th iteration, following the above procedures.
Figure~\ref{fig:var-bias-mse} shows the variance and squared bias over $20$ iterations.

\begin{figure}[t]
\centering
    \subfigure[IW-PGPE]{\includegraphics[clip,width=0.48\columnwidth]{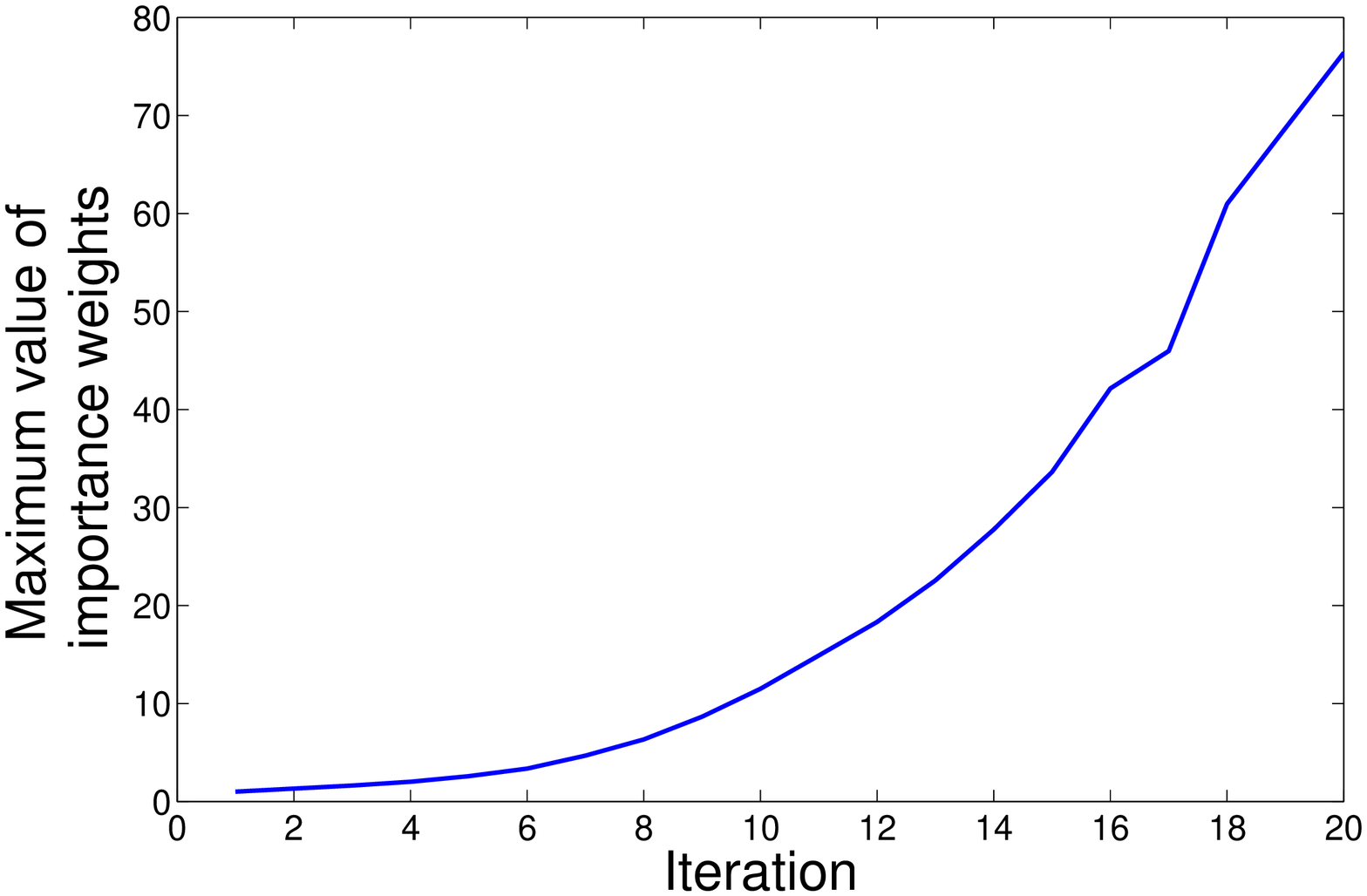}\label{fig:maxiw}}
    \subfigure[IW-PGPE\textsubscript{OB}]{\includegraphics[clip,width=0.48\columnwidth]{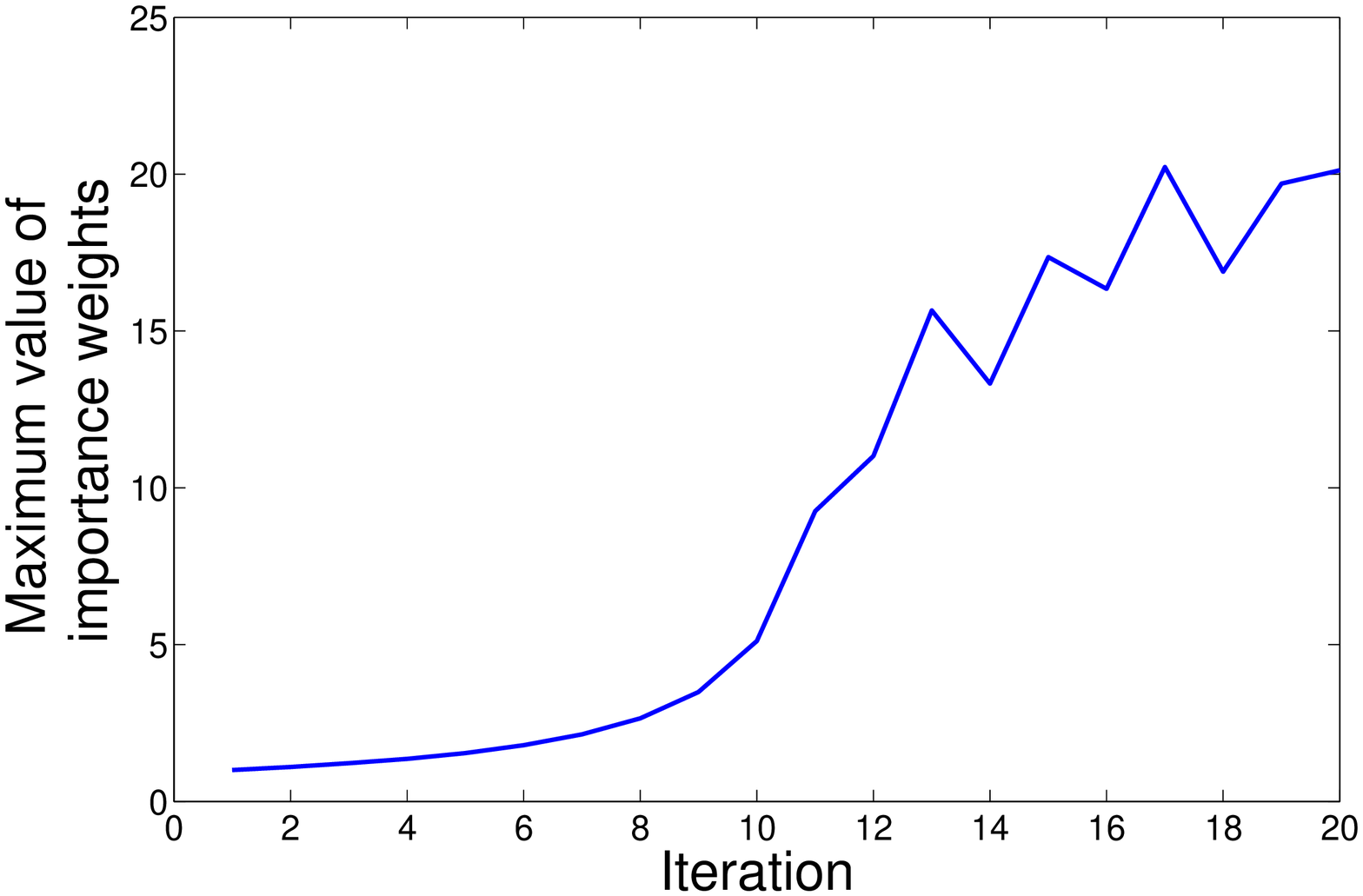}\label{fig:maxobiw}}
    \caption{Average maximum values of importance weights over $20$ runs through parameter update iterations.}
    \label{fig:maxvalue}
\end{figure}

From Figure~\ref{fig:var},
we can see that IW-PGPE\textsubscript{OB} provides gradient estimates with the lowest variance
among the compared methods.
IW-PGPE has a larger variance than NIW-PGPE, which well agrees with our theoretical analysis:
According to Theorem~\ref{theorem:variance-bound-PGPE},
upper bounds of the variance
are proportional to the importance weight,
which is always $1$ in NIW-PGPE,
but is very large in IW-PGPE if the target distribution
is significantly different from the sampling distribution.
In order to see whether the upper bound of importance weights is really large, we measure the maximum value of importance weights over iterations, which is shown in Figure~\ref{fig:maxvalue}. Figure~\ref{fig:maxiw} shows that the maximum value of importance weights tends to be larger over iterations, which further illustrates how importance weights influence the variance of gradient estimates in IW-PGPE.

We can also see that
the gap in the variance between IW-PGPE and IW-PGPE\textsubscript{OB}
tends to be larger over iterations, which is also consistent with our theoretical analysis:
According to Theorem~\ref{theorem:variance-bound-gap}, the larger the importance weight is,
the more the optimal constant baseline contributes to reducing the variance.
The importance weight may get larger at later iterations,
because distributions in the first and the last iterations
may be significantly different (Figure~\ref{fig:maxvalue} exactly illustrates this phenomenon).
Thus,  variance reduction from IW-PGPE to IW-PGPE\textsubscript{OB} by the optimal constant baseline
tends to be more significant in later iterations.
Gradient estimates in both NIW-PGPE\textsubscript{OB} and IW-PGPE\textsubscript{OB} are with smaller variance than the plain PGPE\textsubscript{OB} method,
because the more data we use, the smaller variance of gradient estimates we can obtain as expected from the theory.
IW-PGPE\textsubscript{OB} provides smaller variance than NIW-PGPE\textsubscript{OB}, which is our expected result: According to Theorem \ref{theorem:variance-bound-gap},
if the importance weights are larger, using the optimal constant baseline can reduce variance more, while the importance weights are always $1$ in NIW-PGPE\textsubscript{OB} (see Figure~\ref{fig:maxobiw}).
The plain PGPE\textsubscript{OB} has smaller variance than the plain PGPE, which well agrees with the results reported in \cite{NN2012-ting}.

Figure~\ref{fig:bias} shows that
introduction of the optimal baseline does not increase the bias.
NIW-PGPE and NIW-PGPE\textsubscript{OB} have very large bias, because naively reusing previous data
leads to an inconsistent and biased gradient estimator.
The bias of gradient estimates in IW-PGPE is fairly small, because IW-PGPE is not only consistent, but also unbiased.
The plain PGPE and plain PGPE\textsubscript{OB} are also with small bias, as expected.

Because our proposed IW-PGPE\textsubscript{OB} has small bias and the smallest variance
among the compared methods,
it also gives the smallest mean squared error
(see Eq.\eqref{bias^2+var=MSE}).

\subsubsection{Hyper-Parameter Trajectories}
\begin{figure}[p]
\centering
    \subfigure[\small{NIW-PGPE ($N=10$)}]{\includegraphics[clip,width=0.48\columnwidth]{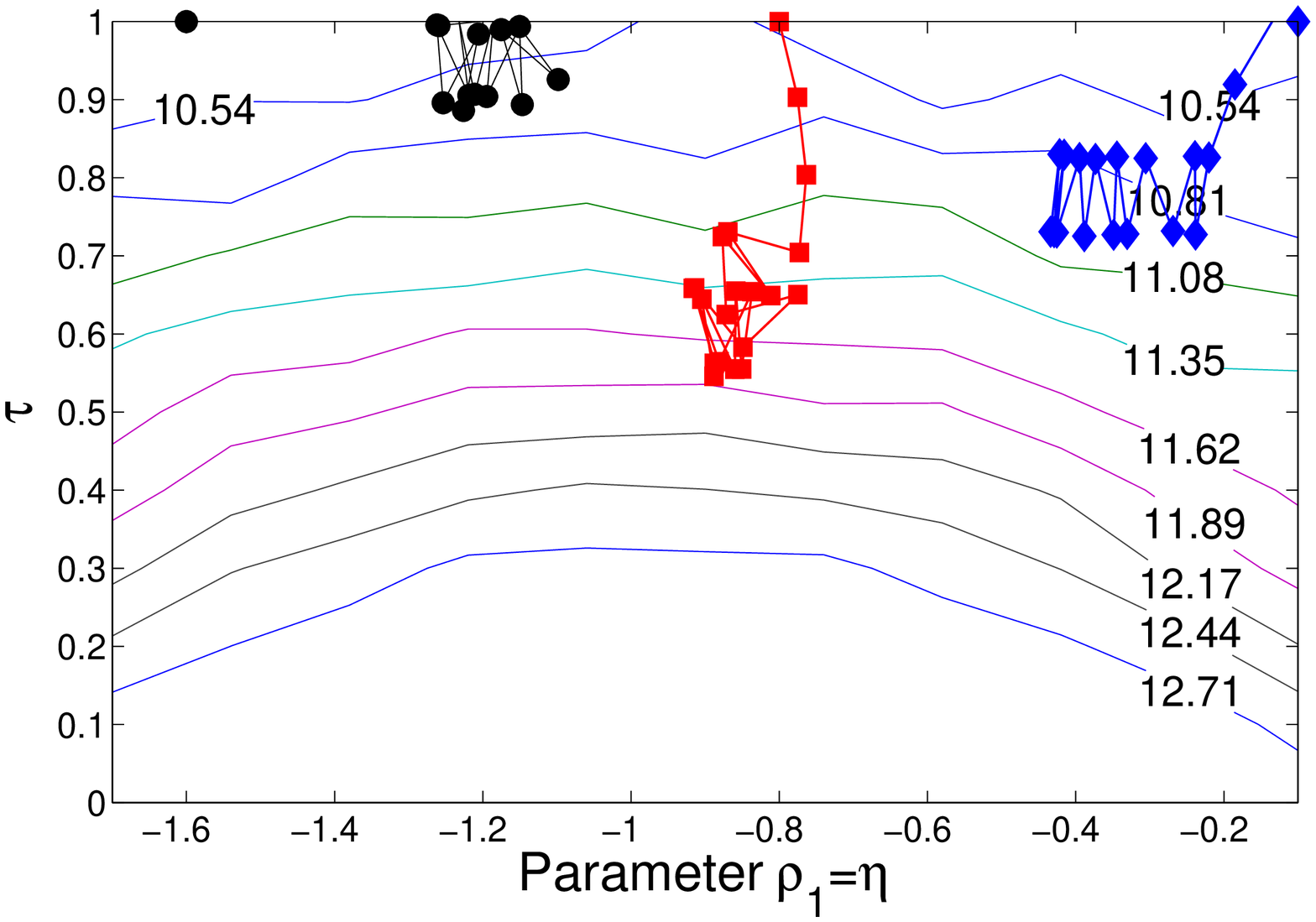}\label{fig:parniw-n10}}
    \subfigure[\small{NIW-PGPE ($N=1$)}]{\includegraphics[clip,width=0.48\columnwidth]{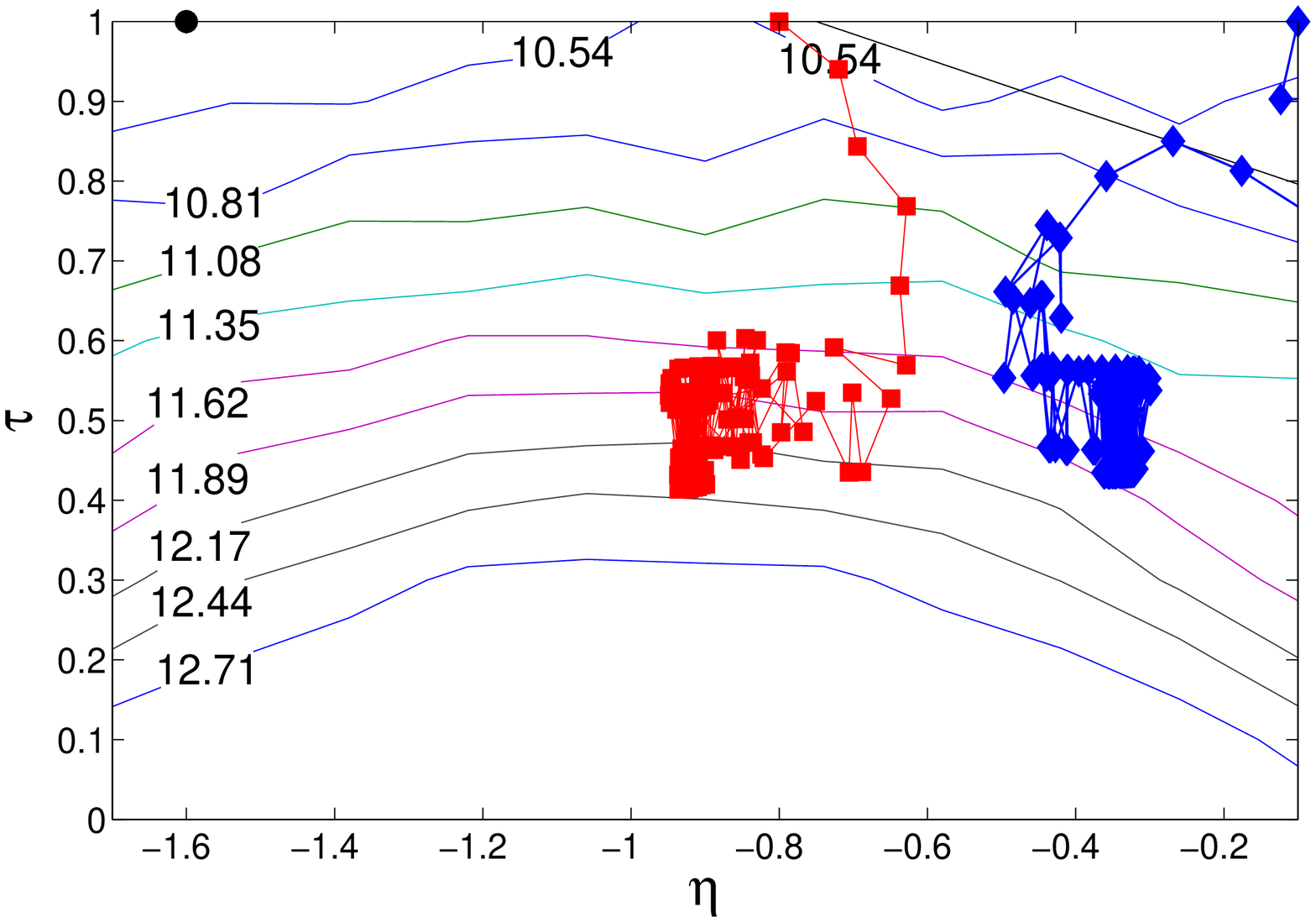}\label{fig:parniw-n1}}
    \subfigure[\small{IW-PGPE ($N=10$)}]{\includegraphics[clip,width=0.48\columnwidth]{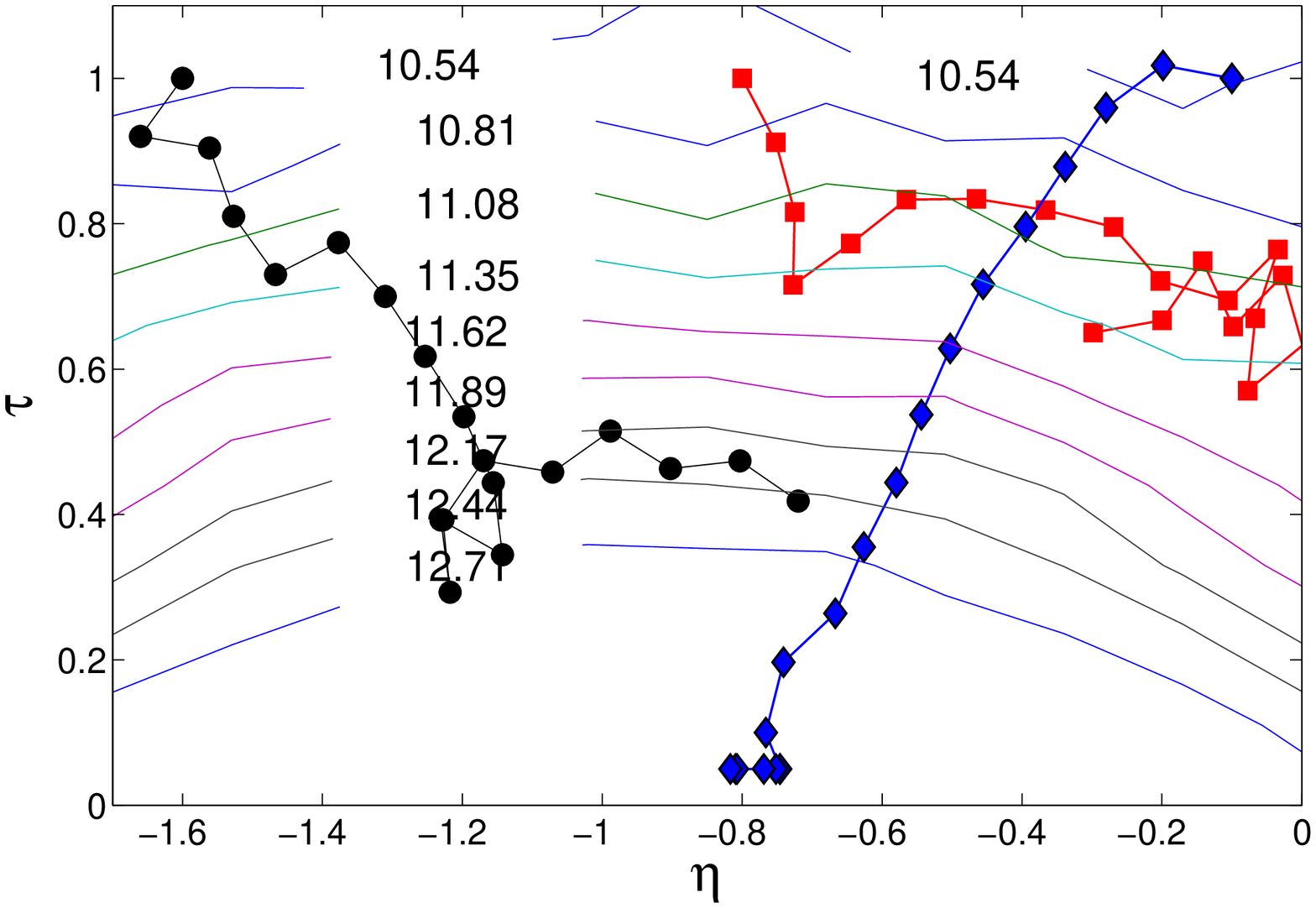}\label{fig:pariw-n10}}
    \subfigure[\small{IW-PGPE ($N=1$)}]{\includegraphics[clip,width=0.48\columnwidth]{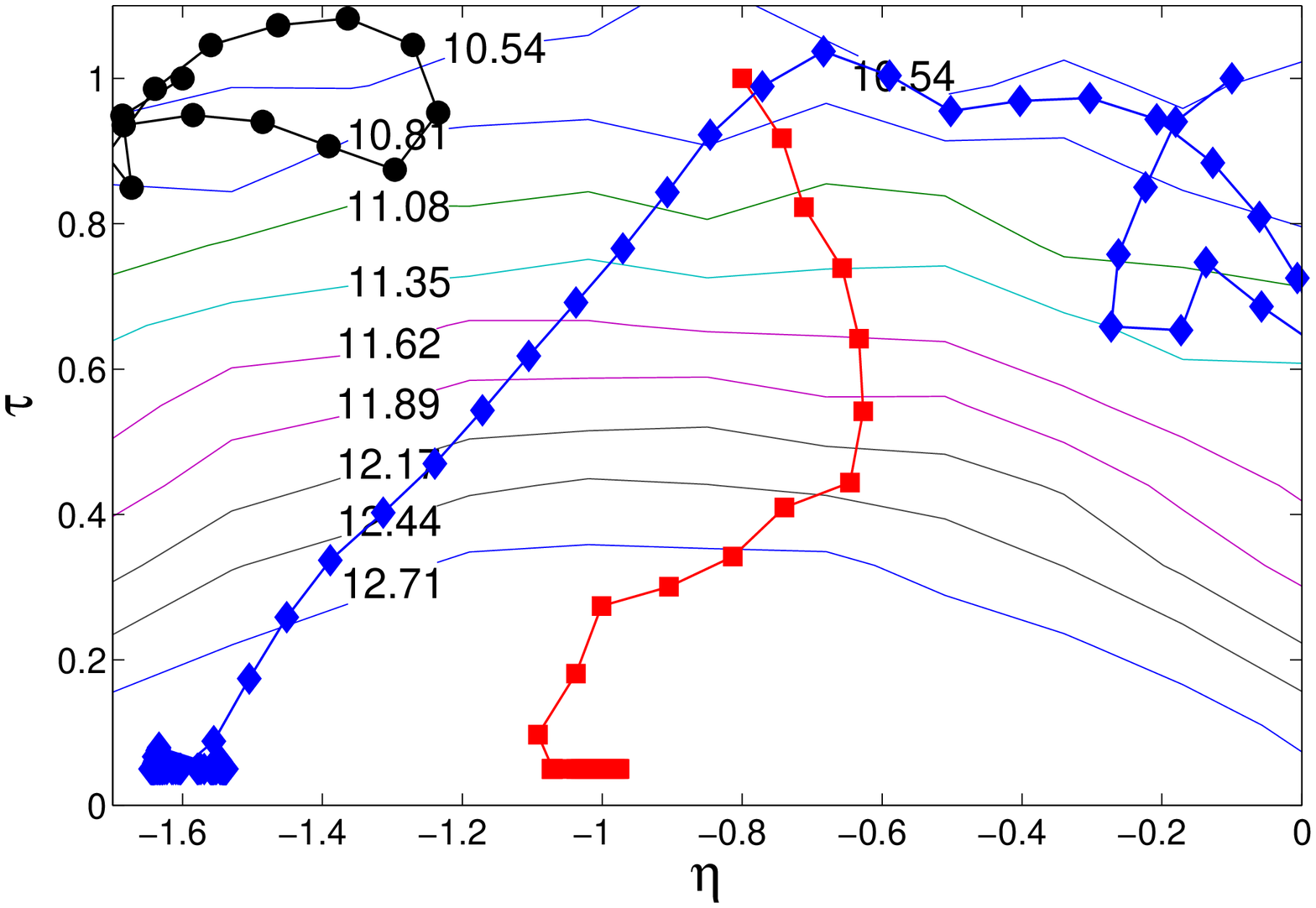}\label{fig:pariw-n1}}
    \subfigure[\small{IW-PGPE\textsubscript{OB} ($N=10$)}]{\includegraphics[clip,width=0.48\columnwidth]{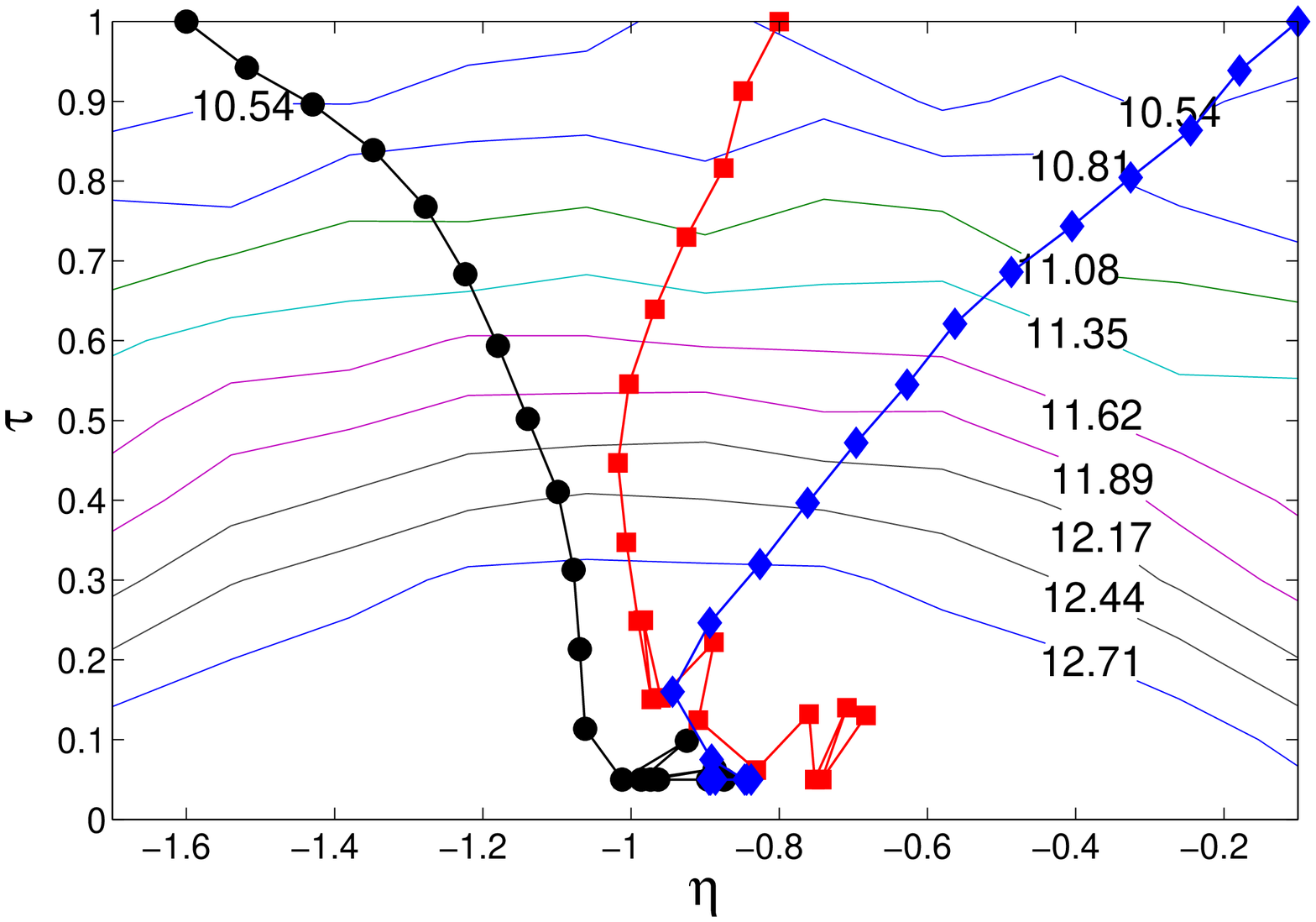}\label{fig:parobiw-n10}}
    \subfigure[\small{IW-PGPE\textsubscript{OB} ($N=1$)}]{\includegraphics[clip,width=0.48\columnwidth]{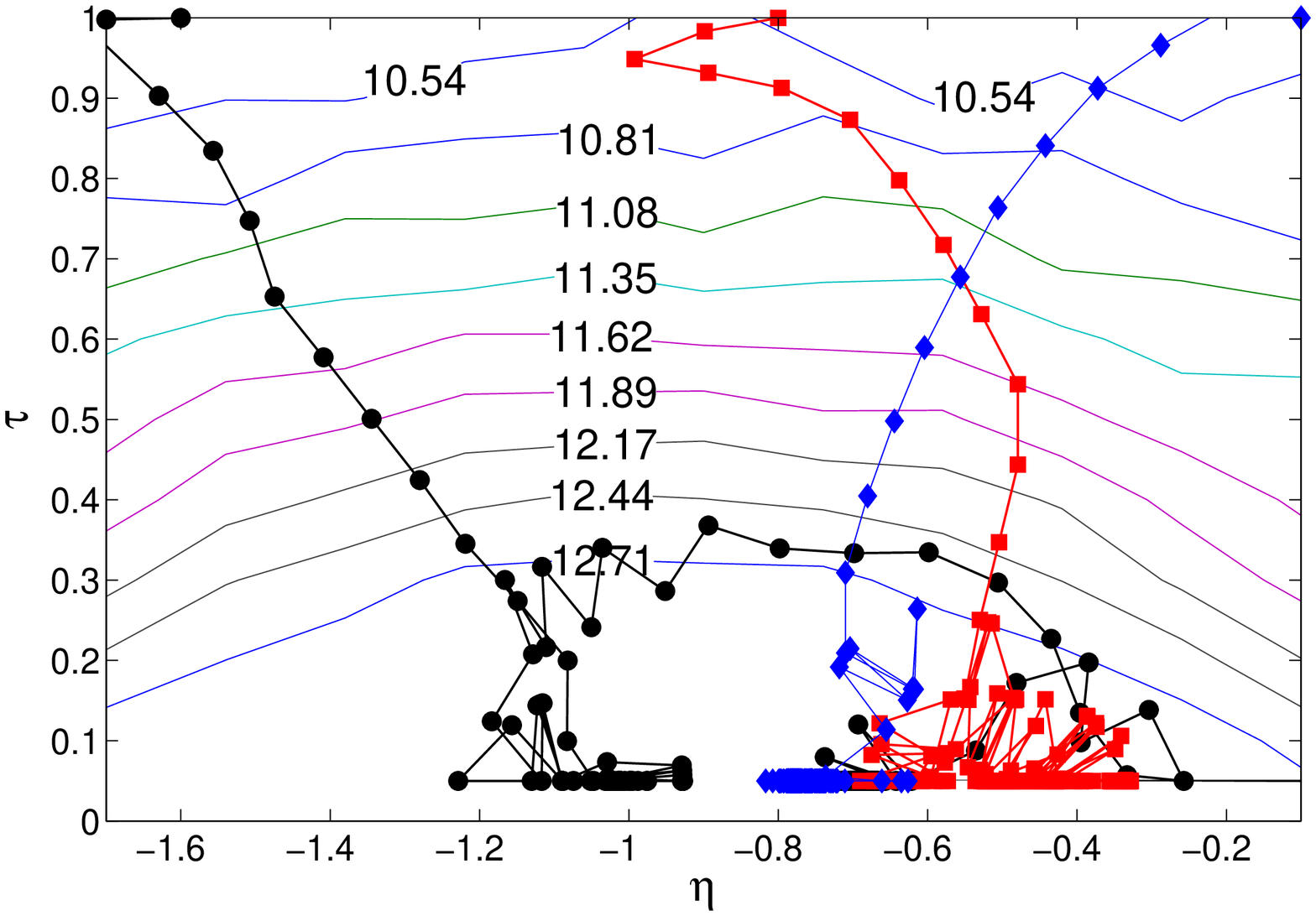}\label{fig:parobiw-n1}}
    \caption{Trajectories of policy hyper-parameters over iterations.}
    \label{fig:parchange-stable}
\end{figure}

Next, we illustrate how learned hyper-parameters change over iterations.
Here we compare the behavior of the following three methods: NIW-PGPE, IW-PGPE and our proposed method IW-PGPE\textsubscript{OB}.
We fix the initial deviation parameter at $\tau=1$,
and test the three different initial mean parameters: $\eta=-1.6$, $-0.8$, and $-0.1$.
Figure~\ref{fig:parchange-stable} depicts the contour of the expected return,
where the maximum of the return surface is located at the middle bottom.

First, let us investigate how the hyper-parameters change over 20 iterations in a large-sample case with $N=10$.
From Figure~\ref{fig:parniw-n10}, we can see that NIW-PGPE can not properly update the solutions,
which means that the inconsistency can not be overcome
by increasing the number of samples.
On the other hand, Figure~\ref{fig:pariw-n10} shows that IW-PGPE can lead
the solutions to an area with large returns sometimes, but can not always reach an area with large returns after 20 iterations.
This indicates that the consistency of importance weighting tends to be helpful
when the number of samples is large, but it can not converge rapidly because of the large variance.
Figure~\ref{fig:parobiw-n10} shows that IW-PGPE\textsubscript{OB} gives
the reliable update directions and the three paths converge rapidly to
the vicinity of the maximum point without detours.
This shows that the optimal constant baseline highly contributes to
improving the convergence property of IW-PGPE.

Next, we investigate the performance over $200$ iterations
with only $N=1$.
Figure~\ref{fig:parniw-n1} shows that NIW-PGPE can not properly update the solutions
to the maximum point because of the inconsistency,
and Figure~\ref{fig:pariw-n1} shows that
the IW-PGPE solutions can not always reach an area with large returns (middle bottom) after 200 iterations,
which is because the variance in IW-PGPE is crucial in this extreme scenario.
However, Figure~\ref{fig:parobiw-n10} shows that
the proposed IW-PGPE\textsubscript{OB} can still find fairly reliable update directions with only $N=1$.

\begin{figure}[h]
\centering
    \subfigure[NIW-PGPE ]{\includegraphics[clip,width=0.48\columnwidth]{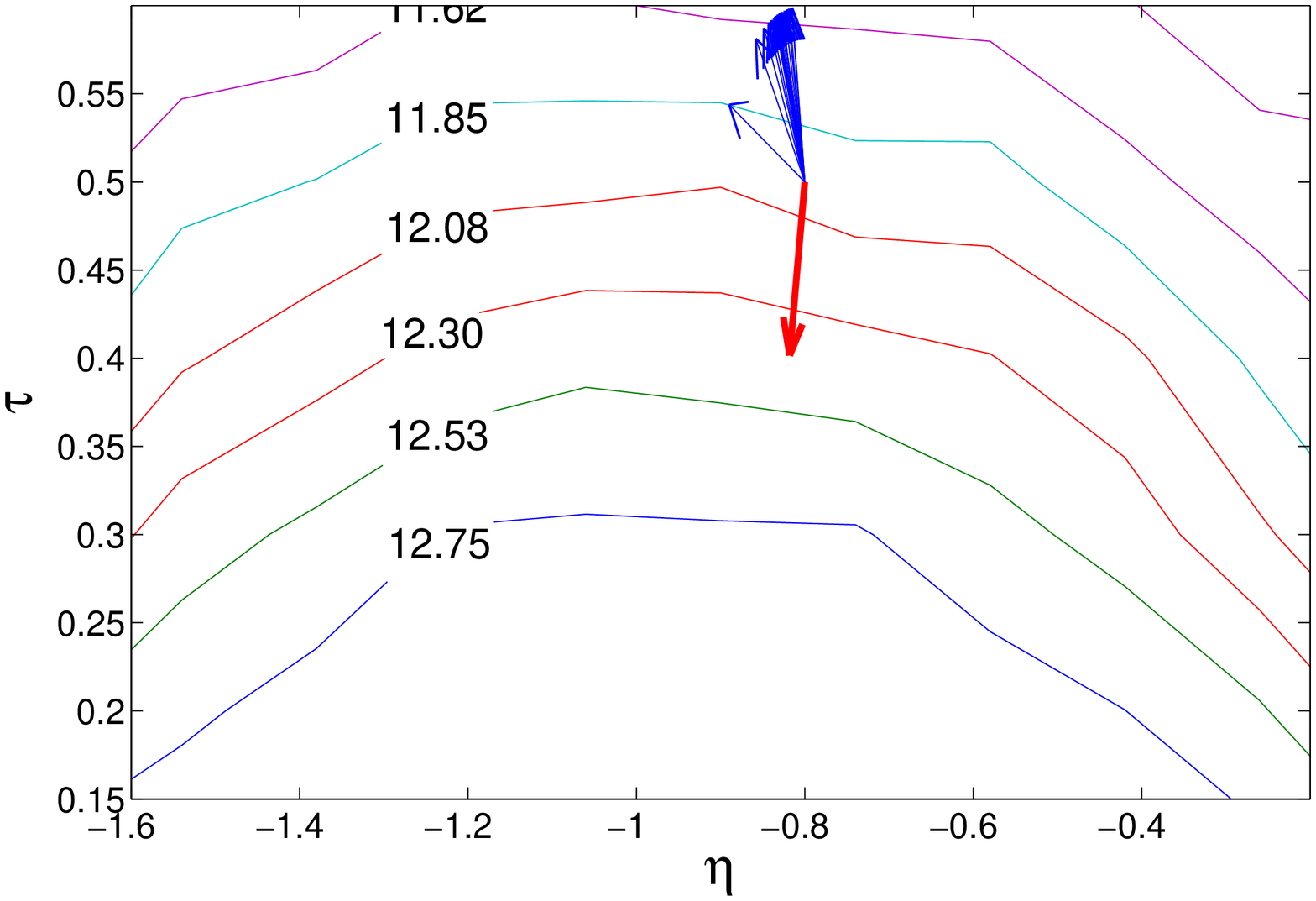}\label{fig:one-niw}}
    \subfigure[NIW-PGPE ]{\includegraphics[clip,width=0.48\columnwidth]{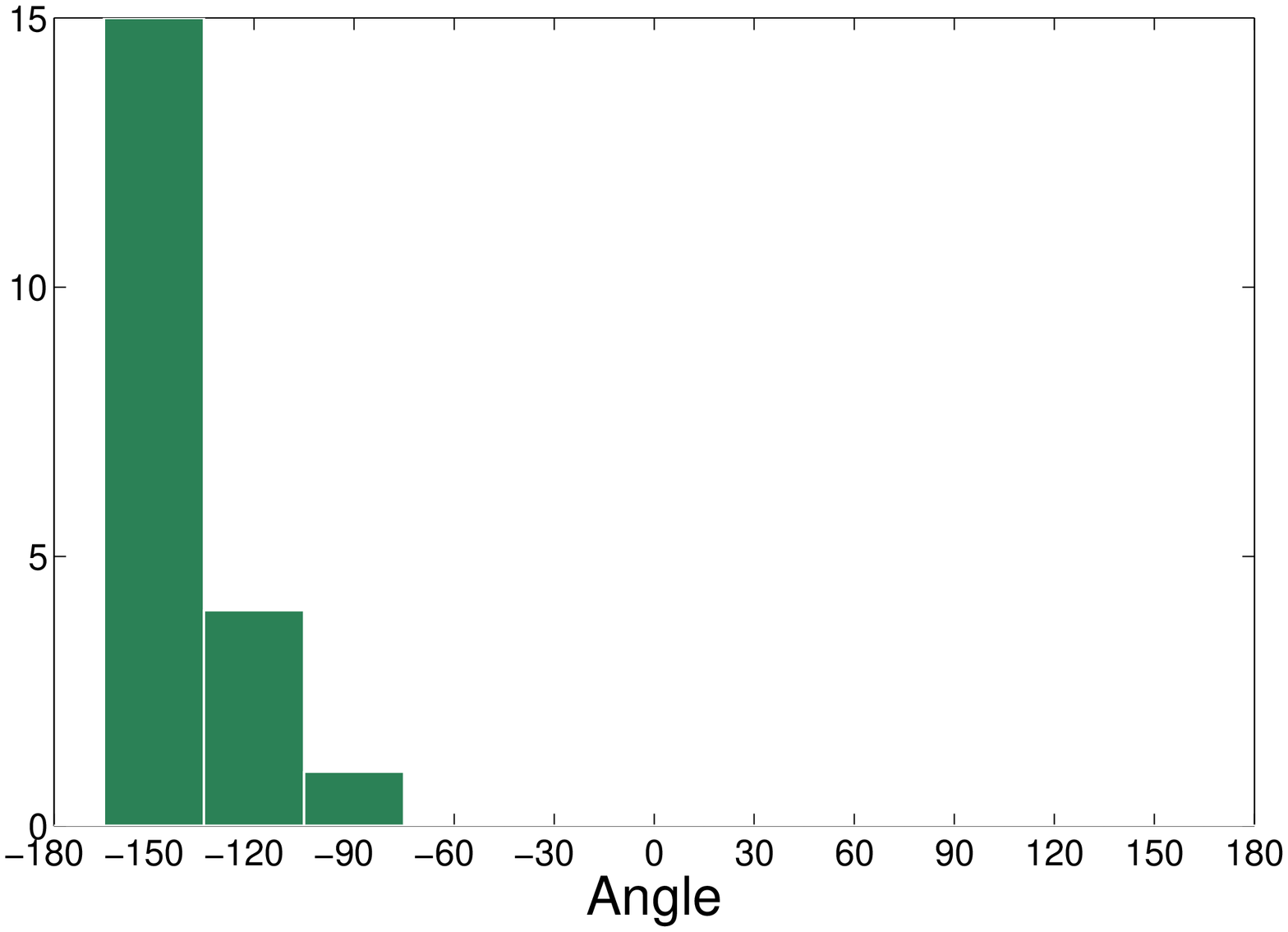}\label{fig:hist-niw}}
    \subfigure[IW-PGPE]{\includegraphics[clip,width=0.48\columnwidth]{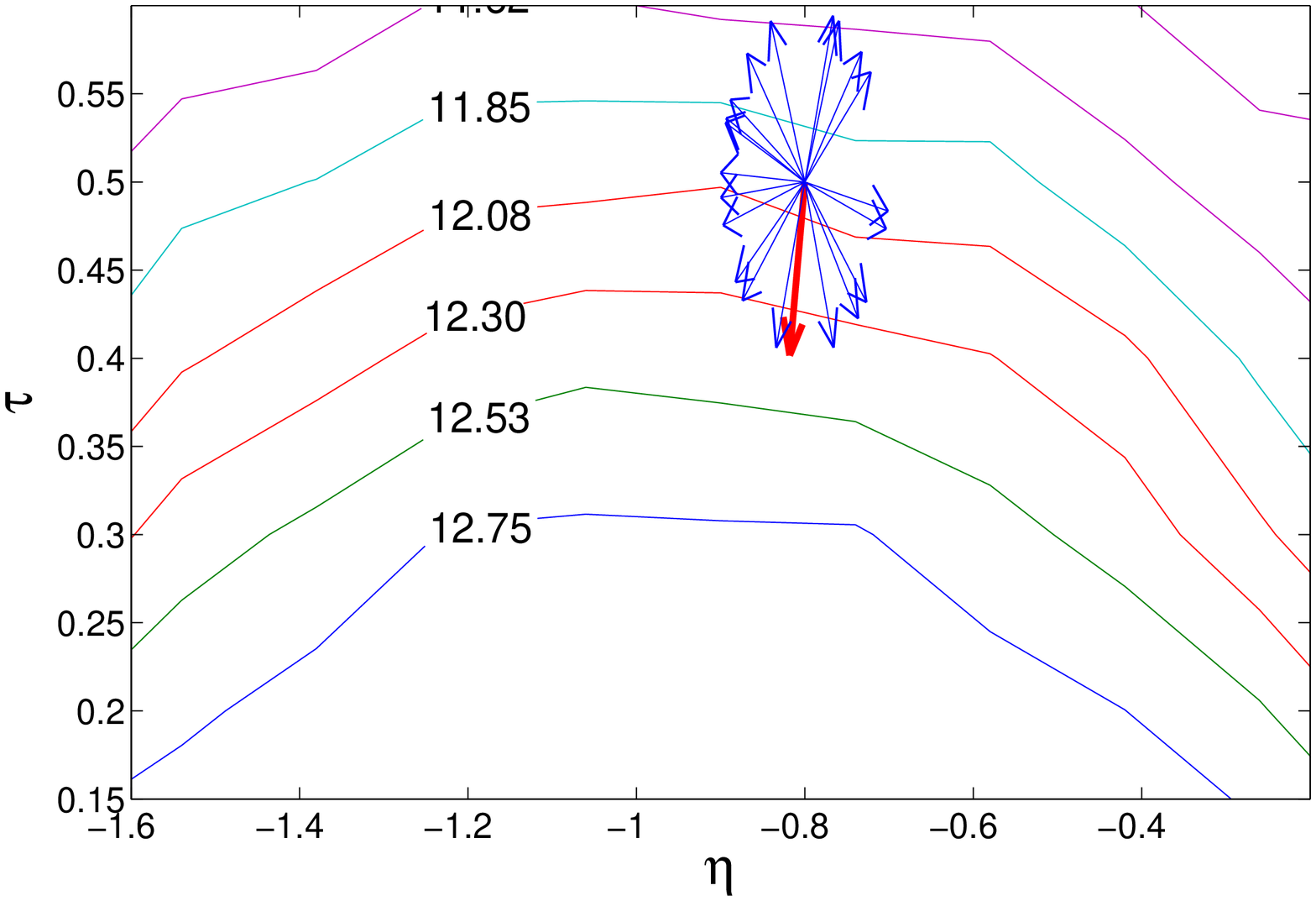}\label{fig:one-iw}}
    \subfigure[IW-PGPE]{\includegraphics[clip,width=0.48\columnwidth]{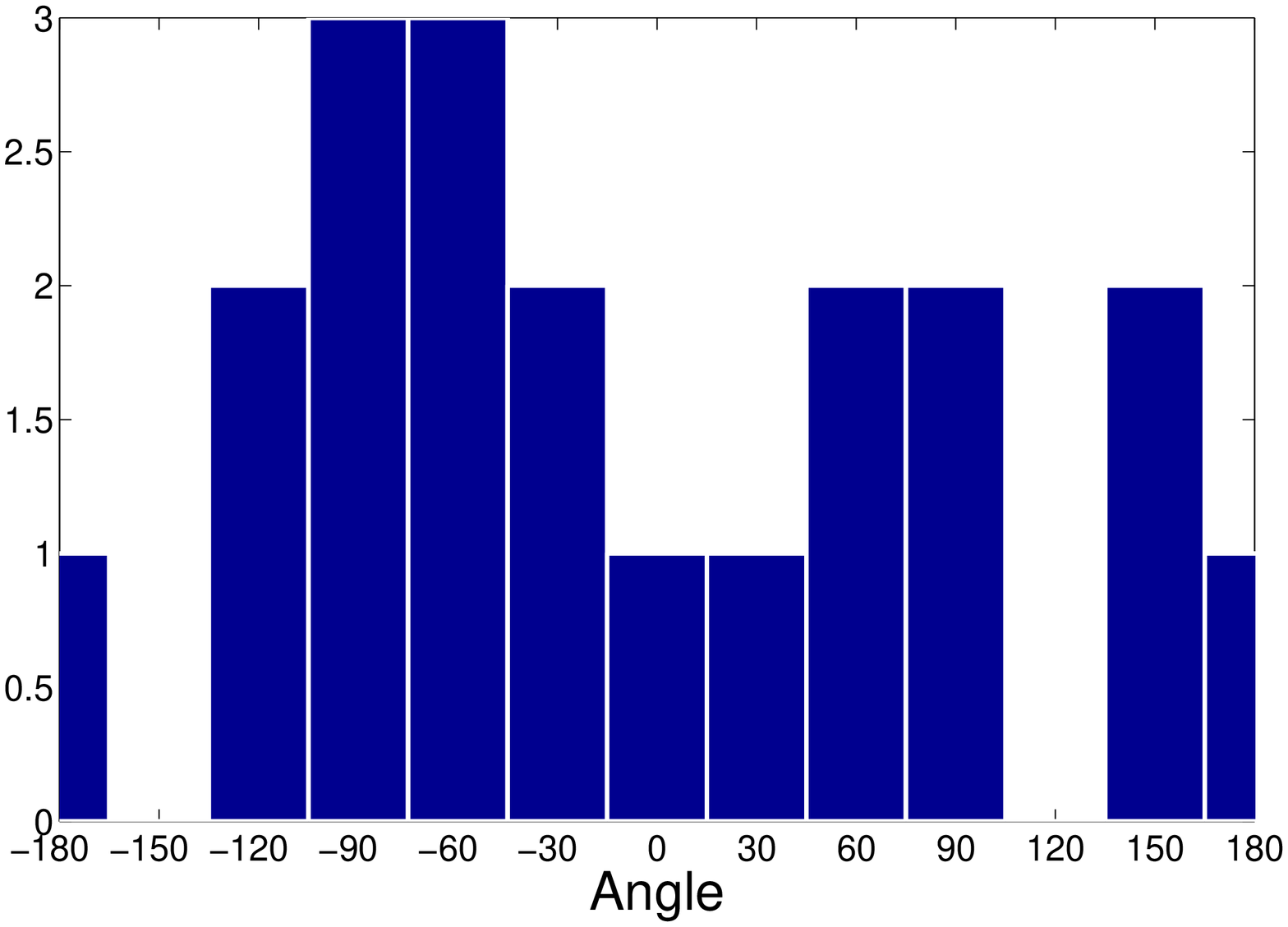}\label{fig:hist-iw}}
    \subfigure[IW-PGPE\textsubscript{OB}]{\includegraphics[clip,width=0.48\columnwidth]{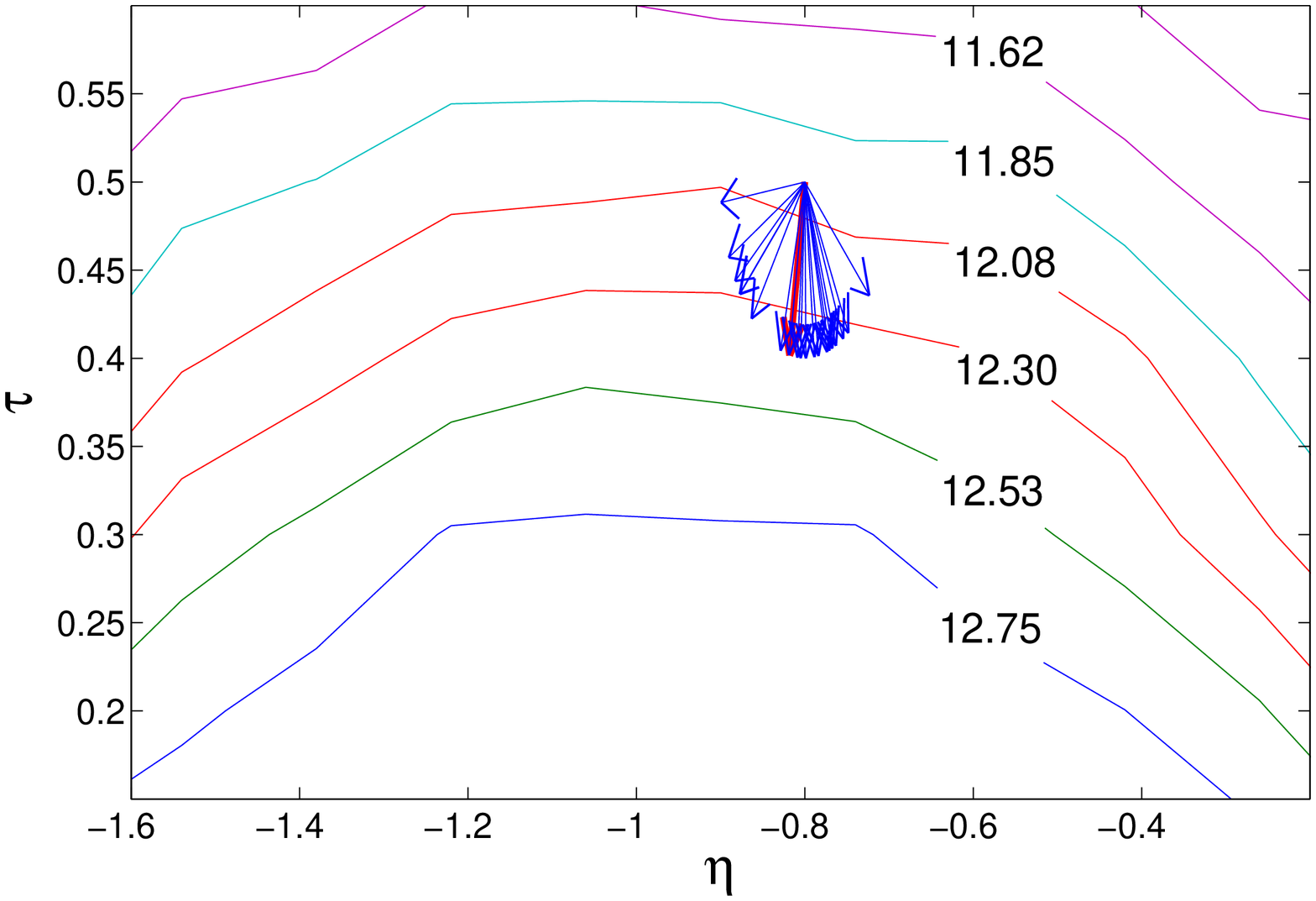}\label{fig:one-obiw}}
    \subfigure[IW-PGPE\textsubscript{OB}]{\includegraphics[clip,width=0.48\columnwidth]{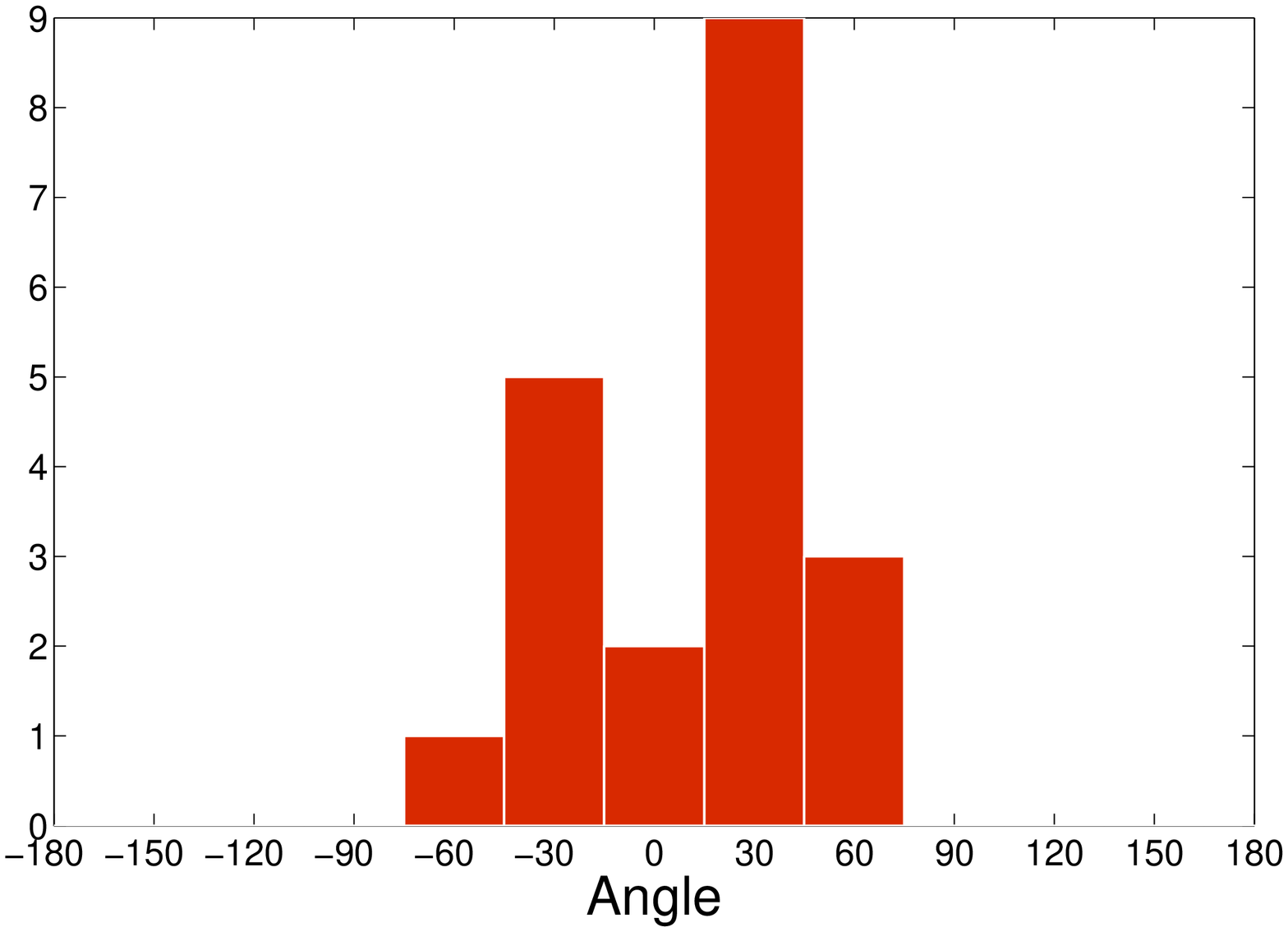}\label{fig:hist-obiw}}
    \caption{Directions of estimated gradients.}
    \label{fig:one-change}
\end{figure}

Next, we investigate the directions of estimated gradients more systematically.
We fix the starting point at $\eta=-0.8$ and $\tau=0.5$.
The true gradient direction is calculated by the plain PGPE method with 10000 on-policy samples.
In this experiment, we first collect $N'=10$ off-policy samples, which are drawn from $\cN(-1.6,1)$.
We then reuse these off-policy samples to estimate the gradients in the data-reuse methods.
We calculate the gradients 20 times with different random seeds, and investigate the angle between the true gradient and the estimated gradients.
The results are summarized in Figure~\ref{fig:one-change}.
In Figure~\ref{fig:one-niw}, the red line denotes the true gradient and blue lines are the estimated gradients by the NIW-PGPE method.
The histograms of angles between the true gradient and the estimated gradients are plotted in Figure~\ref{fig:hist-niw}.
The graph shows that the angles are concentrated in $[-150,-90]$, which further explains the inconsistent property of the NIW-PGPE method.
Observing the angle distribution for IW-PGPE in Figure~\ref{fig:hist-iw}, we can see that the angles are widely distributed in $[-180, 180]$, which clearly illustrates the large variance problem of IW-PGPE.
On the other hand, the angles for the IW-PGPE\textsubscript{OB} method are concentrated in $[-60, 60]$, which highlights the small variance and consistent properties of IW-PGPE\textsubscript{OB}.

\subsubsection{Performance of Learned Policies}
Finally, we evaluate average expected returns obtained by each method over $20$ runs.
The expected return at each trial is approximated using $100$ newly-drawn test episodic data (which are not used for policy learning).
The initial mean parameter $\eta$ is chosen randomly from the standard normal distribution,
and the deviation parameter is fixed at $\tau=1$.

\begin{figure}[t]
\centering
    \includegraphics[width=0.7\columnwidth]{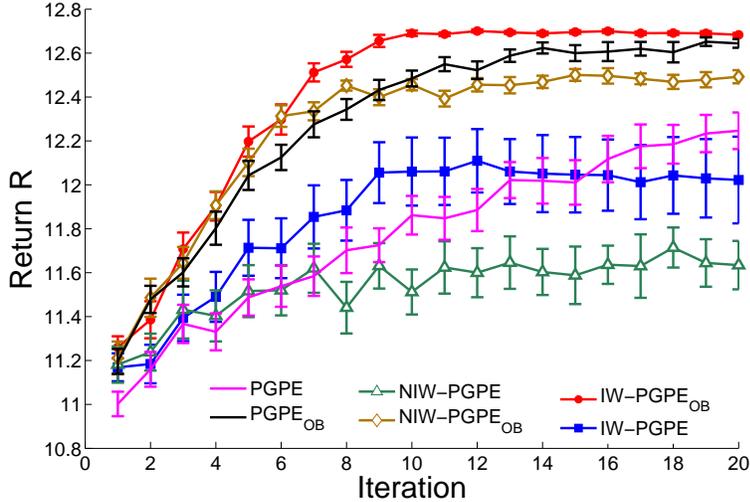}
    \caption{Average expected returns through policy update iterations over $20$ runs for toy data. Error bars denote standard errors.}
    \label{fig:performance-n10}
\end{figure}

Figure~\ref{fig:performance-n10} shows that IW-PGPE\textsubscript{OB}
improves the performance over iterations and converges very fast.
The performance of NIW-PGPE is not largely improved over iterations,
which is caused by biased gradient estimates (see Figure~\ref{fig:parniw-n10} again).
IW-PGPE works better than NIW-PGPE, but the performance is saturated after $9$ iterations.
IW-PGPE\textsubscript{OB} does not outperform NIW-PGPE\textsubscript{OB} that much at the first several iterations,
because the difference between the target distribution and a sampling distribution is not that large at the beginning.
However, the upper bound of importance weights tends to become larger over iterations (see Figure~\ref{fig:maxobiw} again),
which makes IW-PGPE\textsubscript{OB} more reliable than NIW-PGPE\textsubscript{OB} in the latter iterations.
The plain PGPE\textsubscript{OB} method works fairly well with $N=10$ on-policy samples,
but it is still not as good as IW-PGPE\textsubscript{OB}.

\subsection{Mountain Car}
\begin{figure}[t]
\centering
\includegraphics[clip,width=0.5\columnwidth]{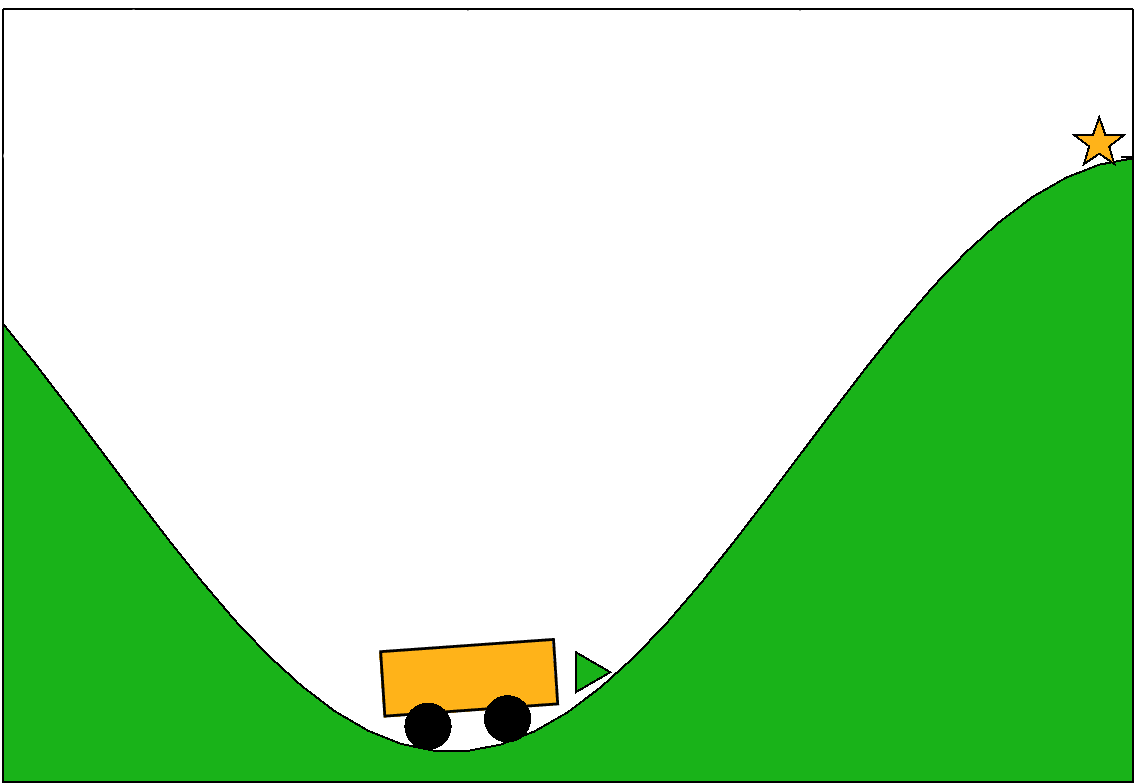}
\caption{Mountain car.}
\label{fig:car}
\vspace*{5mm}
    \includegraphics[clip,width=0.7\columnwidth]{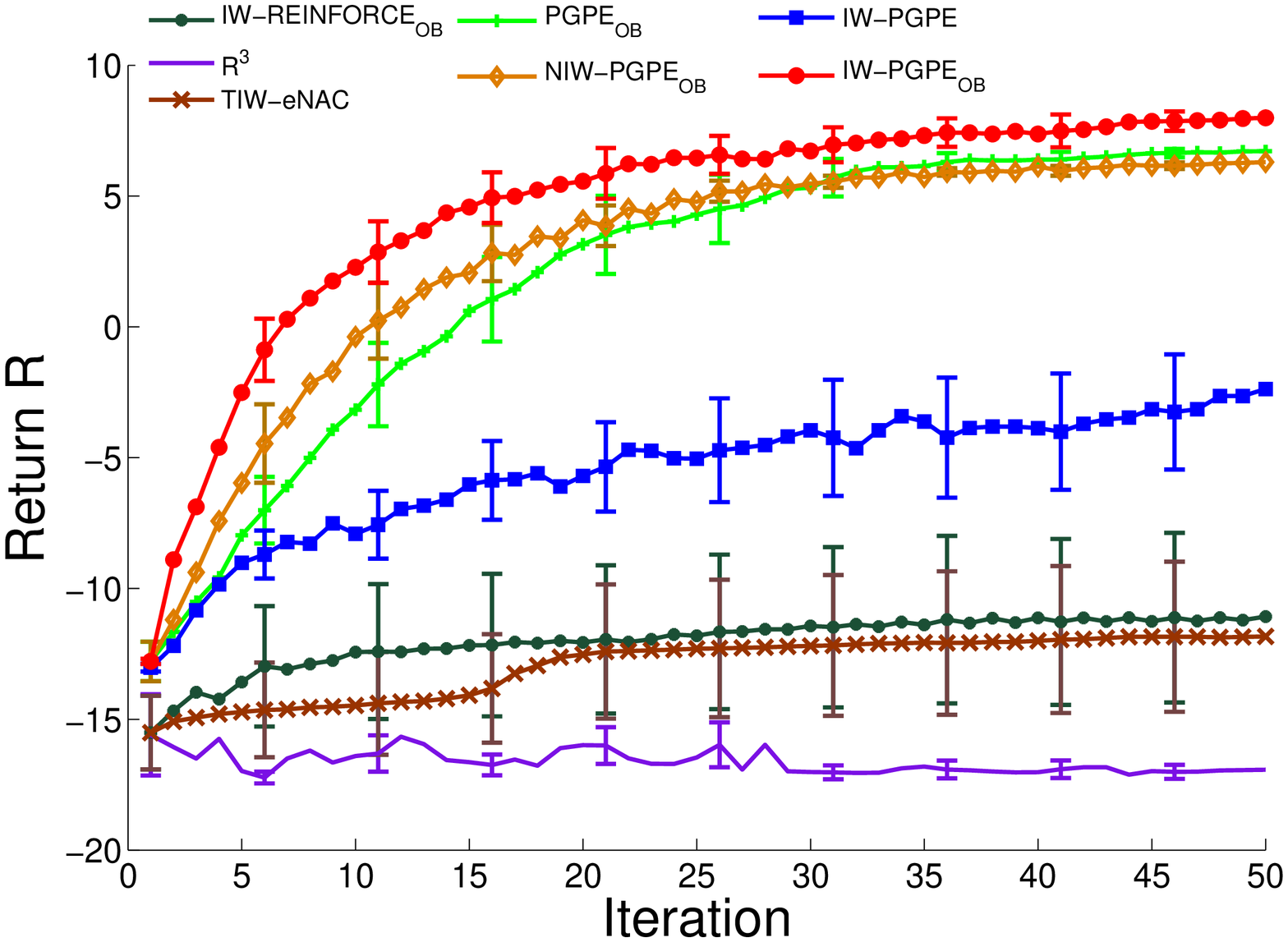}
    \caption{Average expected returns over $10$ runs as functions of the number of iterations for the mountain-car task. Error bars are standard errors.}
    \label{fig:mountain}
    \end{figure}

 Next, we evaluate our proposed method in the \emph{mountain car} task,
which is illustrated in Figure~\ref{fig:car}.
The task consists of a car and two hills whose landscape is described as $\sin(3x)$.
The top of the right hill is the goal to which we want to guide the car.

    We compare the following $7$ methods:
\begin{itemize}
  \item \textbf{TIW-eNAC:}
    Truncated importance-weight episodic natural actor-critic, which
    is an episodic version of the sample-reuse NAC method \cite{truncated, Peters:2008}. Following the same line as \cite{truncated}, we truncate the importance weight as $w=\min\{w,2\}$.
  \item \textbf{IW-REINFORCE\textsubscript{OB}:}
    Importance-weighted REINFORCE with the optimal baseline, which is basically a combination of
    the off-policy implementation of the episodic REINFORCE method \cite{MPK2001} and
    the optimal baseline \cite{IROS:Peters+Schaal:2006},
    although we could not exactly find this method in literature.
  \item \textbf{R\textsuperscript{3}:}
    Reward-weighted regression with sample reuse \cite{R3}.
   \item \textbf{PGPE\textsubscript{OB}:}
    Plain PGPE\textsubscript{OB} without data reuse.
  \item \textbf{NIW-PGPE\textsubscript{OB}:}
    Data-reuse PGPE\textsubscript{OB} without importance weighting.
   \item \textbf{IW-PGPE:}
    Importance-weighted PGPE.
   \item \textbf{IW-PGPE\textsubscript{OB}:}
    Importance-weighted PGPE with the optimal baseline.
  \end{itemize}

The state space ${\cal S}$ is two-dimensional and continuous,
which consists of the horizontal position $x[m]\in[-1.2, 0.5]$
and the velocity $\dot{x}[m/s]\in[-1.5, 1.5]$, i.e.,
$\bm{s}=(x,\dot{x})^\T$.
This is non-linearly transformed
to a feature space via a basis function vector $\bm{\phi}(\bm{s})$.
We use $12$ Gaussian kernels with mean $\bm{c}$
and standard deviation $\kappa=1$ as the basis functions,
 \[\bm{\phi}(\bm{s})=\exp\left(-\frac{\|\bm{s}-\bm{c}\|^2}{2\kappa^2}\right),\]
where the kernel centers $\bm{c}$ are distributed over the following grid points:
\[\{-1.2,-0.35,0.5\}\times\{-1.5,-0.5,0.5,1.5\}.\]

The action space ${\cal A}$ is one-dimensional and continuous,
which corresponds to the force applied to the car
(note that the force of the car is not strong enough to climb up the slope to directly reach the goal).
We use the Gaussian policy model for IW-REINFORCE\textsubscript{OB}, TIW-eNAC,
and R\textsuperscript{3}:
\begin{align}
  p(a|\bm{s},\bm{\theta})=\frac{1}{\sigma \sqrt{2\pi}}
    \exp\left(-\frac{(a-\bm{\mu}^\T \bm{\phi}\left(\bm{s}\right))^2}{2\sigma^2}\right),
\label{Gaussian-policy-model}
\end{align}
where $\bm{\mu}$ is the mean policy parameter
and $\sigma$ is the deviation policy parameter.
We employ a linear deterministic policy model \eqref{linear-policy-model} for the PGPE methods,
which corresponds to Eq.\eqref{Gaussian-policy-model} with $\sigma\to0$.

The dynamics of the car
(i.e., the update rules of the position and the velocity) are given by
    \begin{align*}
      x_{t+1}&=x_t+\dot{x}_{t+1}\Delta t,\\
      \dot{x}_{t+1}&=\dot{x}_t+(-9.8\emph{w}\cos(3x_t)+\frac{a_t}{\emph{w}}-\emph{k}\dot{x}_t)\Delta t,
    \end{align*}
where $a_t$ is the action taken at time $t$.
We set the problem parameters as follows:
The mass of the car $\emph{w}=0.2$[kg],
the friction coefficient $\emph{k}=0.3$,
and the simulation time step  $\Delta t=0.1$[s].
The reward function is defined as
\begin{align*}
r(\bm{s}_t,a_t,\bm{s}_{t+1}) =
\begin{cases}
1& \mathrm{if} ~~x_{t+1}\geq0.45,\\
-1& \mathrm{otherwise}.\\
  \end{cases}
\end{align*}

The initial mean parameter $\bm{\eta}$
is chosen randomly from the standard normal distribution,
and the initial deviation parameter is set at $\tau=1$.
The initial state of the car is set at the bottom of the mountain with the velocity $\dot{x}=0$.
The agent collects $N=10$ episodic samples with trajectory length $T=40$ at each iteration.
In the data reuse methods, we reuses all previous data at later iterations.
In the plain PGPE\textsubscript{OB} method, we just use $N=10$ on-policy samples at each iteration to estimate policy gradients.
The discount factor is set at $\gamma=0.95$. The learning rate is
$\varepsilon=1/\|\nabla_{\rho} \hat{\cJ}(\bm{\rho})\|$.

We investigate average expected returns over $10$ trials
as functions of policy-update iterations.
The expected return at each trial is computed over $100$ newly-drawn test episodic samples
(which are not used for policy learning).
The experimental results are plotted in Figure~\ref{fig:mountain}.
This shows that IW-PGPE\textsubscript{OB} improves the performance very fast over policy-update iterations,
and it achieves superior performance improvement than all other methods.
IW-PGPE can also improve the performance over iterations well, implying that the consistency of the IW estimator is useful in this task.
However, it is outperformed by the proposed IW-PGPE\textsubscript{OB},
perhaps because the estimation variance in IW-PGPE is large.
NIW-PGPE\textsubscript{OB} performs fairly well, which maybe because the bias of policy gradient estimators is not that crucial in this experiment.
The plain PGPE\textsubscript{OB} can improve the performance throughout the iterations,
which indicates that $N=10$ on-policy samples is enough for this mountain-car task.
Other data-reuse methods can improve the performance over iterations, but slowly,
and they are outperformed by the compared PGPE methods.
IW-REINFORCE\textsubscript{OB} outperforms TIW-eNAC, which maybe because the optimal constant baseline contributes significantly in IW-REINFORCE\textsubscript{OB} and
truncating the importance weights can lead to a larger bias over iterations in TIW-eNAC.
R\textsuperscript{3} can not improve the performance over iterations.
Overall, thanks to the low variance,
IW-PGPE\textsubscript{OB} achieves smooth and fast policy improvement throughout iterations,
and its performance is the best among the compared methods.

\subsection{Upper-body Humanoid Control}

\begin{figure}[t]
\centering
    \subfigure[CB-i]{\fcolorbox{black}{yellow}{\includegraphics[clip,height=15em]{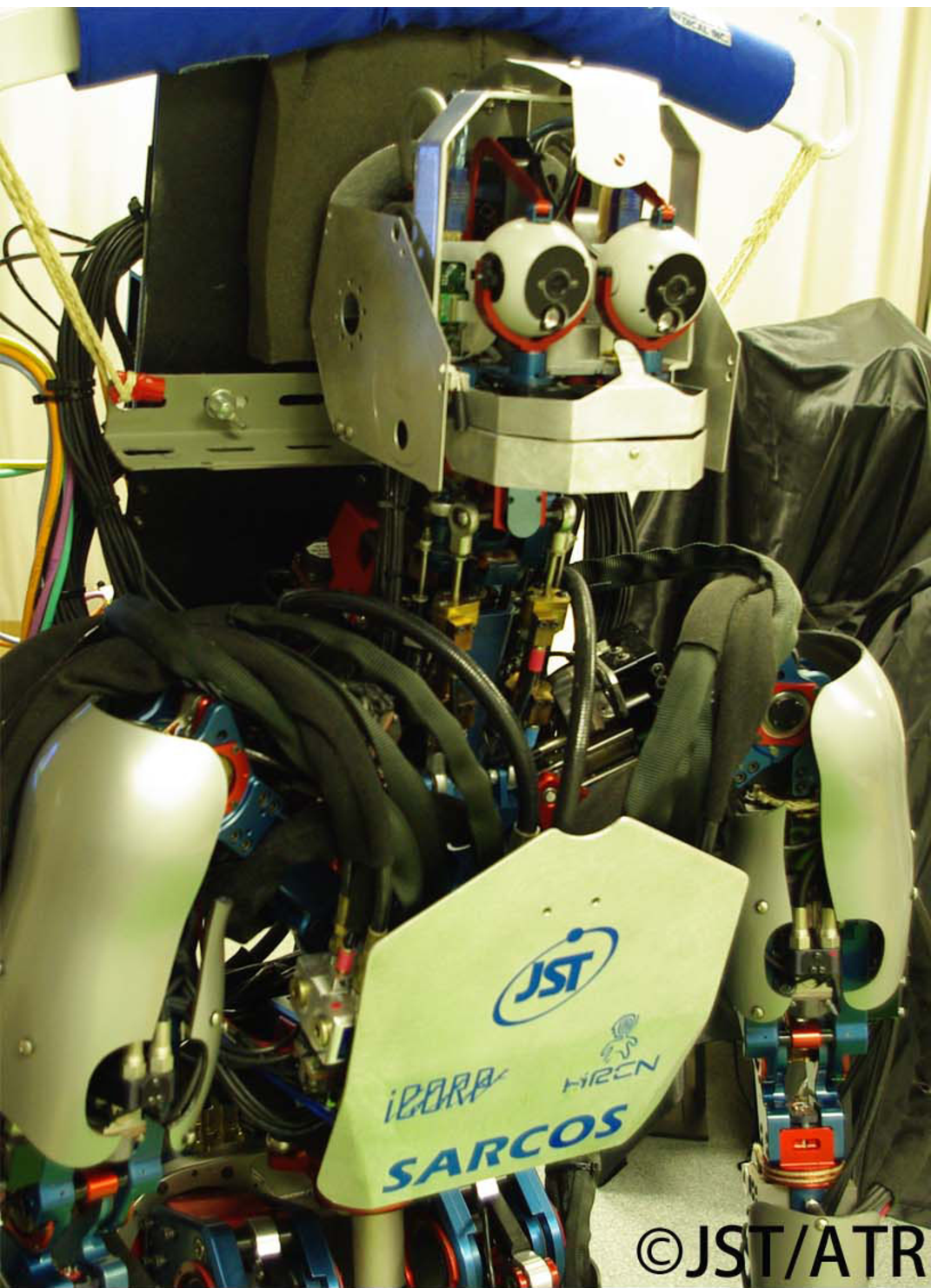}\label{fig:cbi}}}
    ~~
    \subfigure[Simulated upper-body model]{\fcolorbox{black}{yellow}{\includegraphics[clip,height=15em]{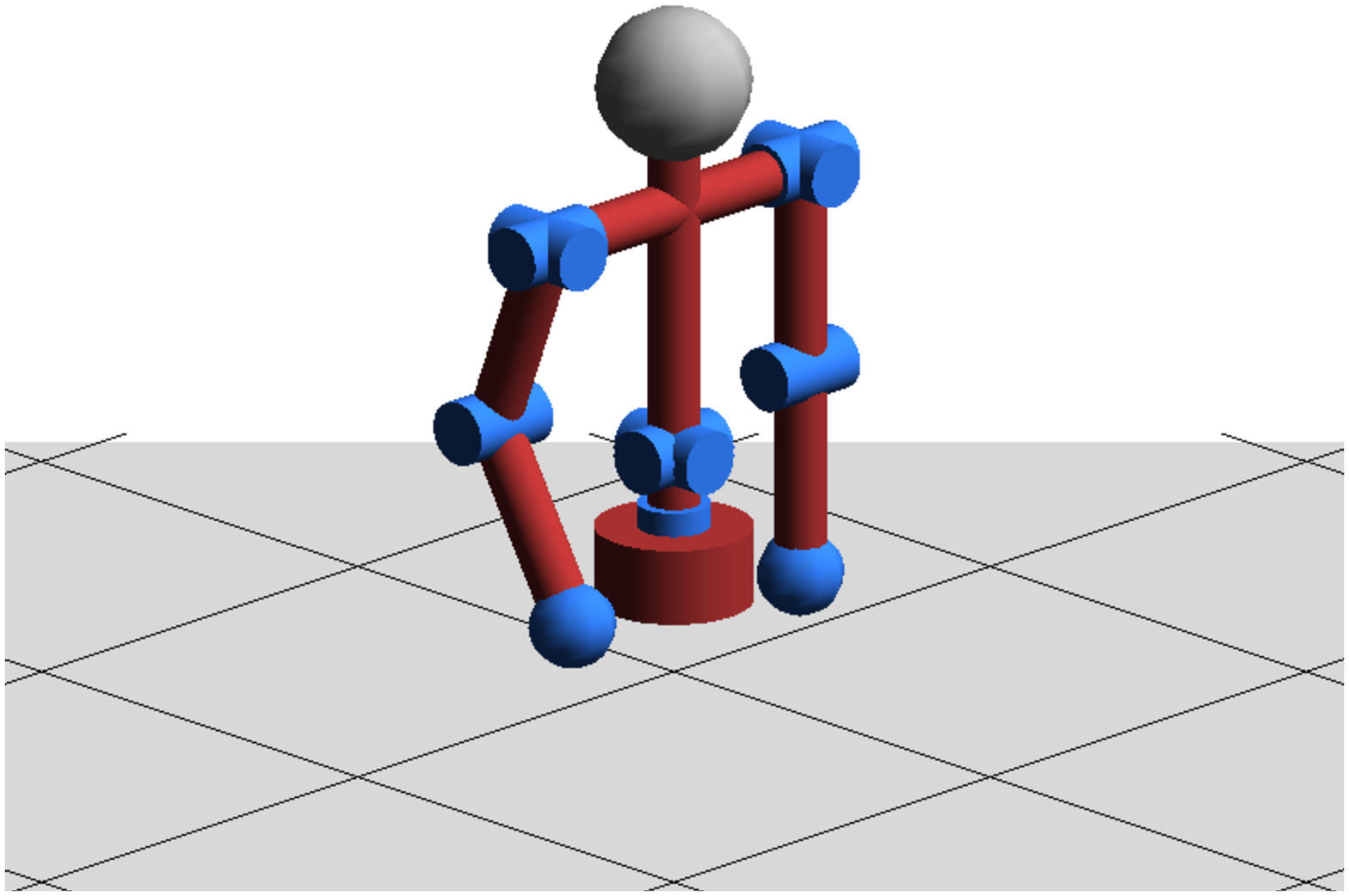}\label{fig:simulator}}}
    \caption{Humanoid robot CB-i and its upper-body model.}
    \label{fig:CBI}
\end{figure}

Finally, we evaluate the performance of our proposed method on a highly nonlinear dynamic control problem of the simulated upper-body model
of the humanoid robot \emph{CB-i} \cite{cbi}
(see Figure \ref{fig:cbi}).
We use its simulator in our experiments (see Figure \ref{fig:simulator}).
The goal is to lead the end-effector of the right arm (right hand) to a target object.

\subsubsection{Setup}
We compare the performance of the following $4$ methods:
\begin{itemize}
  \item \textbf{IW-REINFORCE\textsubscript{OB}:}
    Importance-weighted REINFORCE with the optimal baseline.
  \item \textbf{NIW-PGPE\textsubscript{OB}:}
    Data-reuse PGPE\textsubscript{OB} without importance weighting.
   \item \textbf{PGPE\textsubscript{OB}:}
    Plain PGPE\textsubscript{OB} without data reuse.
   \item \textbf{IW-PGPE\textsubscript{OB}:}
    Importance-weighted PGPE with the optimal baseline.
  \end{itemize}

The simulation is based on the upper body of the CB-i humanoid robot illustrated in Figure \ref{fig:simulator},
which has 9 degrees of freedom corresponding to main joints of the upper body:
The shoulder pitch, shoulder roll, elbow pitch of the right arm, shoulder pitch,
shoulder roll, elbow pitch of the left arm,
waist yaw, torso roll, and torso pitch.

At each time step, the controller receives states from the system and sends out actions.
The state space is 18-dimensional, which corresponds to the current angle and the current angular velocity of each joint.
The action space is 9-dimensional, which corresponds to the target angle of each joint.
Both states and actions are continuous.

The initial positions of the robot and an object are fixed, where
the initial position of the robot is set at the state of standing up straight with the arms down,
and the position of the target object depends on the task.
Note that the position of the target object is only used in the designing of the reward function.
The reward function is given by
\[r_t=k_1\exp(-10 d_t)-k_2 \min\{c_t,10000\},\]
where $k_1=1$, $k_2=0.0005$, $d_t$ is the distance between the robot's right hand and
the target object at the time step $t$, and $c_t$ is the sum of control costs for each joint.
Note that the results may change with different $k_1$ and $k_2$ for the reward function.
In order to keep the value of $\exp(-10 d_t)$ and $c_t$ in the reward function to the same order of magnitude, we need to choose $k_1$ and $k_2$ reasonably.
We use the same policy model as the mountain car experiment, i.e.,
the linear deterministic policy for PGPE and the Gaussian policy for IW-REINFORCE\textsubscript{OB}
with the basis function $\phi (\bm{s})=\bm{s}$.

The initial mean parameter $\eta$ is randomly chosen from the standard normal distribution, and the initial standard deviation parameter $\tau$ is set to $1$.
To evaluate the usefulness of the data reuse methods with a small number of samples,
the agent collects only $N=3$ on-policy samples with trajectory length $T=100$ at each iteration.
In the data reuse methods, we reuse all previous data at later iterations.
In the plain PGPE\textsubscript{OB}, we just use the on-policy samples to estimate the gradients.
The discount factor is set at $\gamma=0.9$, and the learning rate is set at
$\varepsilon=0.1/\|\nabla_{\rho} \hat{\cJ}(\bm{\rho})\|.$

\subsubsection{Reaching Task with 2 Degrees of Freedom}
\begin{figure}[t]
    \centering
    \includegraphics[clip,width=0.7\columnwidth]{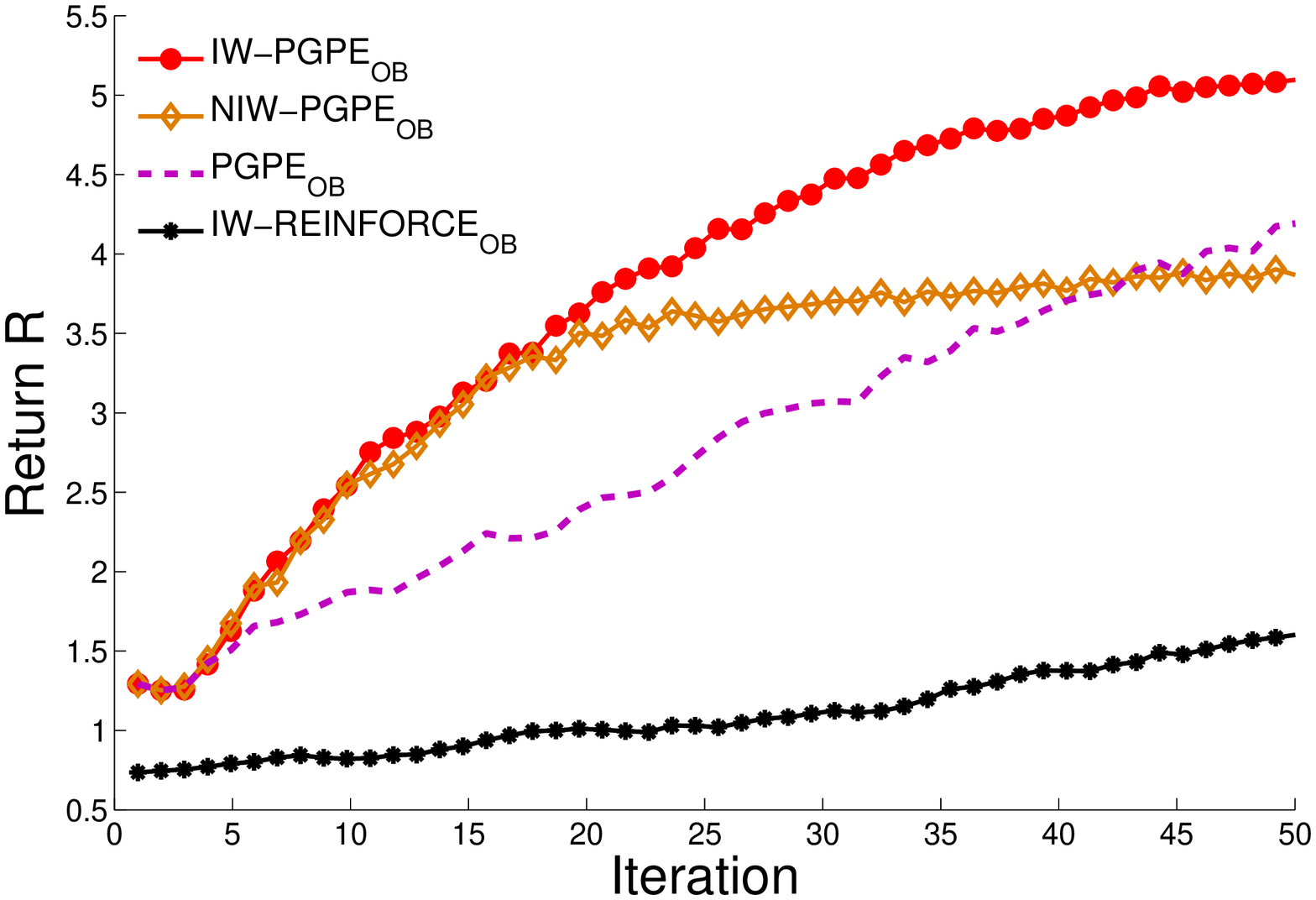}
    \caption{Average expected returns over $10$ runs as functions of the number of iterations for the reaching task with 2 degrees of freedom (right shoulder pitch and right elbow pitch).}
    \label{fig:j2-return}
\vspace*{5mm}
\subfigure[Distance]{{\includegraphics[clip,width=0.48\columnwidth]{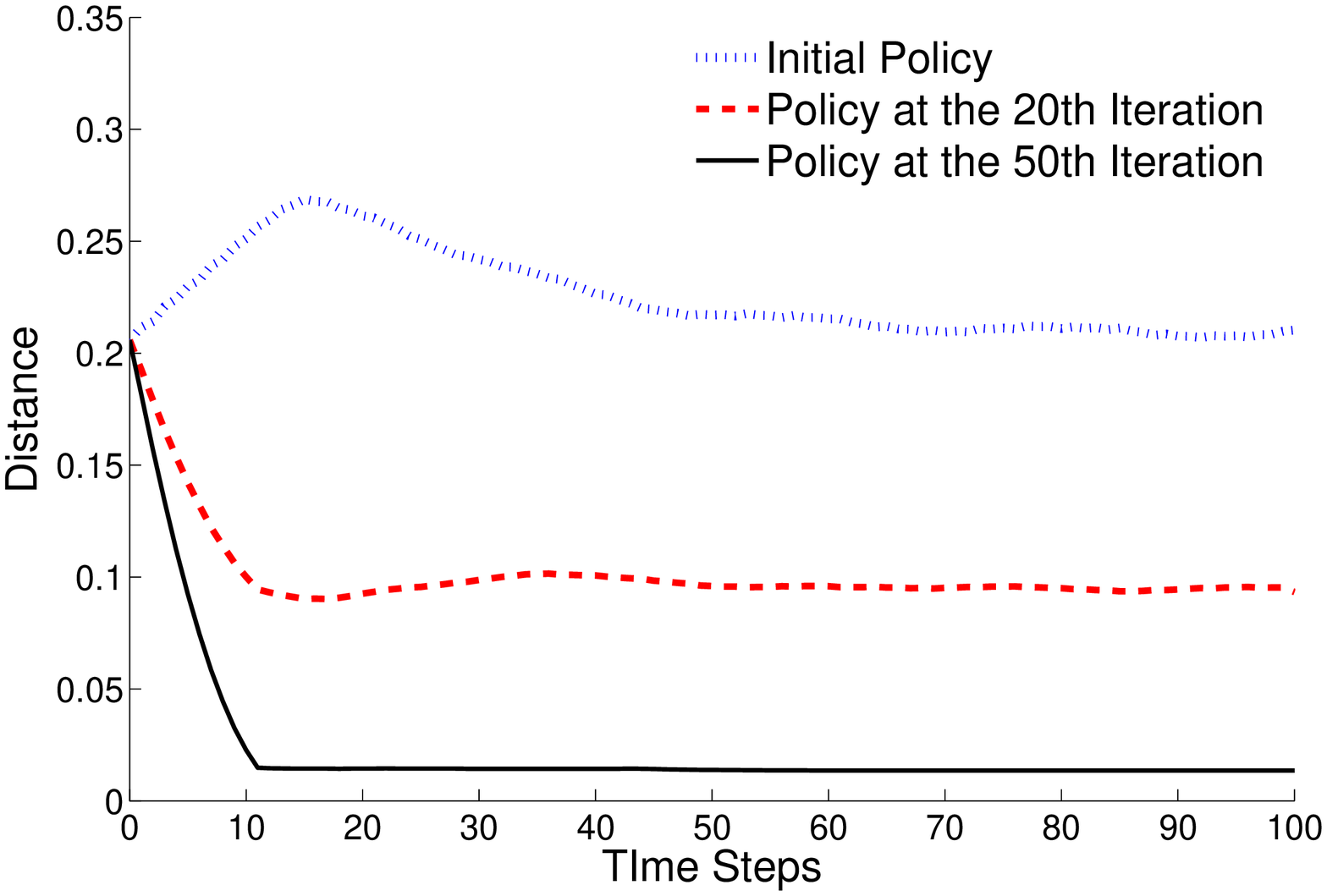}\label{fig:distance}}}
\subfigure[Control costs]{{\includegraphics[clip,width=0.48\columnwidth]{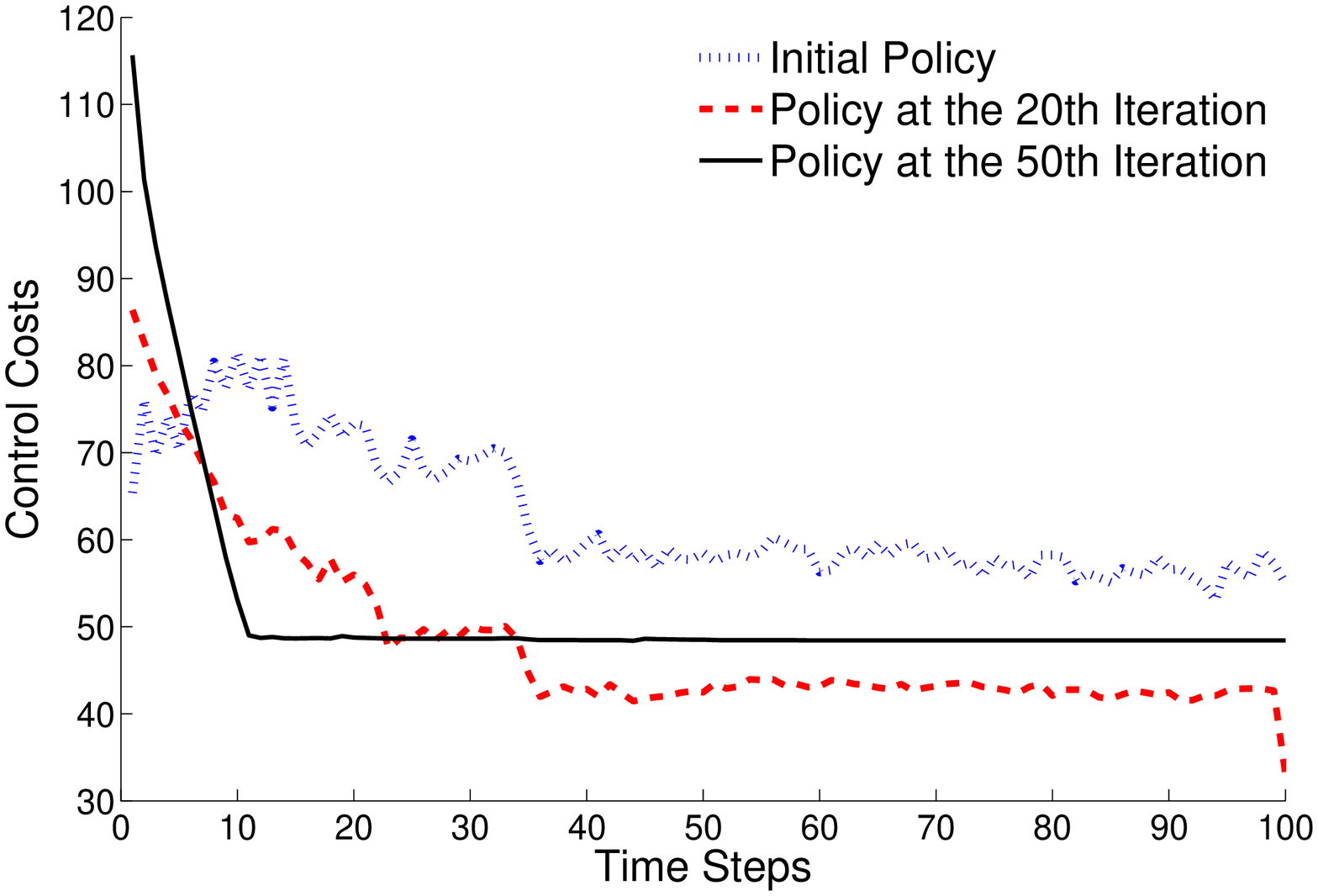}\label{fig:cost}}}
\caption{Distance and control costs of arm reaching with 2 degrees of freedom using the policy learned by IW-PGPE\textsubscript{OB}.}
\label{fig:J2_dtct}
    \end{figure}

First, we investigate the performance on the reaching task with only 2 degrees of freedom.
We fix the body of the robot and use only the right shoulder pitch and right elbow pitch.
Figure \ref{fig:j2-return} depicts the averaged expected return over 10 trials as a function of the number of iterations.
The expected return at each trial is computed from 50 newly-drawn test episodic data (which are not used for policy learning).
The graph shows that IW-PGPE\textsubscript{OB} nicely improves the performance over iterations
only with a small number of on-policy samples.
The plain PGPE\textsubscript{OB} can also improve the performance over iterations, but slowly.
NIW-PGPE\textsubscript{OB} is not as good as IW-PGPE\textsubscript{OB} especially at the later iterations, which is because of the inconsistent property of the NIW estimator.
The initial mean parameter is randomly chosen in this experiment,
which makes IW-REINFORCE\textsubscript{OB} not able to improve the performance significantly over iterations.
This result is consistent with the observation that the REINFORCE method is sensitive to the initial parameter values \cite{NN2012-ting}.

The distance from the right hand to the object and the control costs along the trajectory are also investigated.
We test the initial policy, the policy obtained at the $20$th iteration by IW-PGPE\textsubscript{OB}, and the policy obtained at the 50th iteration by IW-PGPE\textsubscript{OB}.
The results are shown in Figure \ref{fig:J2_dtct}.
From Figure \ref{fig:distance}, it is clear to see that the policy obtained at the 50th iteration decreases the distance fastest
compared with the initial policy and the policy obtained at the $20$th iteration.
This means the robot can reach the object fast by using the learned policy.
On the other hand, Figure \ref{fig:cost} shows that the control cost required
for executing the policy
obtained at the 50th iteration decreases steadily until the reaching
task is completed.
This is because the robot mainly adjusts the shoulder pitch in the beginning,
which consumes a larger amount of energy than the energy required for
controlling the elbow pitch.
Then, once the right hand gets closer to the target object, the robot
starts to adjust the elbow pitch
reach the target object.
The policy obtained at the 20th iteration actually consumes less control costs,
but it cannot move the arm to the target object.

\begin{figure}[t]
 \centering
\subfigure[]{\fcolorbox{black}{yellow}{\includegraphics[clip,width=0.18\columnwidth]{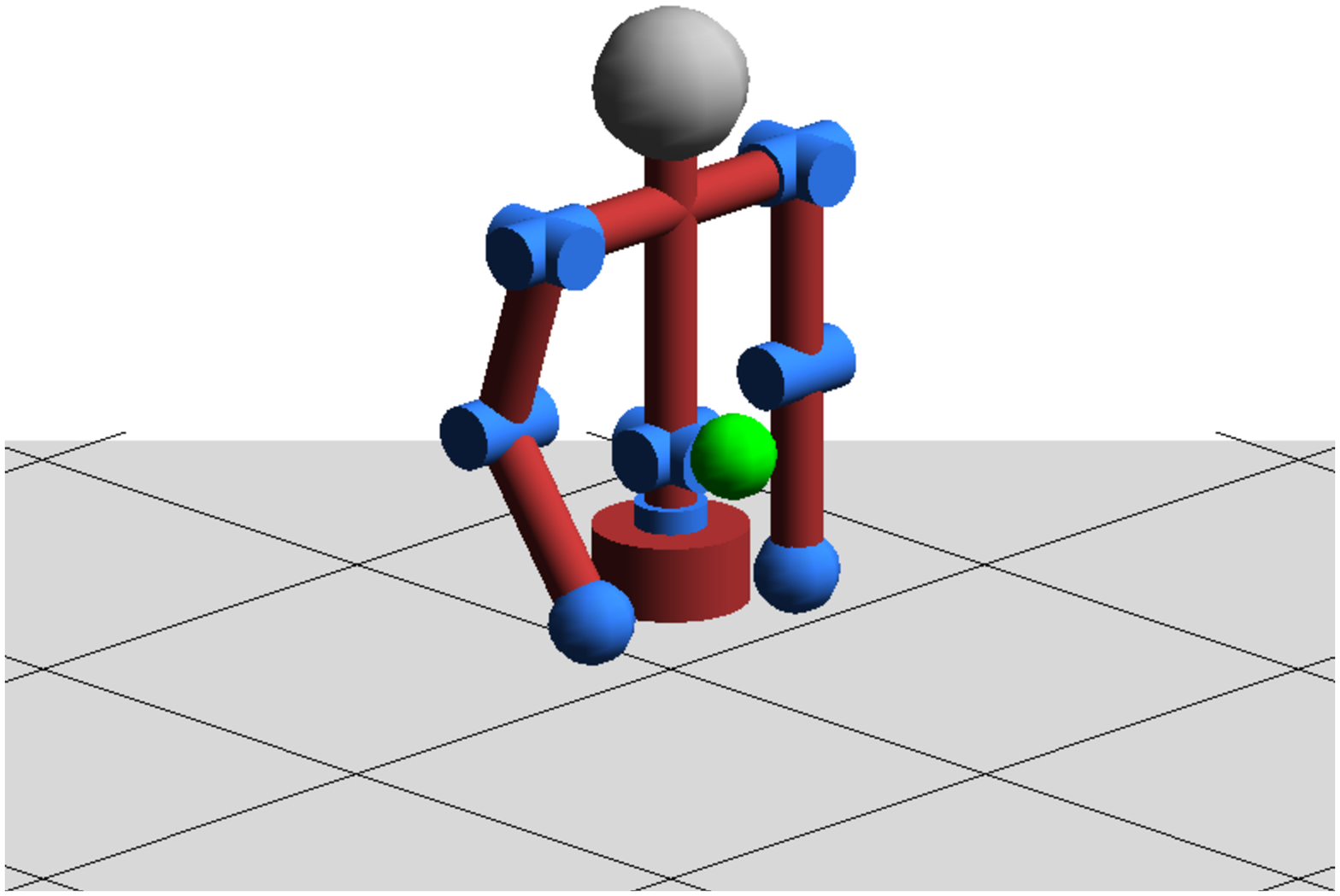}\label{fig:js-s1}}}
\subfigure[]{\fcolorbox{black}{yellow}{\includegraphics[clip,width=0.18\columnwidth]{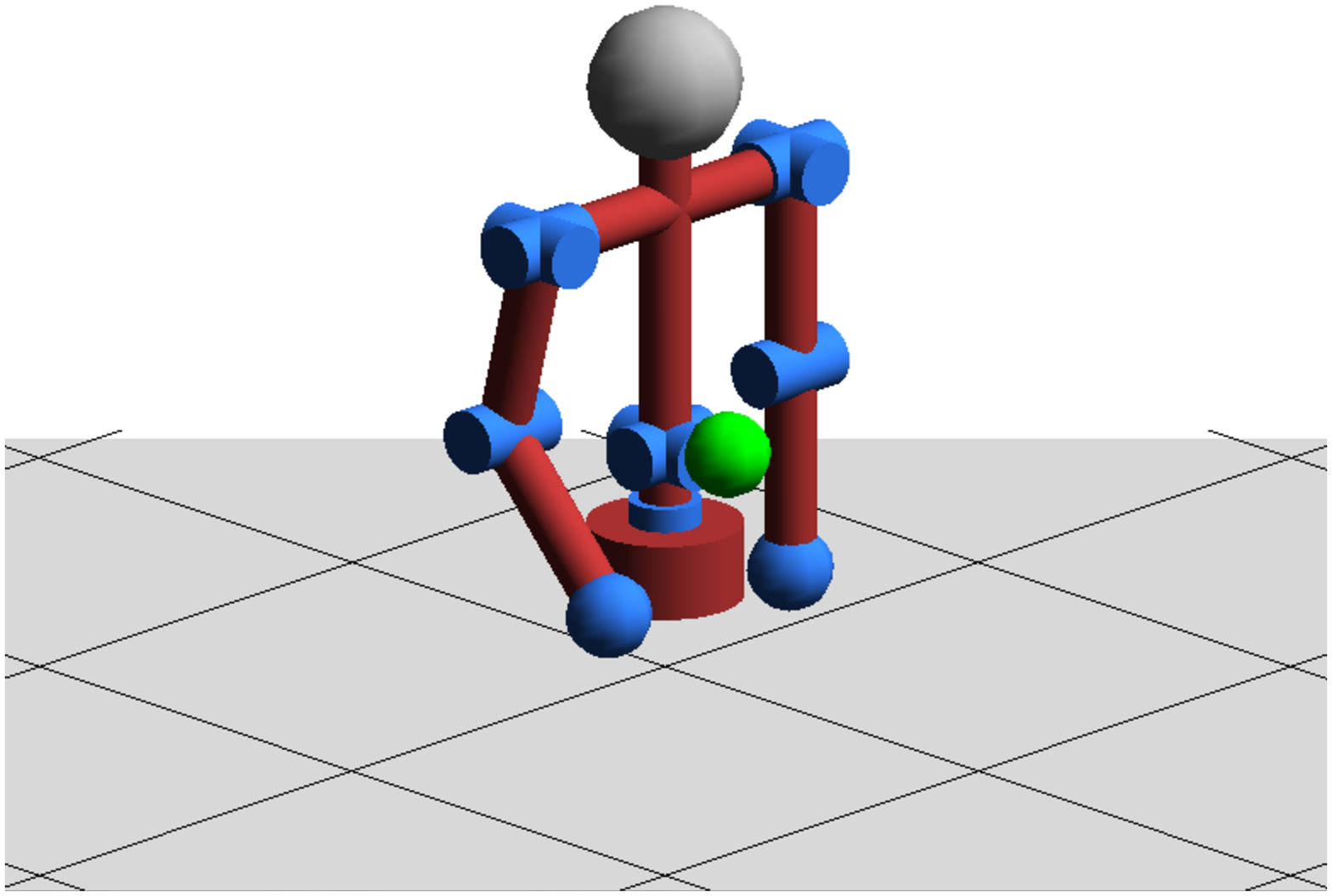}\label{fig:js-s2}}}
\subfigure[]{\fcolorbox{black}{yellow}{\includegraphics[clip,width=0.18\columnwidth]{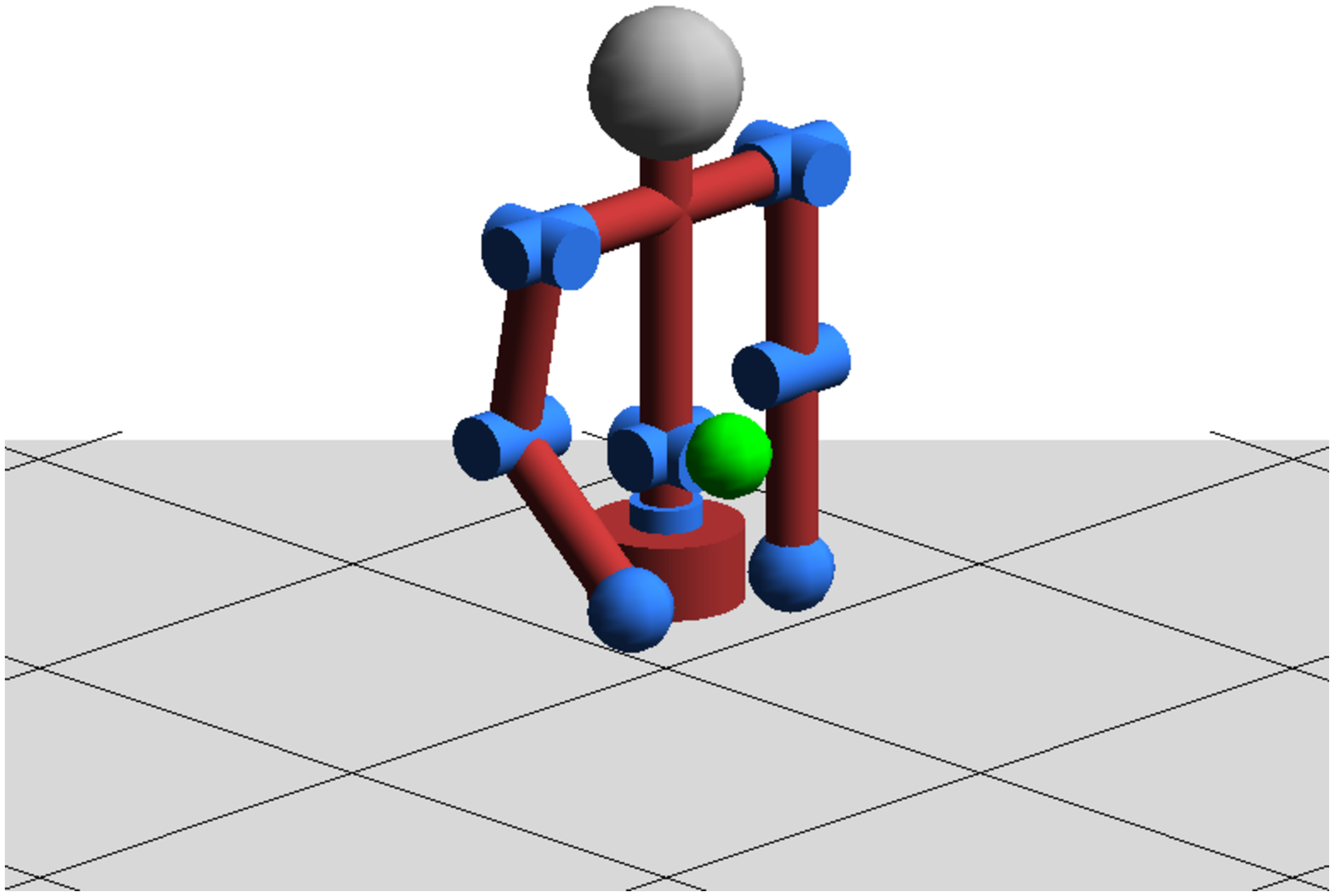}\label{fig:js-s3}}}
\subfigure[]{\fcolorbox{black}{yellow}{\includegraphics[clip,width=0.18\columnwidth]{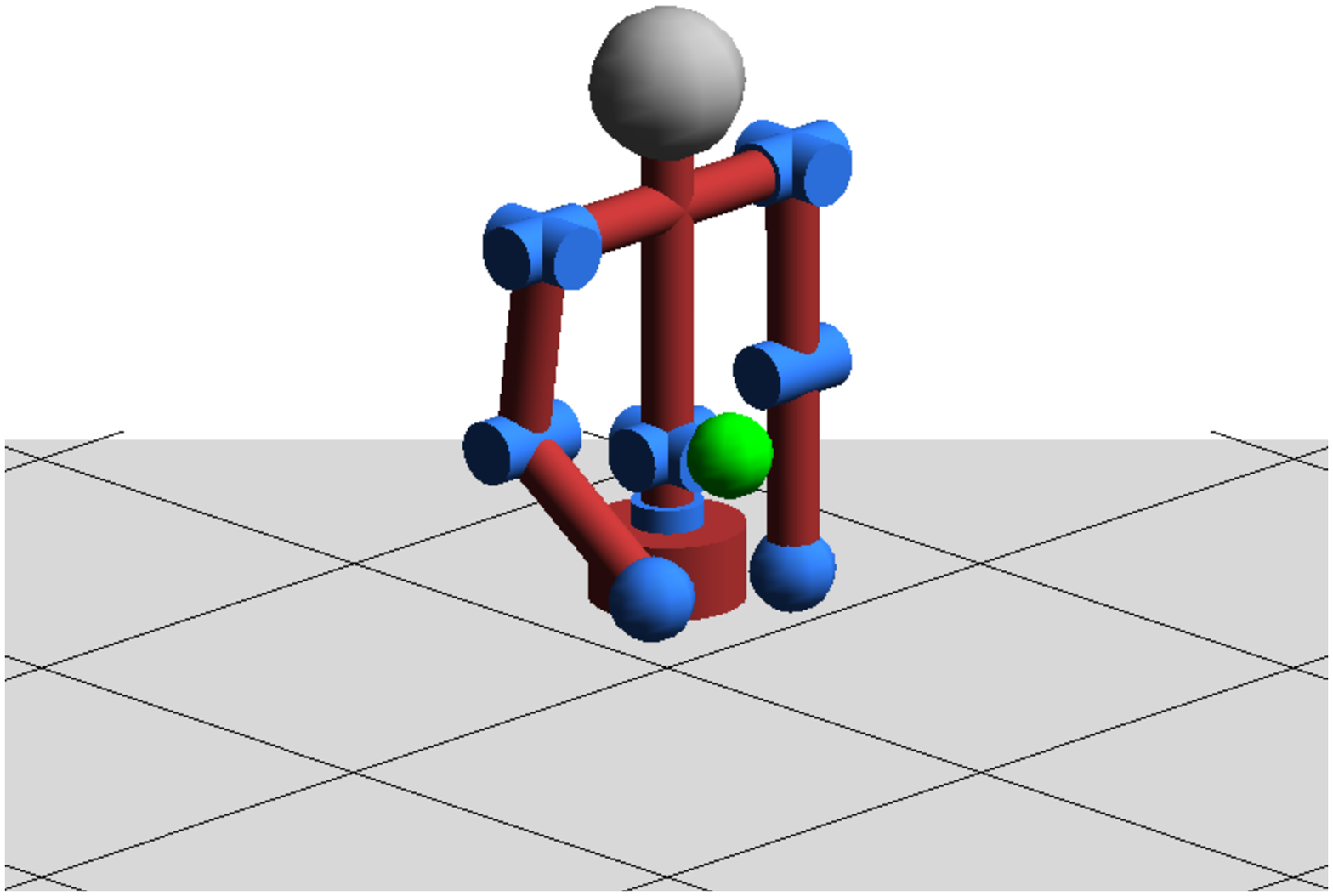}\label{fig:js-s4}}}
\subfigure[]{\fcolorbox{black}{yellow}{\includegraphics[clip,width=0.18\columnwidth]{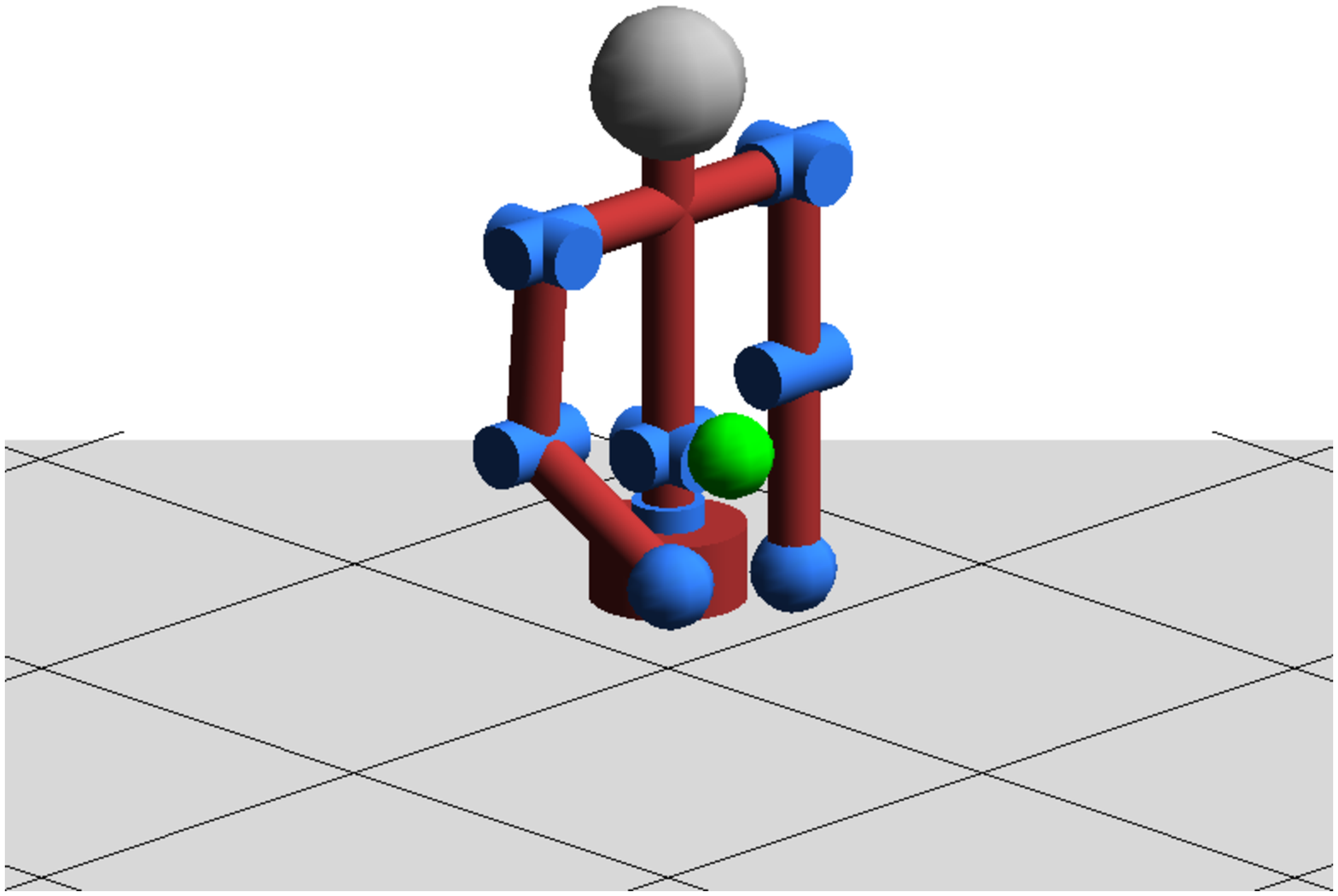}\label{fig:js-s5}}}
\subfigure[]{\fcolorbox{black}{yellow}{\includegraphics[clip,width=0.18\columnwidth]{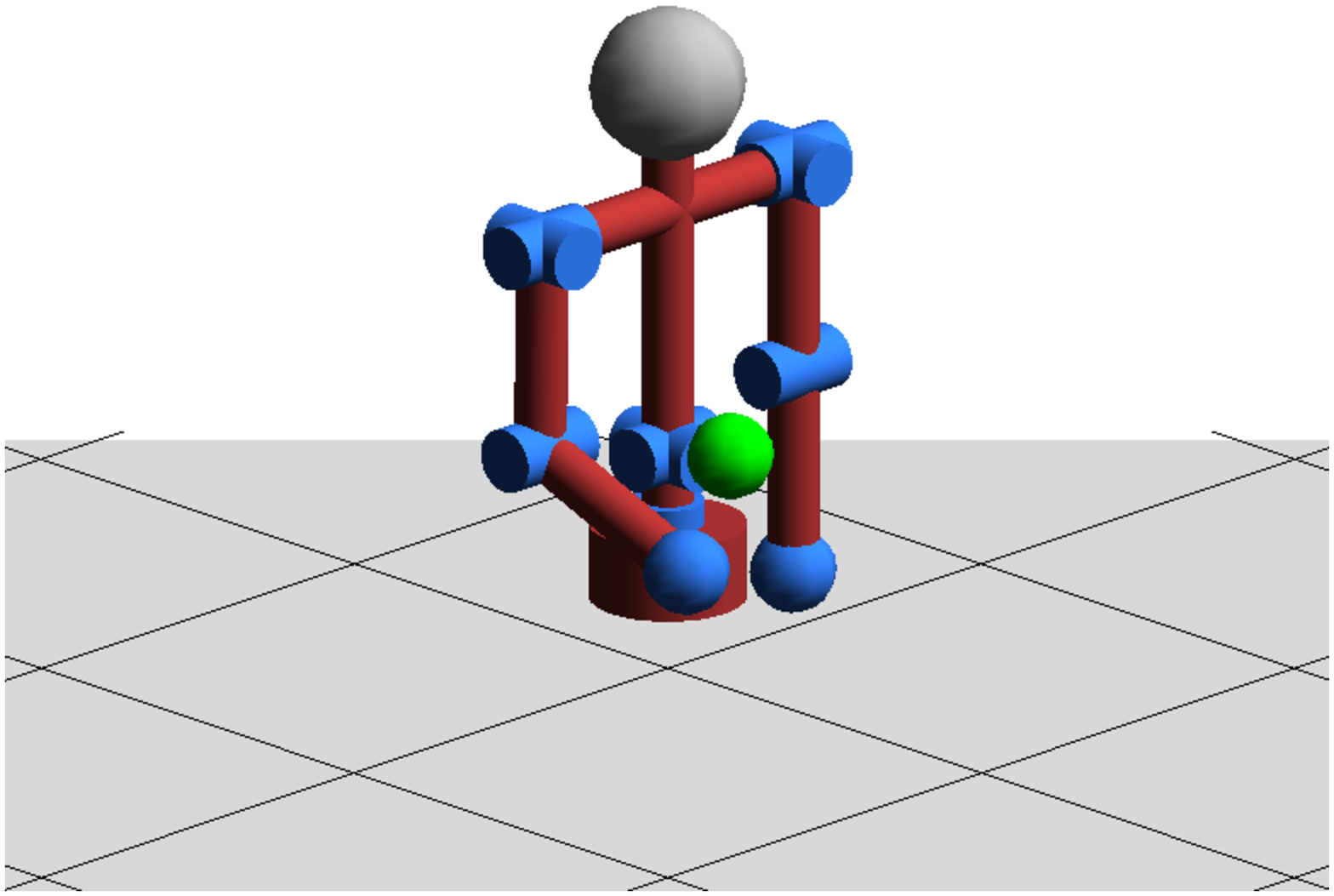}\label{fig:js-s6}}}
\subfigure[]{\fcolorbox{black}{yellow}{\includegraphics[clip,width=0.18\columnwidth]{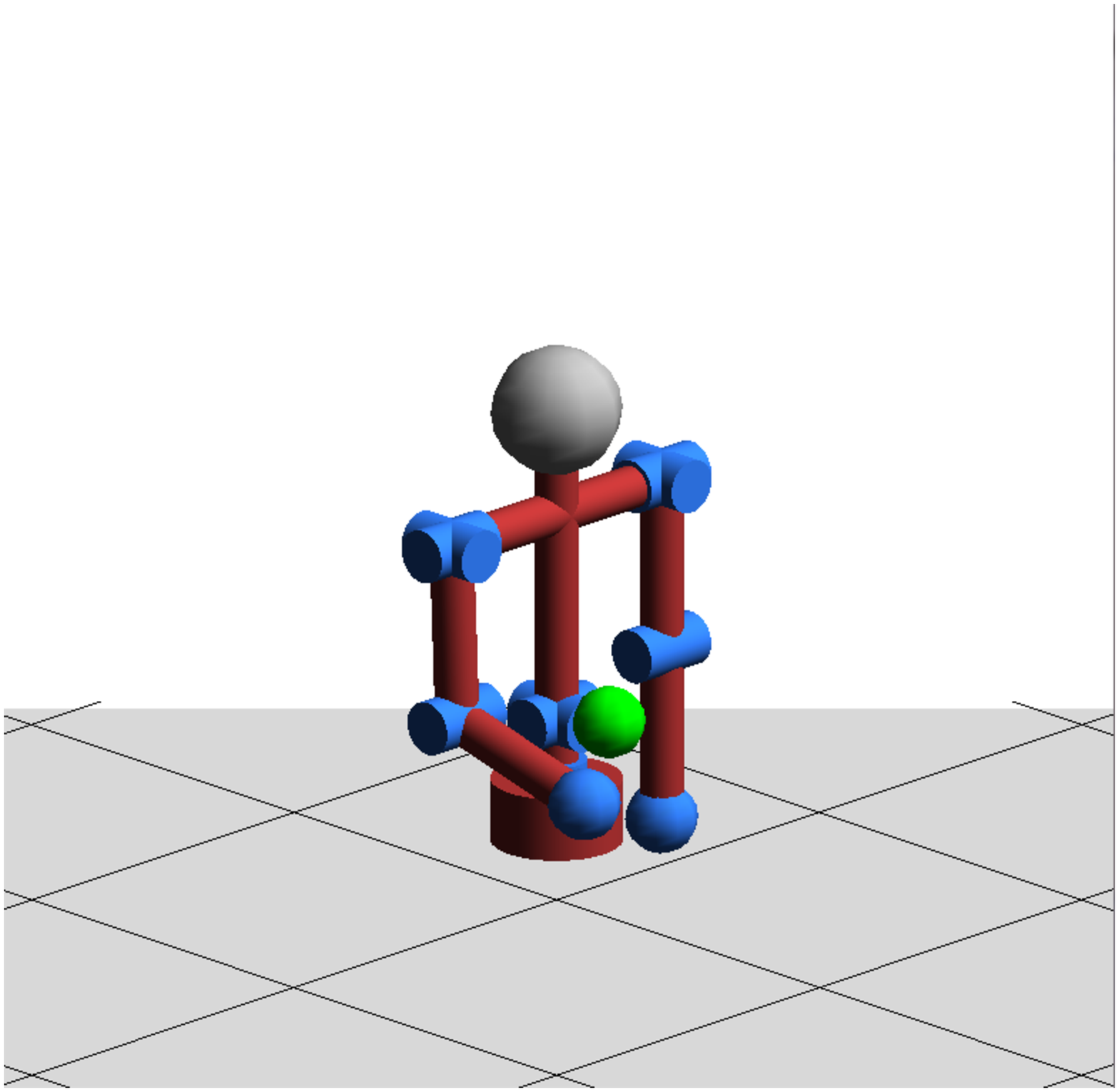}\label{fig:js-s7}}}
\subfigure[]{\fcolorbox{black}{yellow}{\includegraphics[clip,width=0.18\columnwidth]{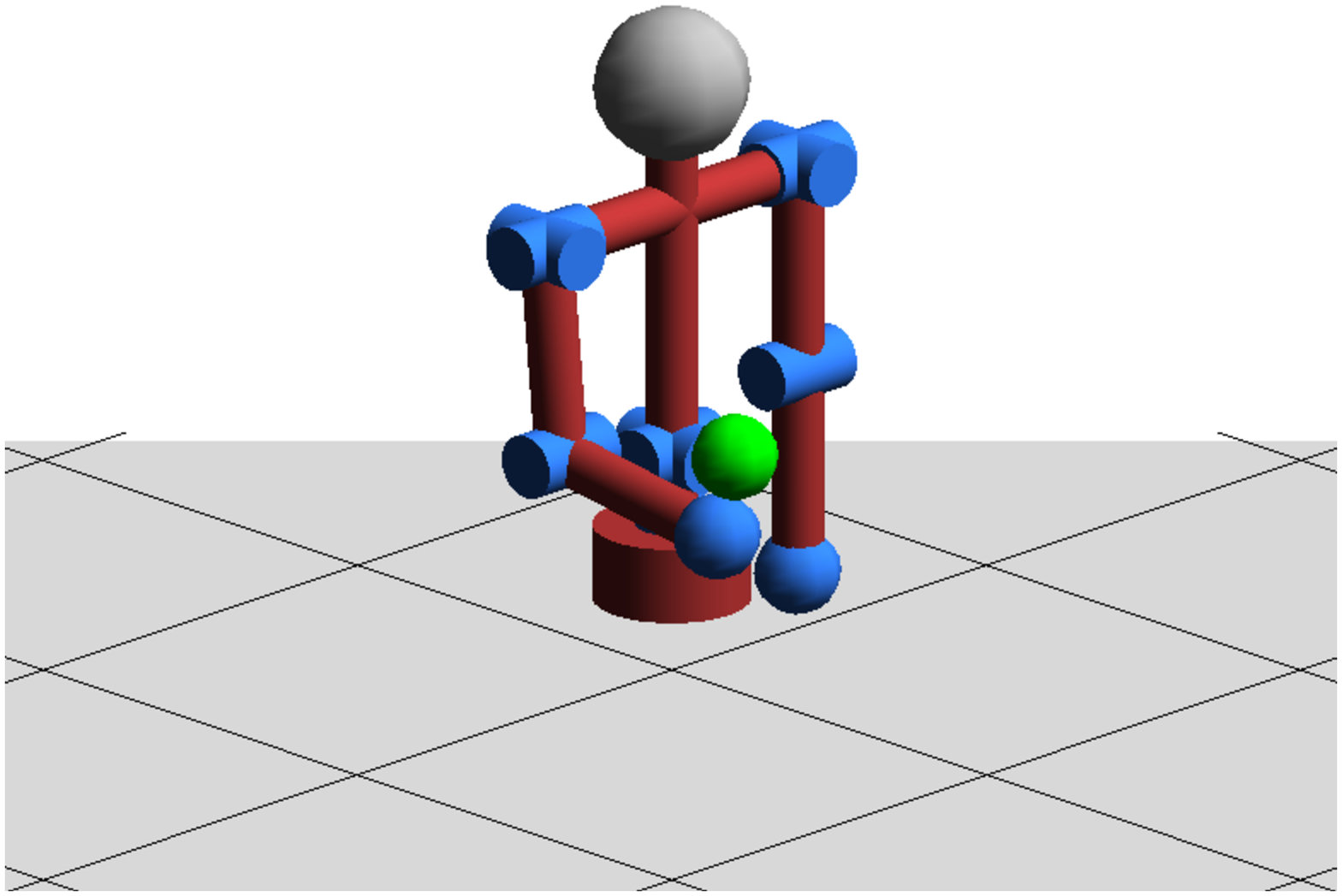}\label{fig:js-s8}}}
\subfigure[]{\fcolorbox{black}{yellow}{\includegraphics[clip,width=0.18\columnwidth]{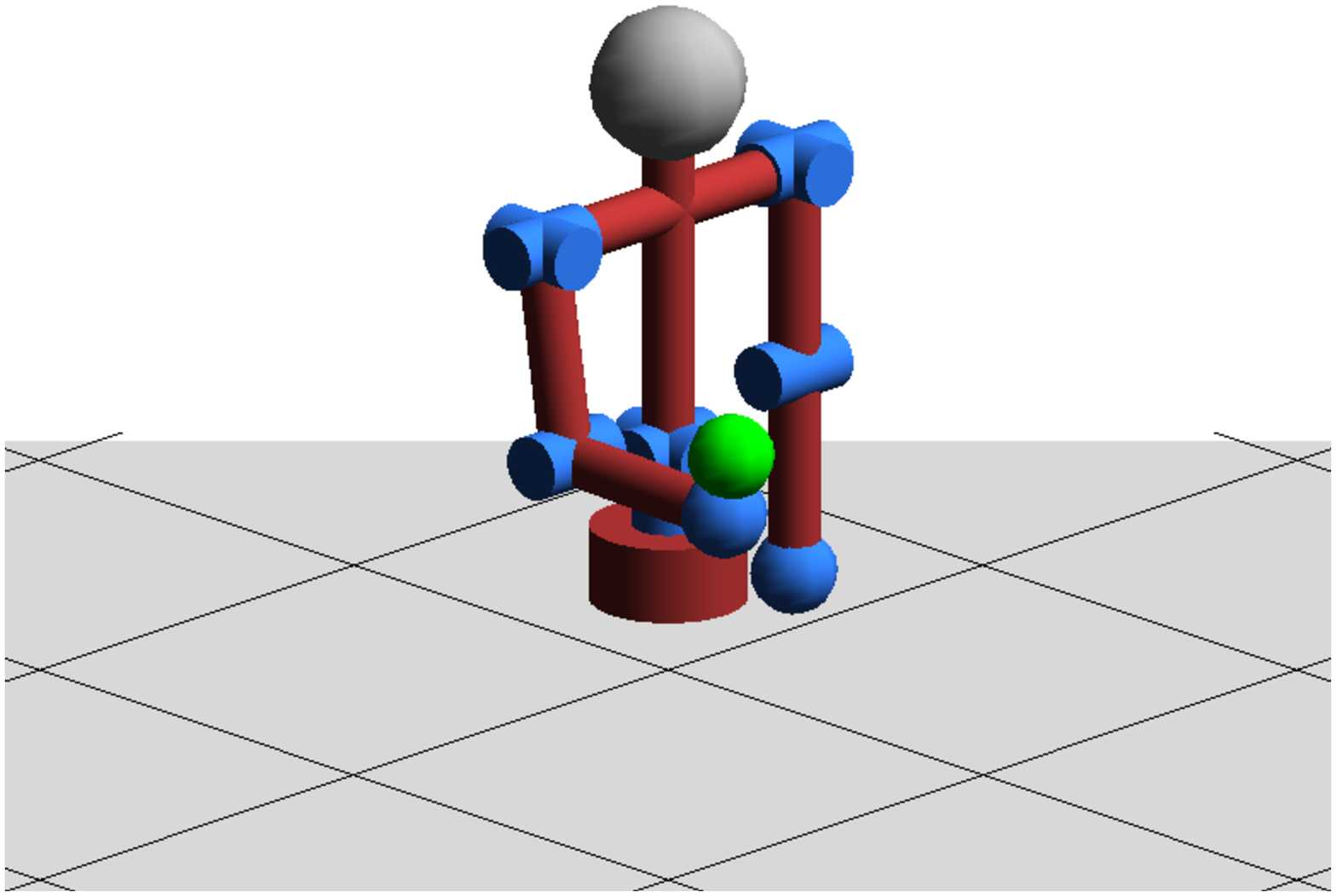}\label{fig:js-s9}}}
\subfigure[]{\fcolorbox{black}{yellow}{\includegraphics[clip,width=0.18\columnwidth]{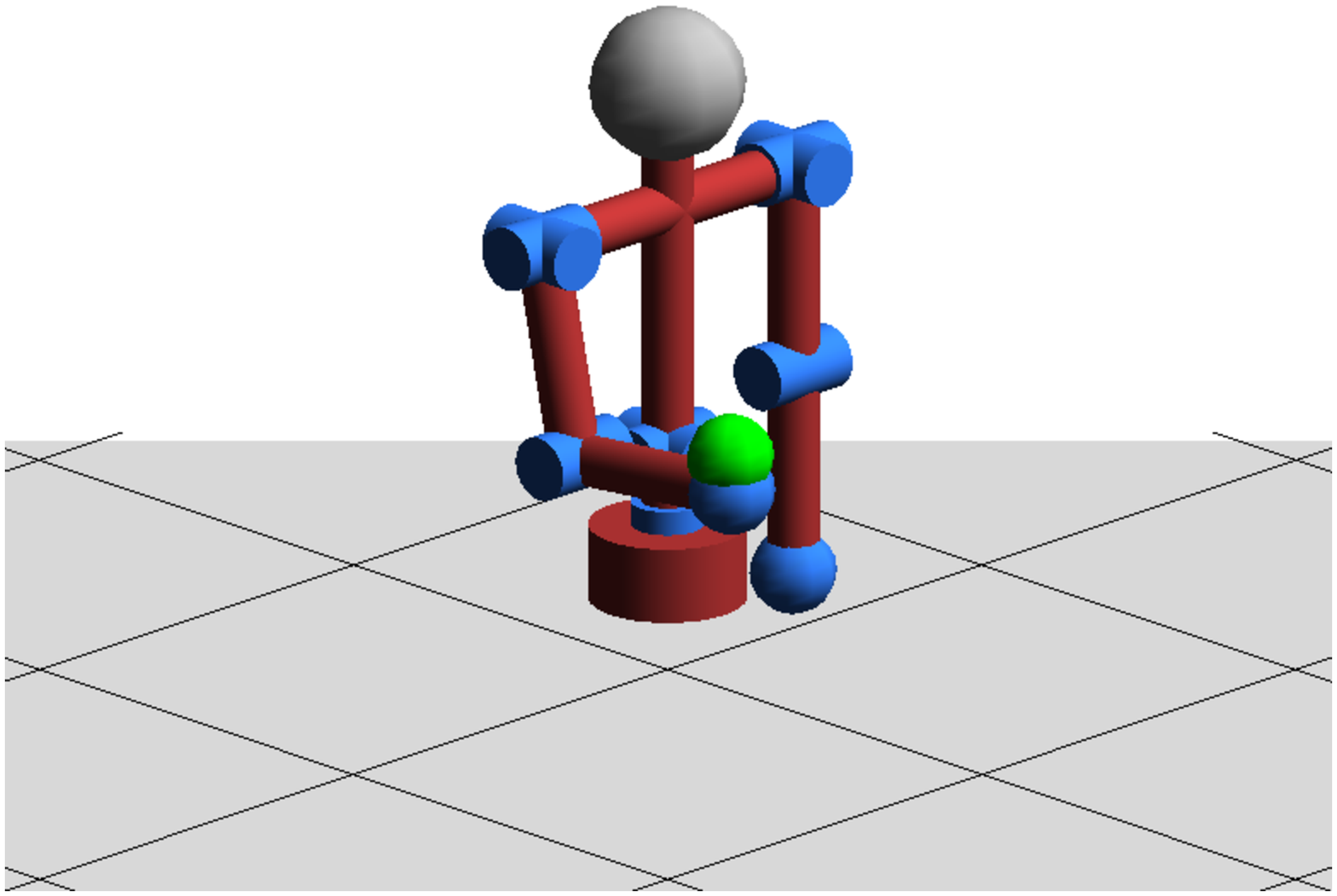}\label{fig:js-s10}}}
\caption{Typical example of arm reaching with 2 degrees of freedom using the policy obtained by
IW-PGPE\textsubscript{OB} at the 50th iteration.}
\label{fig:J2_trajectory}
\end{figure}

Figure \ref{fig:J2_trajectory} shows a typical solution of the reaching task with 2 degrees of freedom by IW-PGPE\textsubscript{OB}
(with the policy obtained at the 50th iteration). The images show that the policy learned by our proposed method successfully leads the right hand to the target object within only 10 time steps.

\subsubsection{Reaching Task with 4 Degrees of Freedom}
\begin{figure}[p]
    \centering
    \includegraphics[clip,width=0.7\columnwidth]{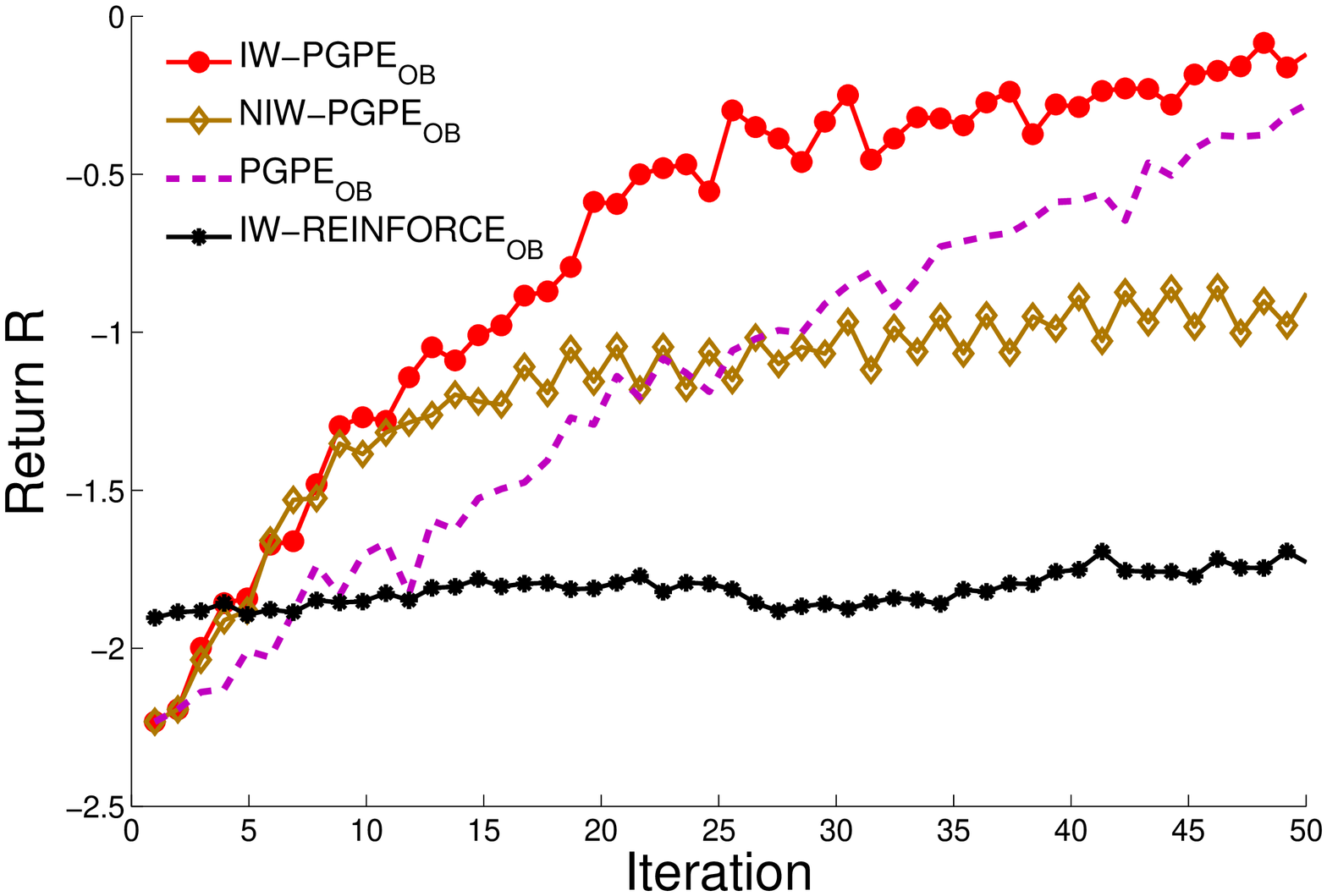}
    \caption{Average expected returns over $10$ runs as functions of the number of iterations for the reaching task with 4 degrees of freedom (right shoulder pitch, right elbow pitch, right shoulder roll, and torso yaw joint).}
    \label{fig:j4-return}
\vspace*{5mm}
\subfigure[]{\fcolorbox{black}{yellow}{\includegraphics[clip,width=0.2\columnwidth]{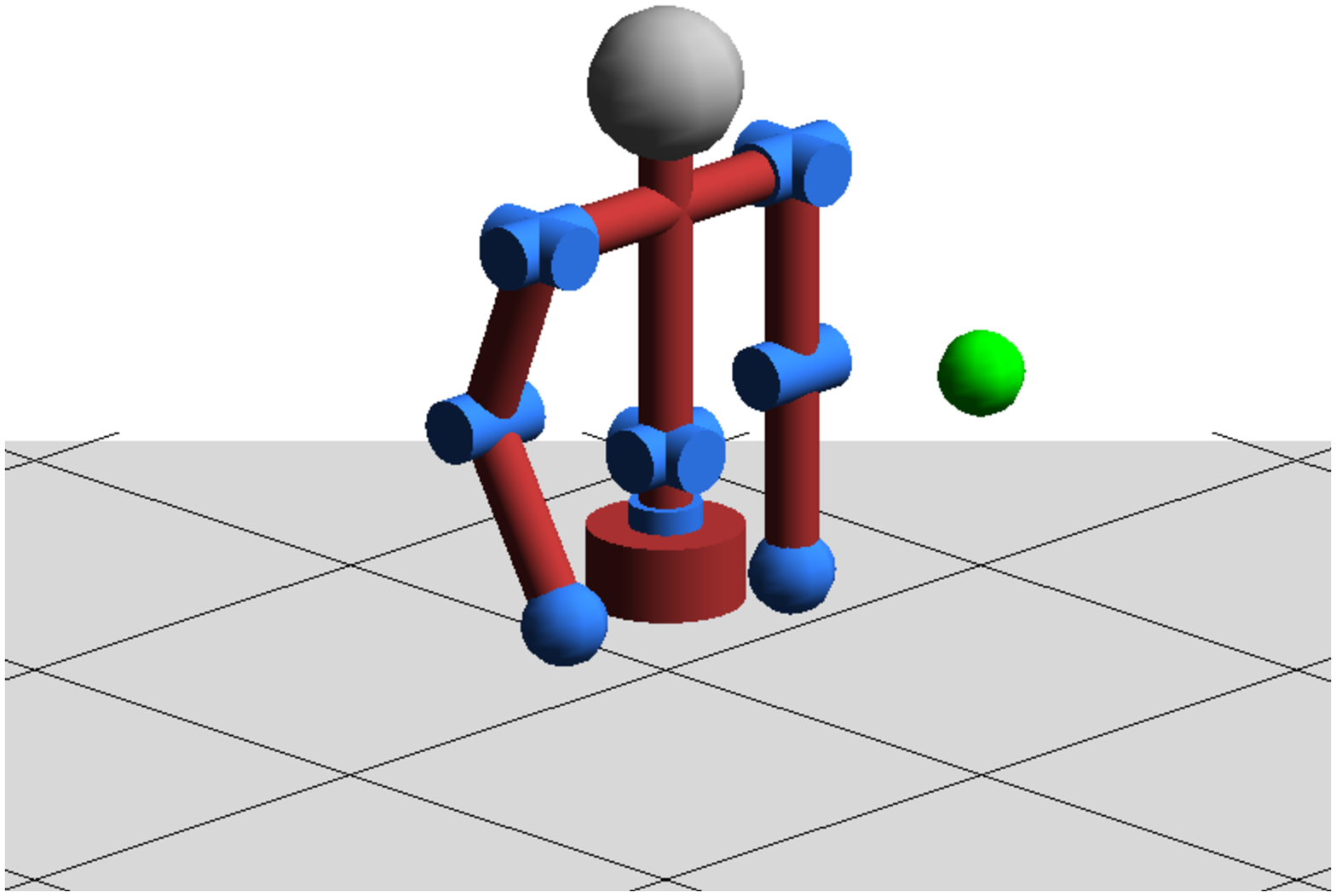}\label{fig:js4-s0}}}
\subfigure[]{\fcolorbox{black}{yellow}{\includegraphics[clip,width=0.2\columnwidth]{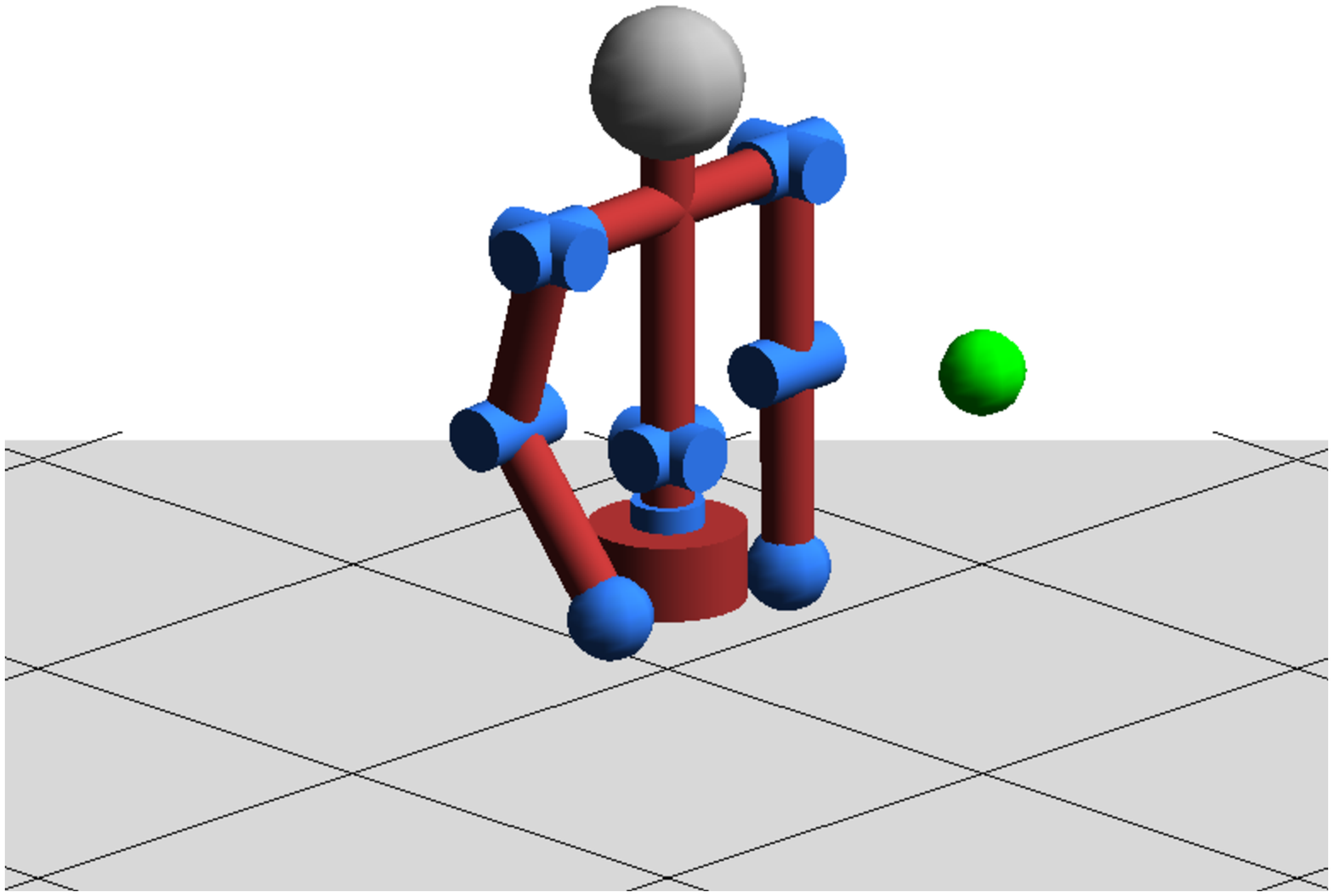}\label{fig:js4-s1}}}
\subfigure[]{\fcolorbox{black}{yellow}{\includegraphics[clip,width=0.2\columnwidth]{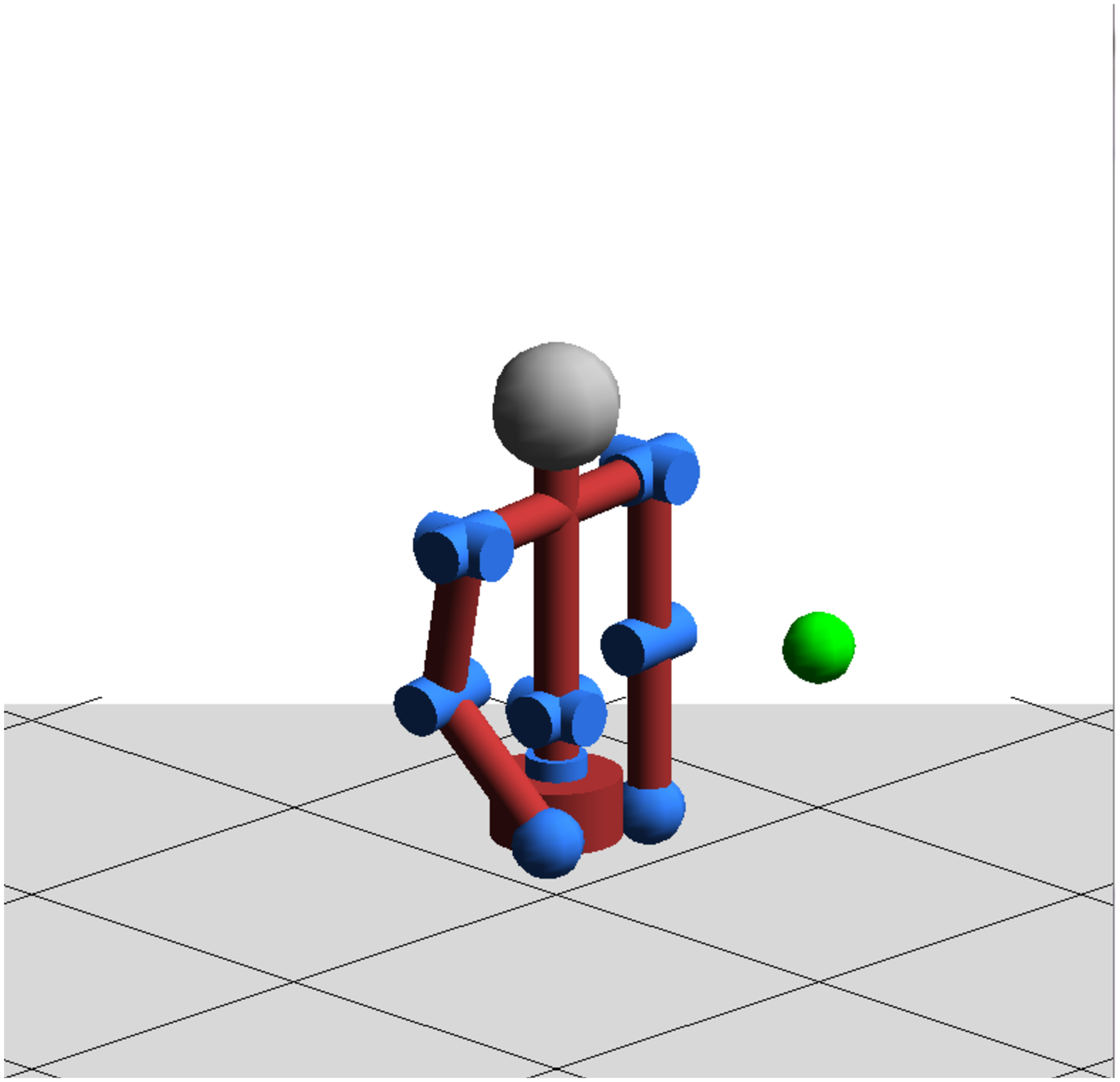}\label{fig:js4-s2}}}
\subfigure[]{\fcolorbox{black}{yellow}{\includegraphics[clip,width=0.2\columnwidth]{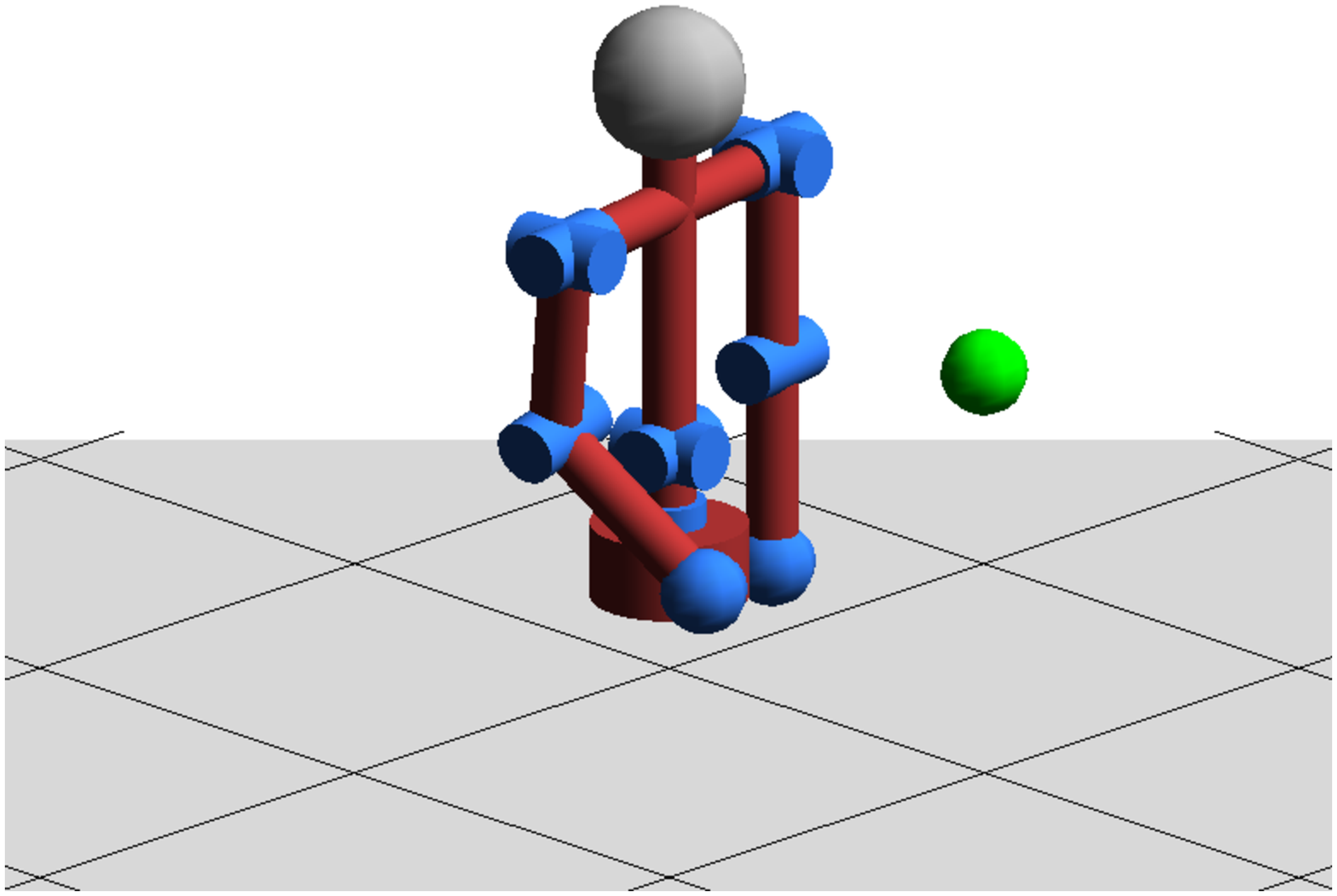}\label{fig:js4-s3}}}
\subfigure[]{\fcolorbox{black}{yellow}{\includegraphics[clip,width=0.2\columnwidth]{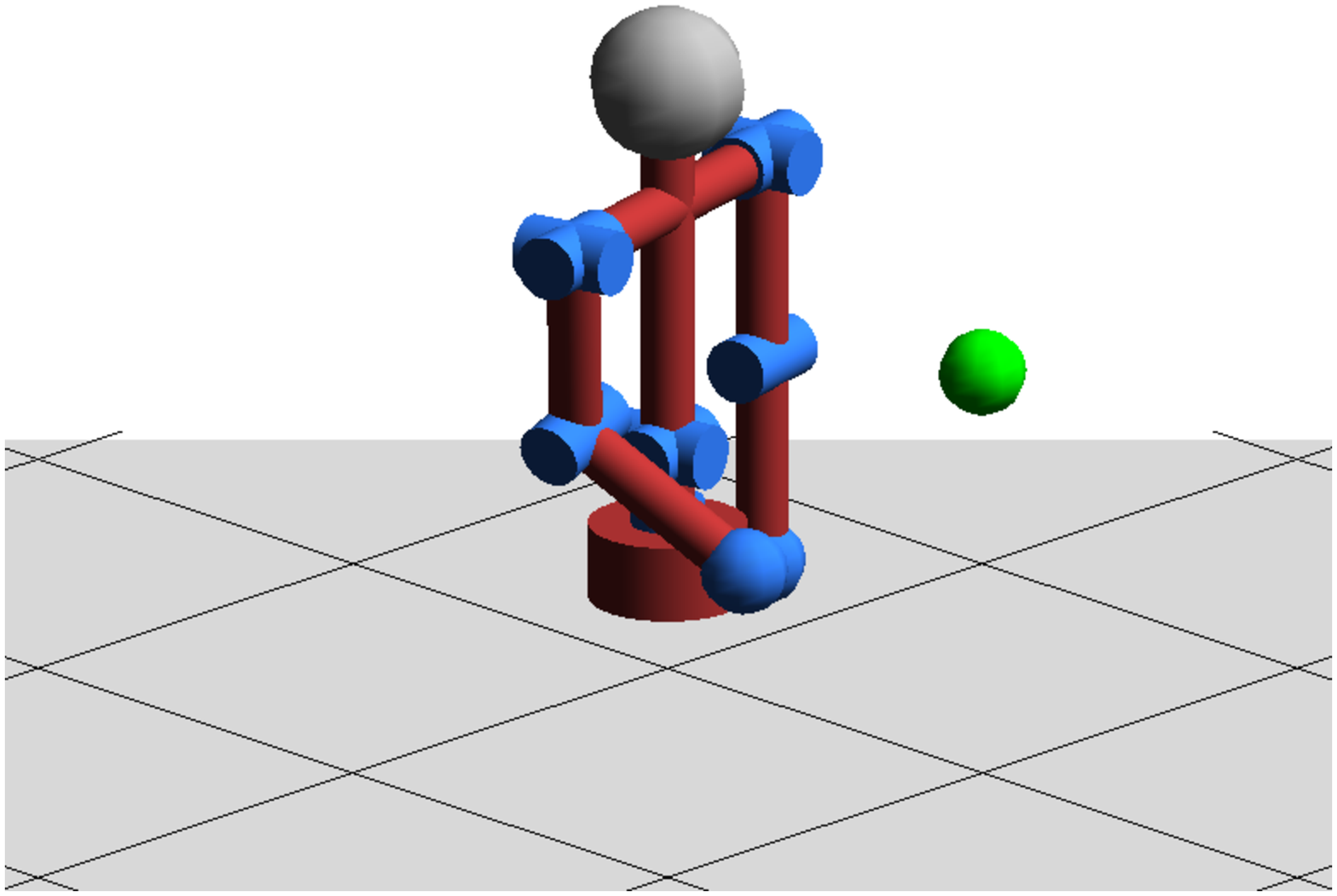}\label{fig:js4-s4}}}
\subfigure[]{\fcolorbox{black}{yellow}{\includegraphics[clip,width=0.2\columnwidth]{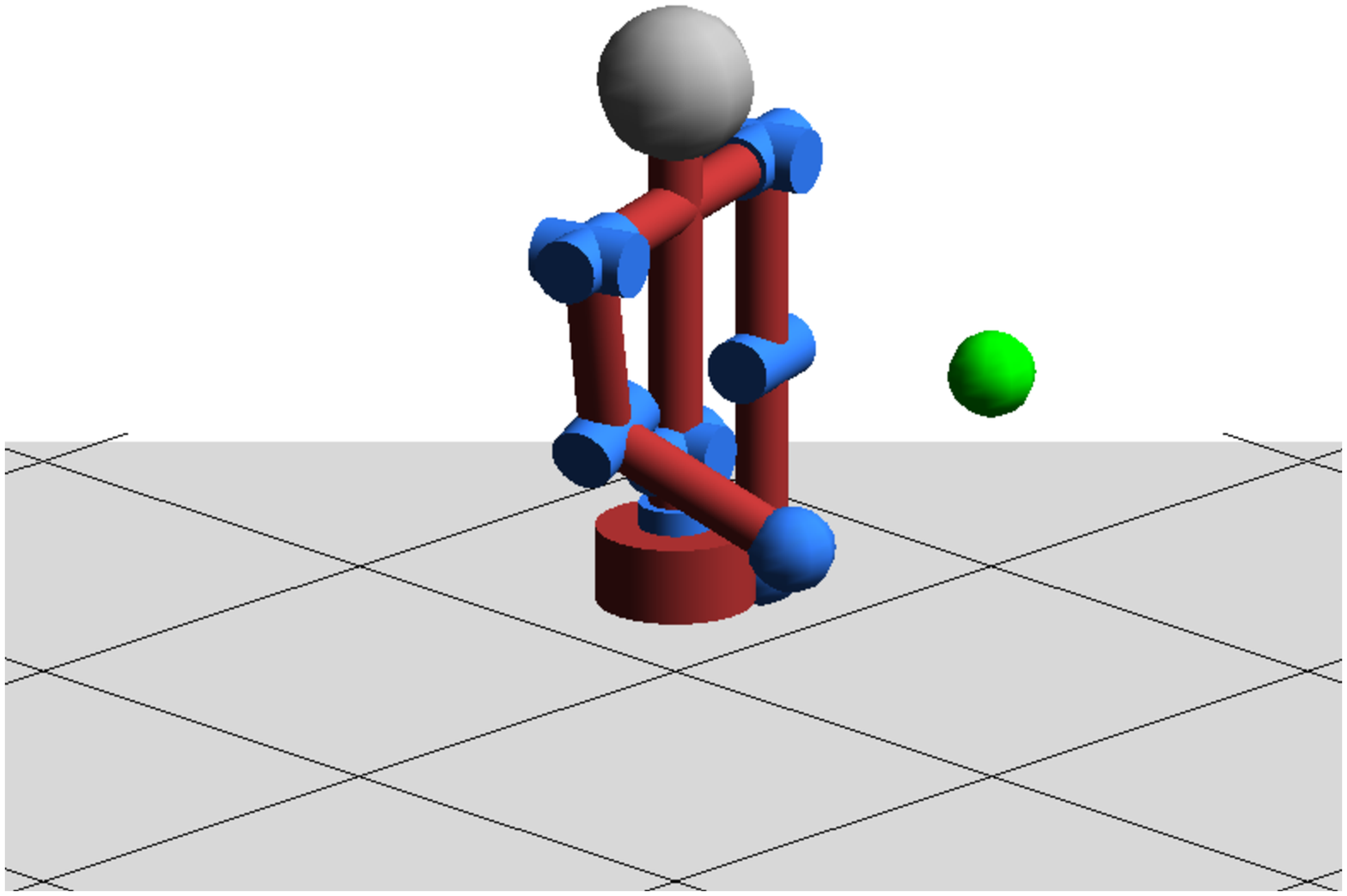}\label{fig:js4-s5}}}
\subfigure[]{\fcolorbox{black}{yellow}{\includegraphics[clip,width=0.2\columnwidth]{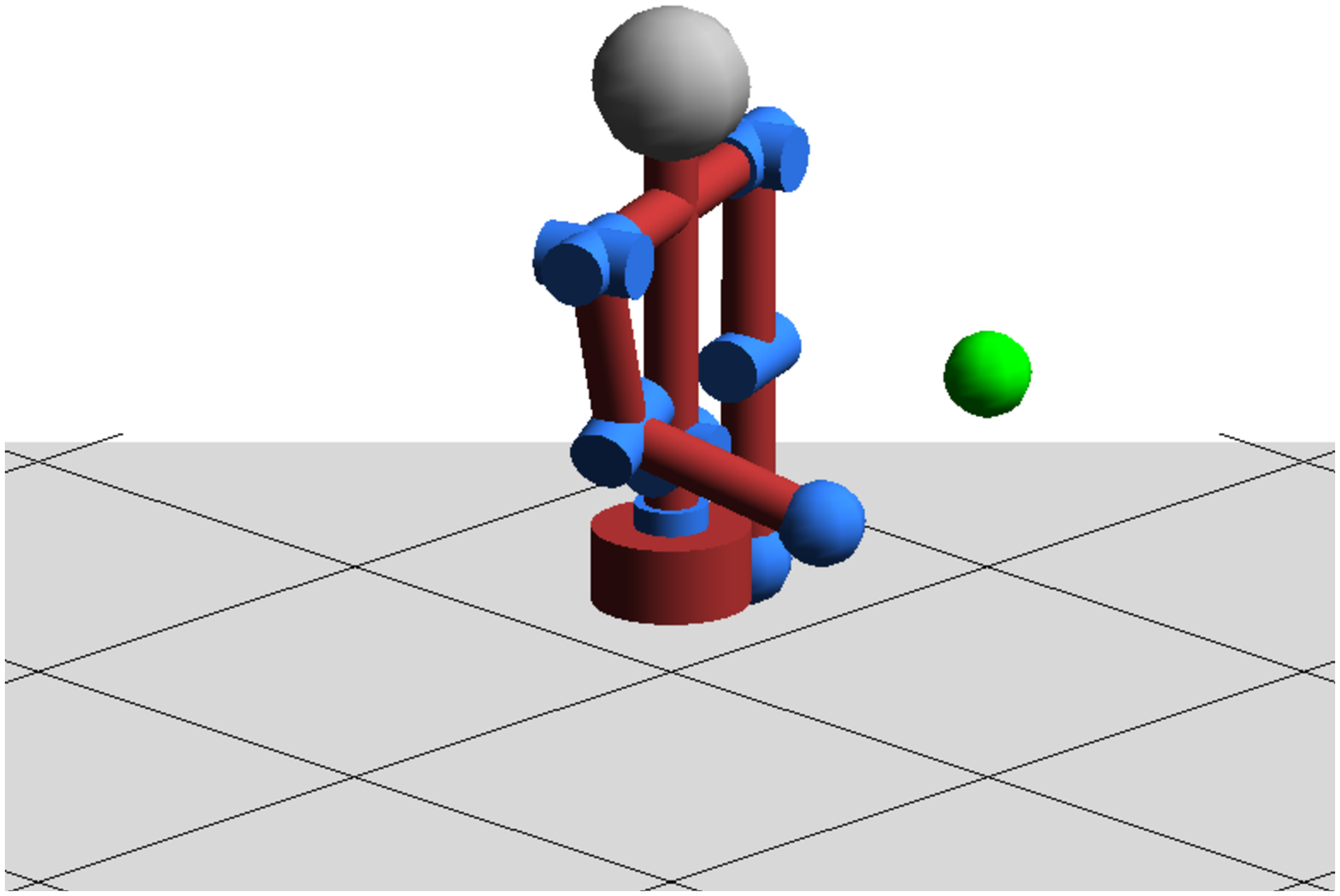}\label{fig:js4-s6}}}
\subfigure[]{\fcolorbox{black}{yellow}{\includegraphics[clip,width=0.2\columnwidth]{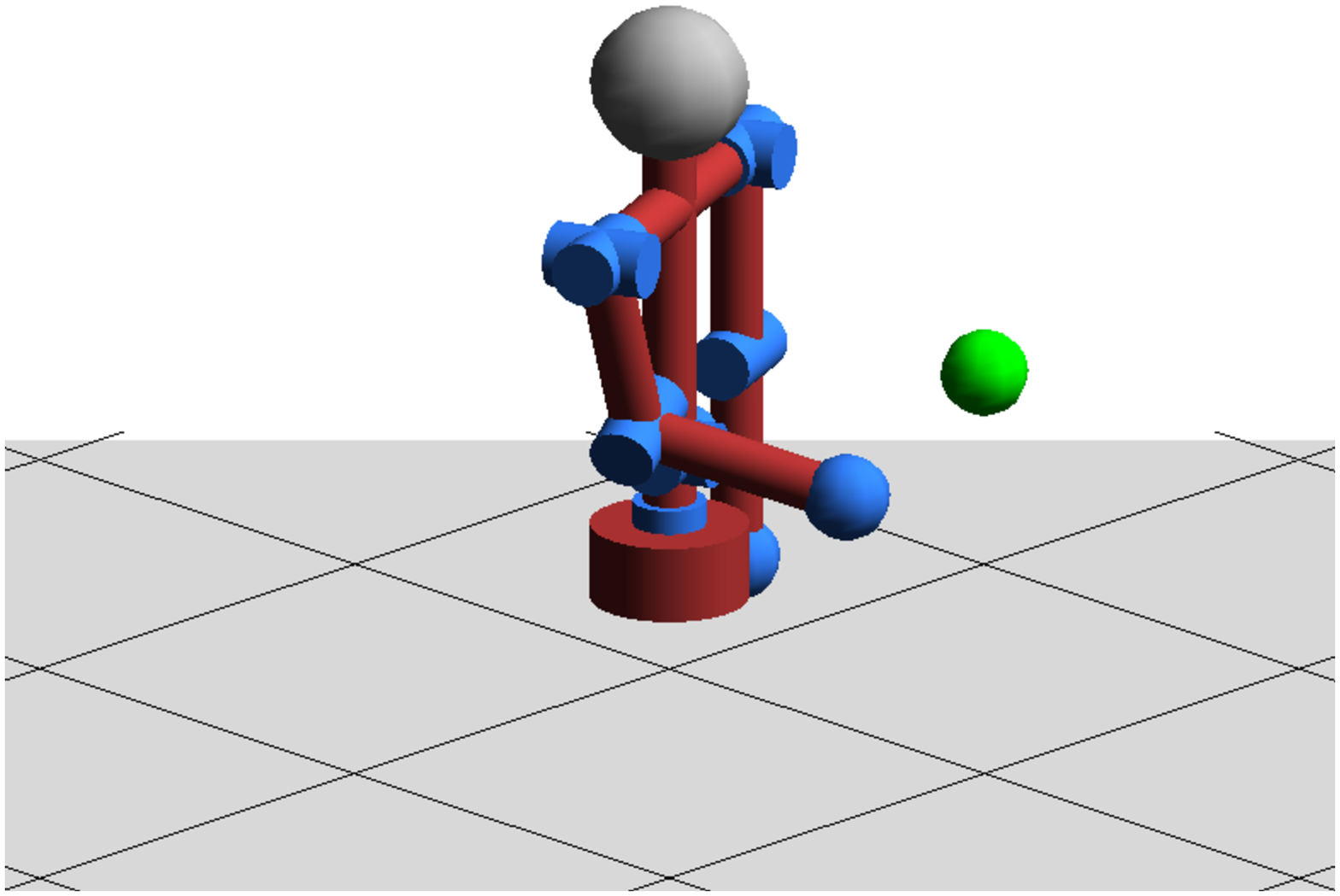}\label{fig:js4-s7}}}
\subfigure[]{\fcolorbox{black}{yellow}{\includegraphics[clip,width=0.2\columnwidth]{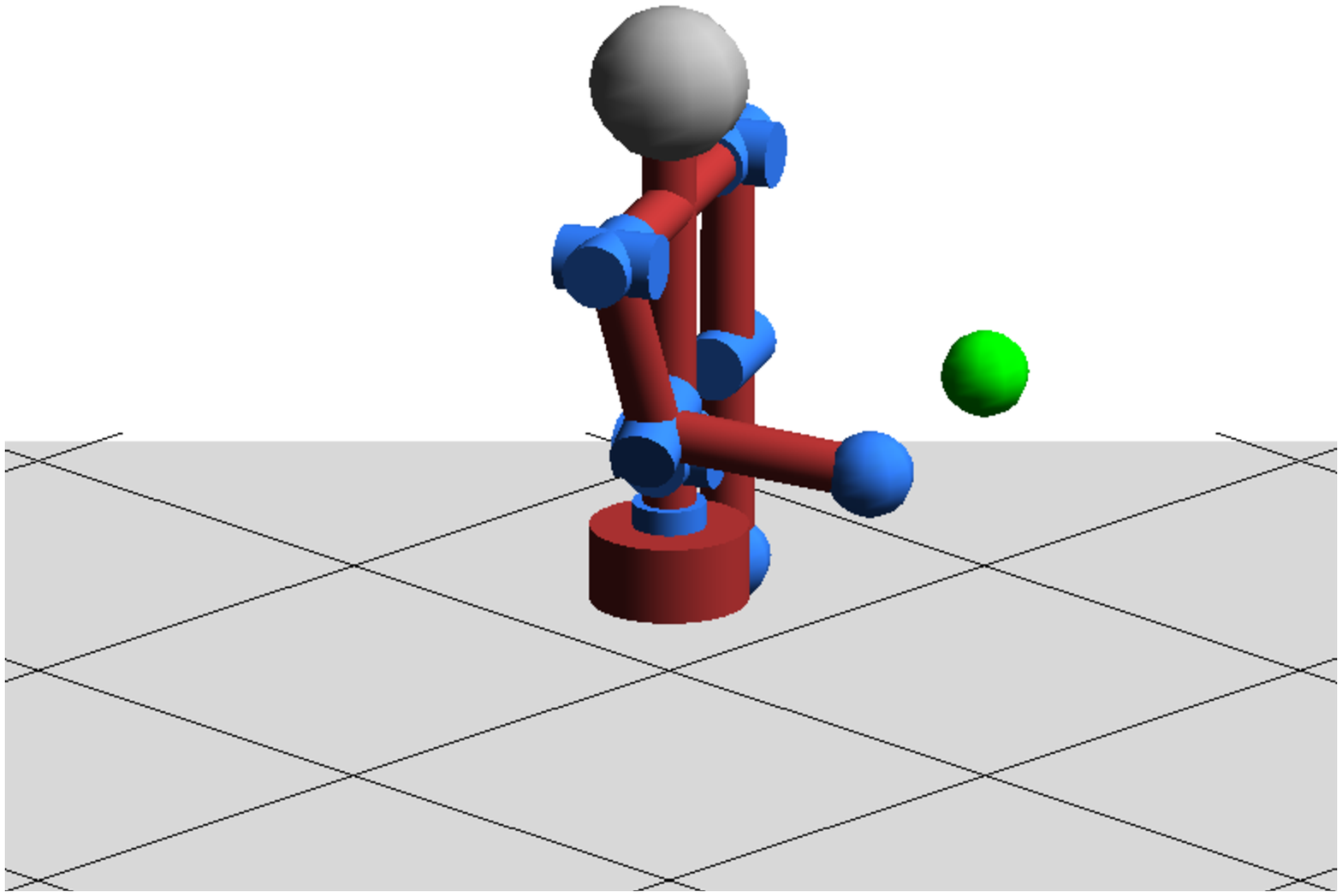}\label{fig:js4-s8}}}
\subfigure[]{\fcolorbox{black}{yellow}{\includegraphics[clip,width=0.2\columnwidth]{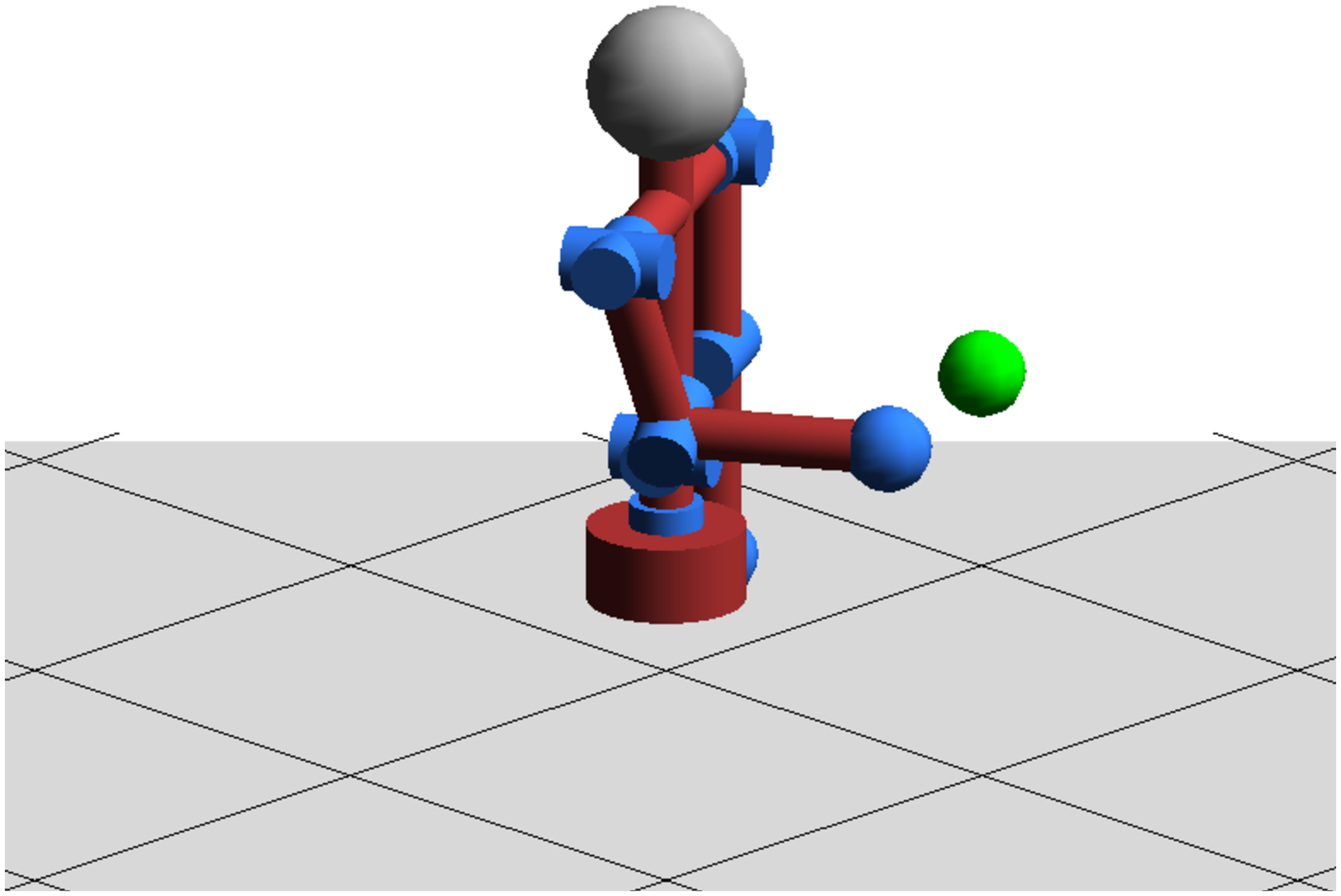}\label{fig:js4-s9}}}
\subfigure[]{\fcolorbox{black}{yellow}{\includegraphics[clip,width=0.2\columnwidth]{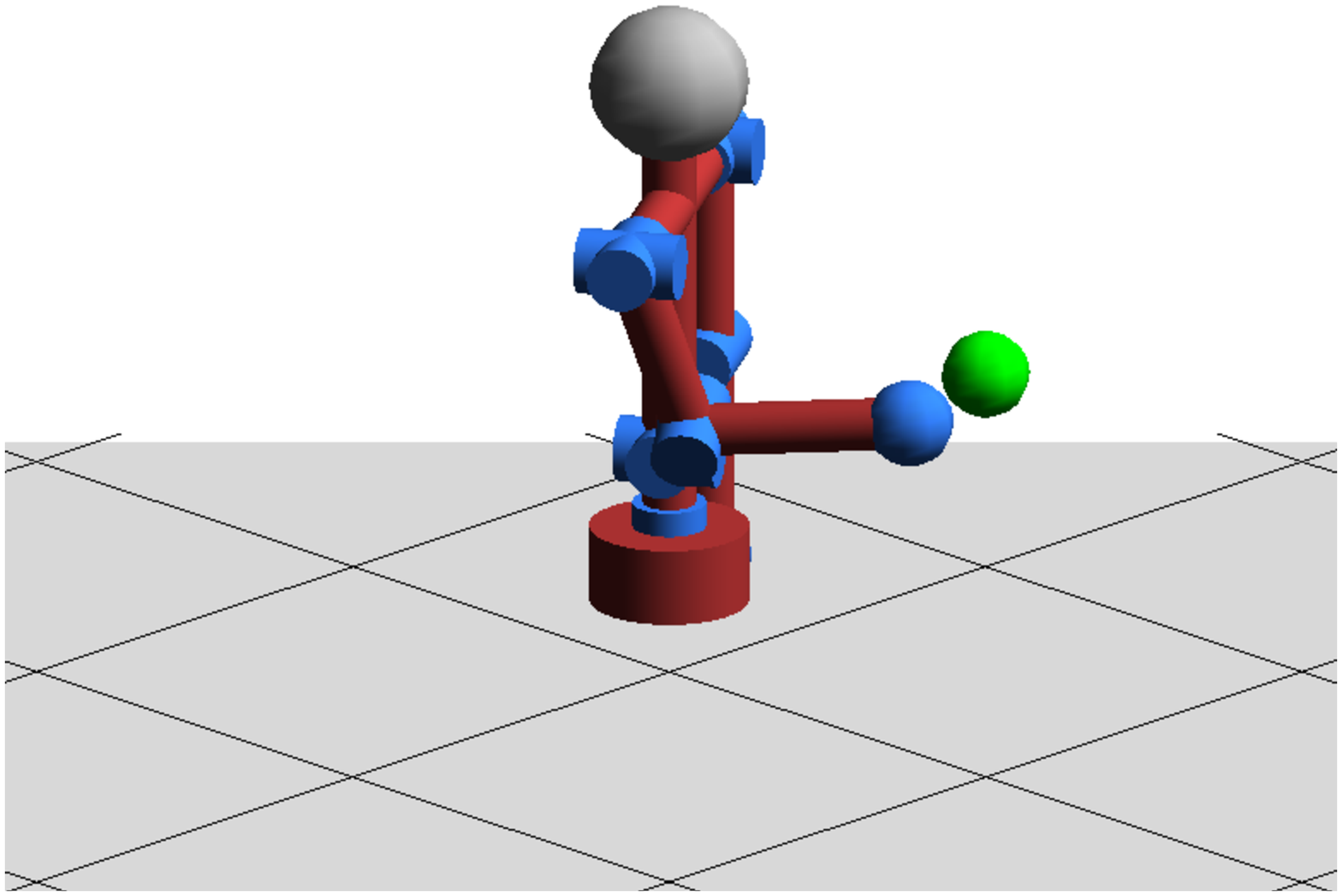}\label{fig:js4-s10}}}
\subfigure[]{\fcolorbox{black}{yellow}{\includegraphics[clip,width=0.2\columnwidth]{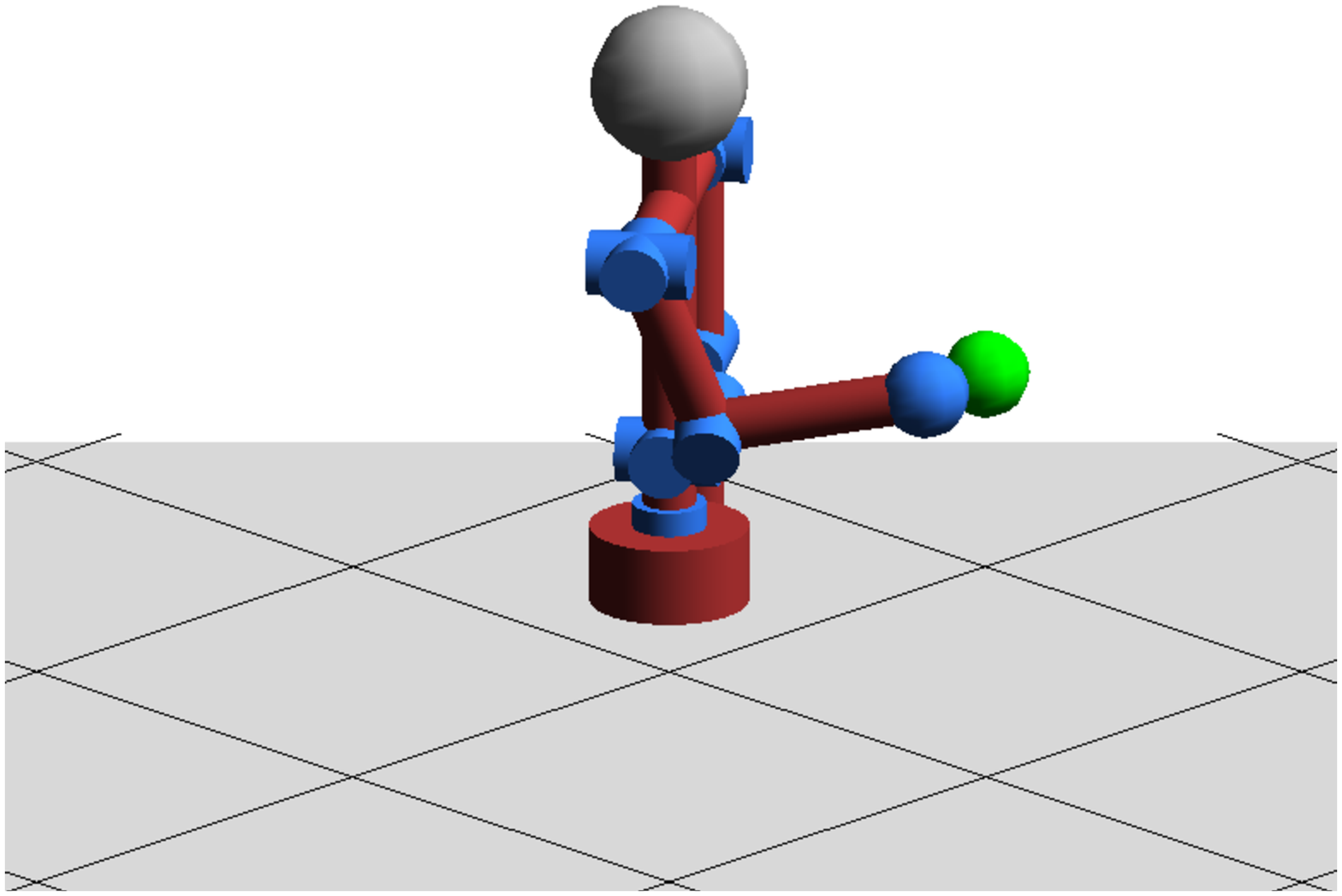}\label{fig:js4-s11}}}
\caption{Typical example of arm reaching with 4 degrees of freedom using the policy obtained by
IW-PGPE\textsubscript{OB} at the 50th iteration.}
\label{fig:J4_trajectory}
\end{figure}

Next, we evaluate the performance on the reaching task
with 4 degrees of freedom.
We use the right shoulder pitch, right elbow pitch, right shoulder roll, and torso yaw joint.
By using the torso yaw joint, the robot can reach a distant object
which can not be achieved by only using the right arm.
The results are shown in Figure \ref{fig:j4-return}.
The graph shows that IW-PGPE\textsubscript{OB} achieves fast policy improvement throughout iterations,
and the performance is the best among the compared methods.

Figure \ref{fig:J4_trajectory} depicts a representative example of object reaching
with $4$ degrees of freedom by IW-PGPE\textsubscript{OB}.
Note that the object is distant from the robot
and it can not be reached by only using the right arm.
The robot first adjusts the torso yaw joint, and then uses the right arm to reach the object.
The images show that the policy learned by our proposed method successfully leads the right hand to the distant object.

    \begin{figure}[t]
    \centering
    \includegraphics[clip,width=0.7\columnwidth]{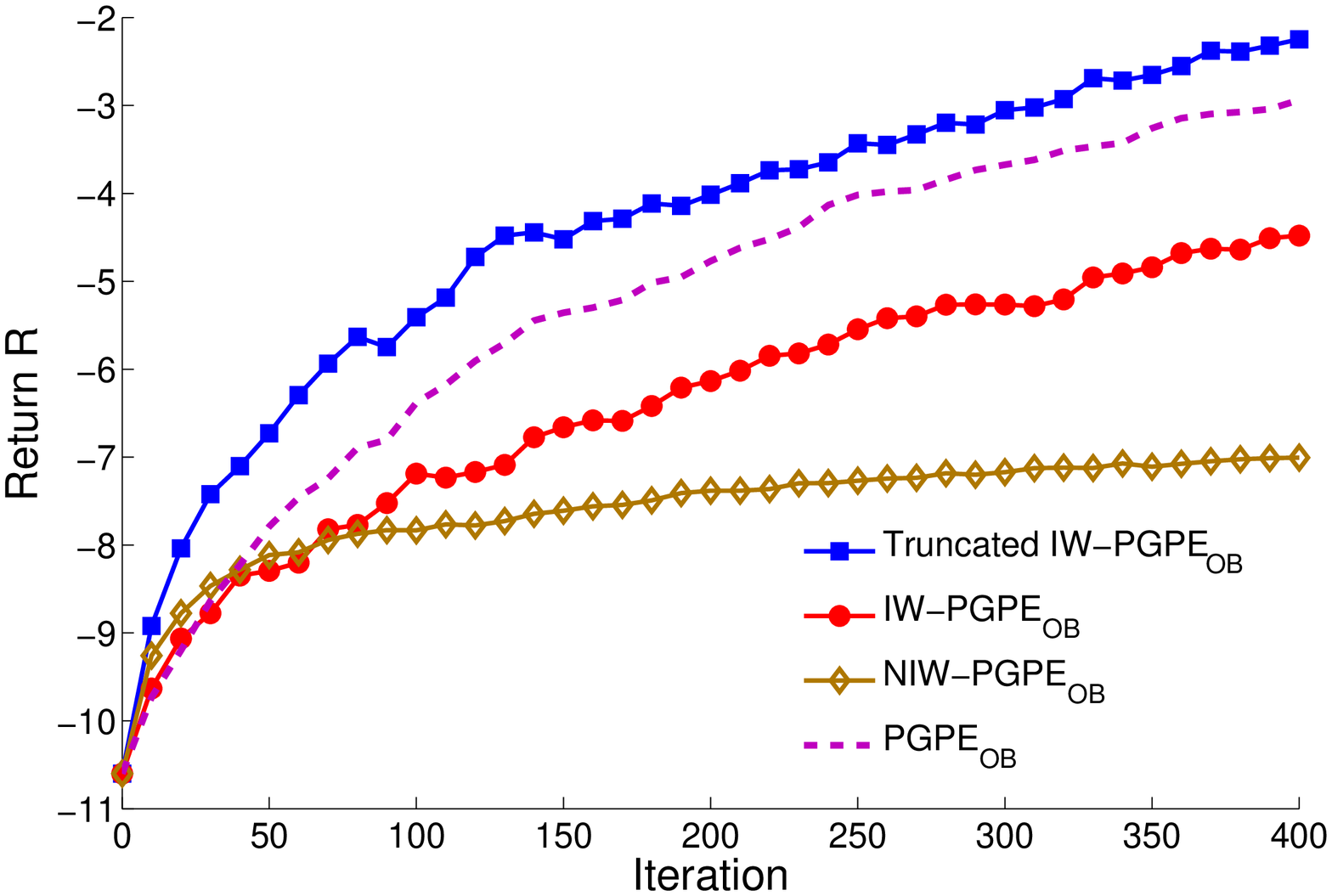}
    \caption{Average expected returns over $10$ runs as functions of the number of iterations for the reaching task with all degrees of freedom.}
    \label{fig:j9-return}
\end{figure}

\subsubsection{Reaching Task with All Degrees of Freedom}
At last, we evaluate the performance on the reaching task with all degrees of freedom.
The position of the target object is the same as the task in the 4-degrees-of-freedom setting.

In this experiment, we use all degrees of freedom to reach the object.
This increases the dimensionality of the state space, which actually
may grow the values of importance weights exponentially \cite{Shimodaira2000, IWbound}.
In order to mitigate the large values of importance weights,
we decided not to reuse all previously collected samples,
but only samples collected in the last 5 iterations.
This allows us to keep the difference between the sampling distribution
and the target distribution reasonably small, and thus the values of
importance weights
can be suppressed to some extent.
Furthermore, following \cite{truncated}, we truncate the importance weights as $w=\min\{w,2\}$.
This version of IW-PGPE\textsubscript{OB} is denoted as Truncated IW-PGPE\textsubscript{OB} below.

The results are shown in Figure \ref{fig:j9-return}.
The graph shows that the performance of Truncated IW-PGPE\textsubscript{OB} is the best,
which implies that the truncation of importance weights is helpful when applying our proposed method to high-dimensional problems.

Through all the arm-reaching experiments,
we can see that the returns tend to be lower as the dimension is increased,
even though we run the higher-dimensional experiment for a larger number of iterations.
In the task with all degrees of freedom (Figure \ref{fig:j9-return}), the largest number of iteration is 400.
If we continue the experiment for more iterations, the returns may sligtly increase, but are still less than the returns in the low-dimensional experiments.
This is because the more joints the robot uses, the larger energy will be consumed, and thus the returns tend to be lower in high-dimensional cases.

Overall, the proposed IW-PGPE\textsubscript{OB} is shown to be a promising method,
although in the last experiment it is obvious that just like other importance weight-based methods, the performance degrades in high-dimensional problems without the use of additional correction techniques such as weight truncation.

\section{Discussions and Conclusions}
In many real-world reinforcement learning problems, reducing the number of training samples
is desirable because the sampling cost is often much higher than the computational cost.
In this paper, we proposed a new policy gradient method equipped with efficient sample reuse,
which systematically combines a reliable policy gradient method, PGPE,
with importance sampling and the optimal constant baseline.
We showed that the introduction of the optimal constant baseline can mitigate
the large-variance problem of importance weighting under some conditions.
Through experiments with an artificial domain, the usefulness of the proposed method was demonstrated.
More over, through robotic experiments, we found that the truncation technique was helpful when applying the proposed method to high-dimensional problems.

The low variance of PGPE was brought by considering a deterministic policy and
introducing the stochasticity by drawing a policy parameter from a prior distribution.
This per-trajectory formulation was indeed shown to be useful
in reducing the variance of policy gradient estimates.
However, PGPE has limitations, too.
For example, the use of a finite horizon is essential in PGPE,
because the gradient estimates need full trajectories.
In particular, it is not straightforward to handle the infinite-horizon case.
Another issue is an extension to a partially-observable case.
It is known that for every finite Markov decision problem (MDP)
there exists a deterministic policy that is optimal \cite{book:Ross1983}.
However, in a partially-observable MDP (POMDP),
the best stationary stochastic policy can be arbitrarily better
than the best stationary deterministic policy \cite{Singh94learningwithout}.
Thus, the deterministic policy in PGPE can be a limitation
when extending it to the POMDP framework.
It is trivial to extend the current formulation to consider stochastic policies.
However, this may lead to an increase of variance and thus slow down convergence.
These issues need to be further investigated in the future work.

The baseline and importance weighting techniques are two independent techniques. More specifically, importance weighting is used in the off-policy scenario to efficiently reuse previously collected samples, by using importance weighting the consistency between the data sampling distribution and the target distribution is kept. On the other hand, the optimal constant baseline is used to reduce the variance of gradient estimates.

The use of a baseline technique has been first proposed in terms of reinforcement comparison in \cite{sutton:dissertation84}, which intuitively means the comparison between the expected return $R$ and the baseline $b$: If $R>b$ we adjust learned parameters $\rho$ so as to increase the probability of $\theta$, and, if $R<b$, we do the opposite.
Based on this idea, Williams \cite{Williams:88} demonstrated that a baseline technique did not introduce bias, which is because the expectation of the coefficient of $b$ is zero, i.e.,
$\bE\left[\frac{\nabla_{\rho} p(\theta|\rho)}{p(\theta|\rho)}\right]=0$.
The effect of the baseline on variance is considered in \cite{Dayan}.
The intuition behind the baseline is that subtracting a baseline from the return reduces the magnitude, and thus reduces the variance.
Technically, subtracting a baseline can be viewed as a \emph{control variate technique} \cite{book:Fishman:1996}, which is an effective approach to reducing variance of Monte Carlo estimates of integrals.
The experimental results in the paper suggest that the removal of the baseline is possibly the primary factor in improving performance
compared with the importance weighting techniques.

In episodic policy gradient methods, the optimal baseline which does not bias policy gradient estimates is given by a single scalar for all trajectories \cite{IROS:Peters+Schaal:2006}.
However, in the non-episodic policy gradient methods,
the optimal baseline can depend on the current state \cite{JMLR:Greensmith+Bartlett+Baxter:2004, morimura, Peters:2008}.
Thus, if a good parameterization for the baseline is known, e.g., in a generalized linear form $b(s_t)= \bm{w}^T \bm{\phi}(s_t)$,
this can significantly improve the gradient estimation process.
However, the selection of the basis function can be difficult and often impractical in robotics \cite{IROS:Peters+Schaal:2006}.
On the other hand, it is interesting to see that if the value function is used as the baseline function in non-episodic policy gradient methods, such as in \cite{Peters:2008, Sutton99policygradient},
the term $Q(s,a)-V(s)$ will lead to the \emph{advantage function} \cite{Baird93advantageupdating}, where $Q(s,a)$ is action value function and $V(s)$ is the value function.

\section*{Acknowledgements}
The authors would like to thank anonymous reviewers for their
feedback on our earlier manuscript, which highly contributed to
improving the readability of this paper.
TZ, VT, JM, and MS were supported by MEXT KAKENHI 23120004.
HH was supported by the FIRST program.
TZ was also supported by the MEXT scholarship,
VT was also supported by the JASSO scholarship,
and JM was also supported by the SRBPS and MEXT.

\appendix
\section*{Appendix}

In the appendix, we give proofs of the theorems.

\section{Proof of Theorem~\ref{theorem:variance-bound-PGPE}}
\label{proof:variance-bound-PGPE}

\begin{proof}
Due to the fact that the sampled data $\{\left(\bm{\theta}'_n, h'_n\right)\}_{n=1}^{N'}$ are independent and identically distributed, we have

\begin{align}
\label{def:meangra}
&\var\left[\nabla_{\bm{\eta}}\widehat{\cJ}_{\mathrm{IW}}(\bm{\rho})\right]
=\frac {1} {N'} \var\left[w(\bm{\theta}) \nabla_{\bm{\eta}} \log p(\bm{\theta}|\bm{\rho})R(h) \right],
\end{align}
where $h$ and $\bm{\theta}$ are random variables and follow the distributions $p(h, \bm{\theta}|\bm{\rho}')$.

Note that we consider the trace of the covariance matrix of gradient vectors, that is, the sum of the variance of the components of the vector.
Then by upper-bounding the variance with the second moment, we have the following upper bound:

\begin{align*}
&\var\left[w(\bm{\theta}) R(h) \nabla_{\bm{\eta}} \log p(\bm{\theta}|\bm{\rho}) \right]\\
&\le\sum_{i=1}^\ell \bE_{p(h,\bm{\theta}|\bm{\rho'})}\left[(w(\bm{\theta}) R(h) \nabla_{{\eta}_i} \log p(\bm{\theta}|\bm{\rho}))^2\right]\\
&=\sum_{i=1}^\ell \iint p(h|\bm{\theta})p(\bm{\theta}|\bm{\rho'}) \left(\frac{p(\bm{\theta}|\bm{\rho})}{p(\bm{\theta}|\bm{\rho}')}\right)^2 (R(h))^2 (\nabla_{{\eta}_i} \log p(\bm{\theta}|\bm{\rho}) )^2 \mathrm{d}h \mathrm{d}\bm{\theta}\\
&=\sum_{i=1}^\ell \iint p(h|\bm{\theta})p(\bm{\theta}|\bm{\rho}) w(\bm{\theta})(R(h))^2 (\nabla_{{\eta}_i} \log p(\bm{\theta}|\bm{\rho}) )^2 \mathrm{d}h \mathrm{d}\bm{\theta}\\
&\le \sum_{i=1}^\ell\left(\frac{\beta(1-\gamma^T)}{1-\gamma}\right)^2  w_{\max}  \iint p(h|\bm{\theta})p(\bm{\theta}|\bm{\rho})(\nabla_{{\eta}_i} \log p(\bm{\theta}|\bm{\rho}) )^2 \mathrm{d}h \mathrm{d}\bm{\theta}\\
&=\sum_{i=1}^\ell\left(\frac{\beta(1-\gamma^T)}{1-\gamma}\right)^2  w_{\max} \bE_{p(\bm{\theta}|\bm{\rho})}\left[(\nabla_{{\eta}_i} \log p(\bm{\theta}|\bm{\rho}) )^2\right],
\end{align*}
where $\bE_{p(\bm{\theta}|\bm{\rho})}[\cdot]$ denotes the expectation of the function of random variable $\bm{\theta}$ with respect to $\bm{\theta} \sim p(\bm{\theta}|\bm{\rho})$.
Subsequently, given the proof of the first part of Theorem 1 in \cite{NN2012-ting},
we get the upper bound of $\var\left[\nabla_{\bm{\eta}}\widehat{\cJ}_{\mathrm{IW}}(\bm{\rho})\right]$.

Similarly, given the same technique and the proof of the later part of Theorem 1 in \cite{NN2012-ting},
we could get the conclusion of the upper bound of $\var\left[\nabla_{\bm{\tau}}\widehat{\cJ}_{\mathrm{IW}}(\bm{\rho})\right]$.

\end{proof}

\section{Proof of Theorem~\ref{theorem:optimal-baseline}}
\label{proof:optimal-baseline}

\begin{proof}
First, let us derive some elementary expressions.
Let $\bm{A}$, $\bm{C}$ be random variables taking values in the $\ell$-dimensional space and let $b$ be a scalar.
Then,
\[\var[\bm{A}-b\bm{C}]=\var[\bm{A}]+b^2\var[\bm{C}]-b\cov[\bm{A},\bm{C}]-b\cov[\bm{C},\bm{A}].\]
We still consider the trace of the covariance matrix of gradient vectors for multi-dimensional space.
Assume that $\bE[\bm{C}]=\bm{0}$.
Then, we could have
\begin{align}
\var[\bm{A}-b\bm{C}]=&\var[\bm{A}]+b^2\var[\bm{C}]-2b\cov[\bm{A},\bm{C}] \nonumber\\
=& \var[\bm{A}]+ \bE[\bm{C}^\T\bm{C}] \left\{b^2-2b\frac{\bE[\bm{A}^\T\bm{C}]}{\bE[\bm{C}^\T\bm{C}]}\right\} \label{varb}\\
=& \var[\bm{A}]+\bE[\bm{C}^\T\bm{C}]\left\{ \left(b-\frac{\bE[\bm{A}^\T\bm{C}]}{\bE[\bm{C}^\T\bm{C}]}\right)^2-\left(\frac{\bE[\bm{A}^\T\bm{C}]}{\bE[\bm{C}^\T\bm{C}]}\right)^2\right\} \nonumber.
\end{align}
Simple calculus shows that the foregoing is minimized when
\[b=\frac{\bE[\bm{A}^\T\bm{C}]}{\bE[\bm{C}^\T\bm{C}]}.\]
The optimal baseline for IW-PGPE follows immediately by plugging in \[\bm{A}=Rw \nabla_{\bm{\rho}} \log p(\bm{\theta}|\bm{\rho}) \] and
\[\bm{C}=w\nabla_{\bm{\rho}} \log p(\bm{\theta}|\bm{\rho})\] for $\bm{A}$ and $\bm{C}$.
Note that Eq.(\ref{varb}) uses the conclusion of
$\bE[w\nabla_{\bm{\rho}} \log p(\bm{\theta}|\bm{\rho})]=\bm{0}$, which can be found in the proof of Theorem 4 in \cite{NN2012-ting}.

As the sampled data are independent and identically distributed, we have
\[\var[\nabla_{\bm{\rho}} \hat{\cJ}_{\mathrm{IW}}^b (\bm{\rho})]=\frac{1}{N'}\var[\bm{A}-b\bm{C}].\]
Then, according to Eq.(\ref{varb}) and the definition of $b^*$, we could have
\begin{align*}
&\var[\nabla_{\bm{\rho}}\widehat{\cJ}^b_{\mathrm{IW}}(\bm{\rho})]-\var[\nabla_{\bm{\rho}} \widehat{\cJ}^{b^*}_{\mathrm{IW}}(\bm{\rho})]\\
&=\frac{1}{N'} \left( b^2\bE[\bm{C}^\T \bm{C}]-2b\bE[\bm{A}^\T \bm{C}]+\frac{(\bE[\bm{A}^\T \bm{C}])^2}{\bE[\bm{C}^\T \bm{C}]}\right)\\
&=\frac{1}{N'} \left(b-b^*\right)^2 \bE [\bm{C}^\T \bm{C}],
\end{align*}
where the expectation is over random variables $h$ and $\bm{\theta}$ such that $(h,\bm{\theta}) \sim p(h,\bm{\theta}|\bm{\rho}')$.
This completes the proof of Theorem \ref{theorem:optimal-baseline}.\qedhere
\end{proof}

\section{Proof of Theorem~\ref{theorem:variance-bound-gap}}
\label{proof:variance-bound-gap}

\begin{proof}
 We define $\nabla_{\bm{\eta}}$ and $\nabla_{\bm{\eta}_i}$ as
\begin{align*}
\nabla_{\bm{\eta}}= & \nabla_{\bm{\eta}} \log p(\bm{\theta}|\bm{\rho}),\\
\nabla_{\bm{\eta}_i}=&\nabla_{\bm{\eta}_i} \log p(\bm{\theta}|\bm{\rho}).
\end{align*}
We still denote the subscripts $\bm{\rho}'$ as $p(h,\bm{\theta}|\bm{\rho}')$.
According to Theorem~\ref{theorem:optimal-baseline}, by setting $b=0$, it is easy to know that
\[
\var\left[\nabla_{\bm{\eta}} \widehat{\cJ}_{\mathrm{IW}}(\bm{\rho})\right]-\var\left[\nabla_{\bm{\eta}} \widehat{\cJ}_{\mathrm{IW}}^{b^*}(\bm{\rho})\right]
=\frac{\left(\bE_{\bm{\rho}'}[R(h)w^2(\bm{\theta}) \nabla_{\bm{\eta}}^\T\nabla_{\bm{\eta}}]\right)^2}{N'\bE_{\bm{\rho}'}[w^2(\bm{\theta})\nabla_{\bm{\eta}} ^\T \nabla_{\bm{\eta}}]}.
\]
We already know that
\[
\bE_{\bm{\rho}'}[R(h)w^2(\bm{\theta})\nabla_{\bm{\eta}}^\T\nabla_{\bm{\eta}}]
\le \frac{\beta(1-\gamma^T)}{(1-\gamma)}\bE_{\bm{\rho}'}\left[w^2(\bm{\theta}) \nabla_{\bm{\eta}}^\T\nabla_{\bm{\eta}}\right].
\]
Hence,
\begin{align}
&\var\left[\nabla_{\bm{\eta}}\widehat{\cJ}_{\mathrm{IW}}(\bm{\rho})\right]-\var\left[\nabla_{\bm{\eta}} \widehat{\cJ}_{\mathrm{IW}}^{b^*}(\bm{\rho})\right] \nonumber\\
&\le\frac{\beta^2(1-\gamma^T)^2}{N'(1-\gamma)^2}\bE_{\bm{\rho}'}\left[w^2(\bm{\theta}) \nabla_{\bm{\eta}}^\T\nabla_{\bm{\eta}}\right]\nonumber\\
&\le \frac{\beta^2(1-\gamma^T)^2}{N'(1-\gamma)^2}w_{\max} \sum_{i=1}^{\ell} \bE_{p(\bm{\theta}|\bm{\rho})}\left[(\nabla_{{\eta}_i})^2\right] \label{uppervar1} \\
&= \frac{\beta^2(1-\gamma^T)^2B}{N'(1-\gamma)^2}w_{\max} \label{uppervar2} ,
\end{align}
where Eq.(\ref{uppervar1}) is based on the same technique used in Section \ref{proof:variance-bound-PGPE},
and Eq.(\ref{uppervar2}) is given by results of the proof of Theorem 1 in \cite{NN2012-ting}.

Similarly, we can have the lower bound as
\[
\var\left[\nabla_{\bm{\eta}}\widehat{\cJ}_{\mathrm{IW}}(\bm{\rho})\right]-\var\left[\nabla_{\bm{\eta}} \widehat{\cJ}_{\mathrm{IW}}^{b^*}(\bm{\rho})\right] \geq
\frac{\alpha^2(1-\gamma^T)^2B}{N'(1-\gamma)^2}w_{\min}.
\]

By using the same techniques, we get the bounds of the variance reduction of gradient estimation with respect to the deviation parameter $\bm{\tau}$,
\begin{align*}
\var\left[\nabla_{\bm{\tau}}\widehat{\cJ}_{\mathrm{IW}}(\bm{\rho})\right]-\var\left[\nabla_{\bm{\tau}} \widehat{\cJ}_{\mathrm{IW}}^{b^*}(\bm{\rho})\right] &\le \frac{2\beta^2(1-\gamma^T)^2B}{N'(1-\gamma)^2}w_{\max},\\
\var\left[\nabla_{\bm{\tau}}\widehat{\cJ}_{\mathrm{IW}}(\bm{\rho})\right]-\var\left[\nabla_{\bm{\tau}} \widehat{\cJ}_{\mathrm{IW}}^{b^*}(\bm{\rho})\right] &\ge
\frac{2\alpha^2(1-\gamma^T)^2B}{N'(1-\gamma)^2}w_{\min},
\end{align*}
which completes the proof.
\end{proof}

 \bibliographystyle{plain}

\end{document}